\documentclass[a4paper,fleqn]{cas-dc}
\usepackage[authoryear,longnamesfirst]{natbib}

\usepackage{amsmath,amsfonts}
\usepackage{algorithmic}
\usepackage{algorithm}
\usepackage[english]{babel}
\usepackage{array}
\usepackage{textcomp}
\usepackage{stfloats}
\usepackage{url}
\usepackage{verbatim}
\usepackage{graphicx}
\usepackage{orcidlink}
\usepackage{booktabs}
\usepackage{graphicx}
\usepackage{subcaption}
\usepackage{tcolorbox}
\usepackage{hyperref}
\usepackage{xcolor}

\def\tsc#1{\csdef{#1}{\textsc{\lowercase{#1}}\xspace}}
\tsc{WGM}
\tsc{QE}
\tsc{EP}
\tsc{PMS}
\tsc{BEC}
\tsc{DE}

\ExplSyntaxOn
\cs_gset:Nn \__first_footerline:
  {
    \rmfamily \itshape
    Accepted~manuscript.~
    \href{https://doi.org/10.1016/j.neunet.2026.109375}
    {Neural~Networks,~doi:10.1016/j.neunet.2026.109375}
  }
\ExplSyntaxOff

\begin{document}
\emergencystretch 1em
\let\WriteBookmarks\relax
\def\floatpagepagefraction{1}
\def\textpagefraction{.001}
\shorttitle{NeuCoReClass AD}
\shortauthors{Sánchez-Ferrera et al.}

\author[1]{Aitor Sánchez-Ferrera}[orcid=0000-0001-6127-0686]
\ead{aitor.sanchezf@ehu.eus}
\cormark[1]
\author[1]{Usue Mori}[orcid=0000-0002-2057-1770]
\ead{usue.mori@ehu.eus}
\author[1]{Borja Calvo}[orcid=0000-0001-9969-9664]
\ead{borja.calvo@ehu.eus}
\author[1,2]{Jose A. Lozano}[orcid=0000-0002-4683-8111]
\ead{ja.lozano@ehu.eus}

\affiliation[1]{organization={Department of Computer Science and Artificial Intelligence, University of the Basque Country UPV/EHU},
addressline={Manuel Lardizabal 1},
postcode={20018},
city={Donostia-San Sebastian},
country={Spain}}

\affiliation[2]{organization={Basque Center for Applied Mathematics},
addressline={Mazarredo Zumarkalea 14},
postcode={48009},
city={Bilbao},
country={Spain}}

\cortext[cor1]{Corresponding author}

\title [mode = title]{NeuCoReClass AD: Redefining Self-Supervised Time Series Anomaly Detection}                      

\begin{abstract}
Time series anomaly detection plays a critical role in a wide range of real-world applications. Among unsupervised approaches, self-supervised learning has gained traction for modeling normal behavior without the need of labeled data. However, many existing methods rely on a single proxy task, limiting their ability to capture meaningful patterns in normal data. Moreover, they often depend on handcrafted transformations tailored to specific domains, hindering their generalization across diverse problems. To address these limitations, we introduce NeuCoReClass AD, a self-supervised multi-task time series anomaly detection framework that combines contrastive, reconstruction, and classification proxy tasks. Our method employs neural transformation learning to generate augmented views that are informative, diverse, and coherent, without requiring domain-specific knowledge.
We evaluate NeuCoReClass AD across a wide range of benchmarks, demonstrating that it consistently outperforms both classical baselines and most deep-learning alternatives. Furthermore, it enables the characterization of distinct anomaly profiles in a fully unsupervised manner.
Our code is publicly available at \href{https://github.com/Aitorzan3/NeuCoReClass-AD}{https://github.com/Aitorzan3/NeuCoReClass-AD}.

\end{abstract}


\maketitle

\section{Introduction}
Time series refer to collections of measurements indexed in chronological order, describing the behavior of a system or entity over time \citep{hamilton2020time}. Among the various tasks in time series data mining, such as classification, forecasting, and clustering, anomaly detection has gained increasing attention in recent years. Time series anomalies—also referred to as outliers, novelties, or out-of-distribution samples—are abnormal events where the behavior of the system described by the time series deviates from an expected normality \citep{carreno2020analyzing}. 

Time series anomaly detection (TSAD) aims to identify these events and has broad applications in intrusion detection \citep{cook2019anomaly}, financial fraud detection \citep{hilal2022financial}, and healthcare \citep{pereira2019learning}, among other fields. In the literature, two main types of TSAD problems are distinguished: (a) identifying abnormal points and subsequences within an individual time series, and (b) detecting anomalous time series in a database comprising multiple complete time series \citep{sanchez2025review}. In this work we address the latter. 

Due to the difficulties of employing supervised learning in anomaly detection contexts (e.g., cost of labeling, class imbalance), unsupervised techniques have gained attention in the past few years \citep{chandola2009anomaly}. In this setting, machine learning models are trained on a dataset that is assumed to contain only normal samples or, at worst, a negligible number of anomalous samples. Once trained, they make use of the learned characteristics about normal data to compute an abnormality degree—commonly referred to as the anomaly score—for new samples \citep{chandola2009anomaly}. 

Within the unsupervised proposals, recent approaches seek to improve representation learning to enhance anomaly detection, with self-supervised learning becoming increasingly prominent \citep{zamanzadeh2024deep}. In self-supervised anomaly detection, models are trained on pretext tasks to learn representations that capture relevant patterns and characteristics of normal training data \citep{liu2021self}. These tasks involve predicting specific data parts, attributes, or relationships within the data \citep{liu2021self}. Then, the learned data representations are leveraged to perform the downstream task, which represents the final objective—in this case, anomaly detection. 

The main difference among self-supervised works in the literature lies in the choice and design of the pretext tasks used to train the models \citep{hojjati2022self}. Typically, normal training samples are transformed to generate augmented views, and proxy tasks are designed to model specific relationships between the original samples and these views. In practice, most methods rely on manually defined transformations tailored to a single proxy task. However, such transformations are often task-dependent and require dataset-specific design choices, which restricts the generality of these methods across diverse anomaly detection problems \citep{yoo2022data}.

To address this limitation, recent approaches have replaced manually defined transformations with learnable neural transformations that generate task-compatible views learned directly from normal training data \citep{qiu2021neural}. While this strategy improves adaptability across datasets, most existing methods still optimize a single proxy task. This constitutes a key limitation, as prior research suggests that combining complementary proxy tasks can improve representation learning \citep{doersch2017multi}. Therefore, extending learnable neural transformations to a multi-task self-supervised framework for TSAD remains largely underexplored.

In this paper, we propose a novel self-supervised TSAD method that integrates learnable neural transformations with a multi-task learning framework. The neural transformations generate augmented views that are suitable for multiple proxy tasks, including contrastive learning, self-supervised reconstruction, and self-supervised classification. Since the transformations are learned from data rather than manually designed, the proposed framework can generalize across diverse datasets and anomaly detection scenarios. Based on the proxy tasks considered, we refer to this method as \textbf{NeuCoReClass AD}.

The contributions of this work are as follows: 

\begin{itemize}
    \item We present a self-supervised TSAD method that combines multiple proxy tasks to improve representation learning, leading to better characterization of normal data patterns and enhanced anomaly detection performance.
    \item We introduce a loss function and architecture that ensure that the augmented views satisfy the requirements of the proxy tasks in our method, enabling their joint optimization.
    \item We propose a method that not only detects anomalies with high performance, but also enables the characterization of different anomaly types in an unsupervised manner.
    \item We release our implementation as open-source code to facilitate reproducibility and future research.
\end{itemize}

The remainder of this paper is organized as follows. Section \ref{sec:relatedwork} reviews the related work. Section \ref{sec:method} presents the proposed method. Section \ref{sec:experimentalsetup} describes the experimental setup and evaluation protocol. Section \ref{sec:results} reports the experimental results and analysis. Finally, Section \ref{sec:conclusions} concludes the paper and outlines directions for future research.

\section{Related Work}
\label{sec:relatedwork}
In self-supervised TSAD, models are trained on proxy tasks to learn representations that capture the underlying structure of normal data. Existing methods mainly differ in (i) the proxy tasks used to learn representations of normality and (ii) the mechanisms used to generate augmented views. In this section, we review self-supervised TSAD from these two perspectives.

\subsection{Proxy Tasks for Self-Supervised TSAD}

Self-supervised TSAD methods can be broadly categorized according to the proxy task used to learn representations of normality. The most common families are reconstruction, self-supervised classification, and contrastive learning \citep{hojjati2022self,sanchez2025review}.

\textbf{Reconstruction-based tasks.} These methods train an autoencoder that attempts to reconstruct normal samples and use the reconstruction error as the anomaly score \citep{hojjati2022self}. To improve generalization, several approaches reconstruct the original signal from transformed inputs that serve as augmented views, introducing variability that encourages the model to capture robust normal patterns rather than memorizing the training set \citep{fu2022mad,liu2021deepfib,sakurada2014anomaly,marchi2015novel}.

\textbf{Self-supervised classification tasks.} In this setting, predefined transformations are applied to normal samples to generate augmented views, and a classifier is trained to predict which transformation generated each view \citep{zhang2021self}. At inference, the anomaly score of a new sample is the classification error across its augmented views \citep{hojjati2022self}.

\textbf{Contrastive learning}. This paradigm has been widely adopted for unsupervised representation learning in time series \citep{yue2022ts2vec,zhang2024smde}. The objective of contrastive learning is to train an encoder to generate representations of samples and their augmented views in a structured latent space \citep{jaiswal2020survey}. Specifically, the encoder is encouraged to bring similar samples and augmented views (positive pairs) closer together in the latent space, while pushing dissimilar ones (negative pairs) apart. The similarities and dissimilarities between samples are imposed by the contrastive loss. For instance, TS2Vec \citep{yue2022ts2vec} defines positive pairs as different augmented views of the same time series, while treating views from different time series as negative pairs. In anomaly detection, at inference time, the anomaly score of a new sample is computed based on how well the representations of its augmented views conform to this learned structure.

Overall, these proxy tasks are complementary: reconstruction focuses on fidelity, classification on transformation awareness, and contrastive learning on latent-space structure. This motivates multi-task formulations that jointly combine these objectives to obtain richer representations of normal data.

\subsection{Transformation Design in Self-Supervised TSAD}

The effectiveness of self-supervised TSAD strongly depends on how augmented views are generated. Different proxy tasks impose different structural requirements on these views. For reconstruction-based tasks, augmented views should introduce sufficient variability, so that models learn robust patterns rather than memorizing to reconstruct training samples. For self-supervised classification tasks, augmented views should disrupt normality in controlled ways and be sufficiently diverse, so that each transformation induces distinguishable patterns. In contrastive settings, requirements are less tied to explicit input-space constraints, since latent representations are shaped by the similarity assumptions encoded in the contrastive objective. Importantly, across all proxy tasks, augmented views should preserve the relevant semantics of the original sample, as inference is ultimately performed from transformed views and their representations.

\textbf{Manually designed transformations.} Early TSAD methods commonly rely on handcrafted transformations tailored to each proxy task. In reconstruction-based settings, transformations such as masking and noise injection are used to increase input variability while retaining enough information to reconstruct the original sample \citep{fu2022mad,liu2021deepfib,sakurada2014anomaly,marchi2015novel}. In self-supervised classification, transformations are usually chosen according to domain-specific anomaly mechanisms to disrupt normality. For instance, \citet{blazquez2021water} generate views by applying different upscaling magnitudes for water leak detection, where upscaling normal sequences emulates abnormal flow behaviors linked to leaks. Likewise, \citet{zheng2022task} use amplitude and frequency perturbations in EEG signals to induce seizure-related distortions. Beyond disruption, diversity across transformations is encouraged so that the resulting transformation-induced classes are separable, yielding more discriminative anomaly scores. In contrastive learning, transformations are designed according to the relationships that the model is expected to capture in the latent space. For example, some works inject noise to normal samples to generate augmented views that serve as positive pairs, encouraging invariance to noise \citep{jaiswal2020survey}. Although effective, the use of manually defined transformations is usually task- and dataset-dependent, which requires domain expertise, and may transfer poorly across different anomaly detection problems \citep{yoo2022data,sanchez2025review}.

\textbf{Learnable neural transformations.} To reduce dependence on expert knowledge and improve transferability across TSAD problems, recent work introduces neural transformation learning, where transformations are parameterized by neural networks and learned from normal training data \citep{qiu2021neural}. These transformations are learned with contrastive learning so that the generated views satisfy desired latent-space relations. A key method is NeuTraL AD \citep{qiu2021neural}, which jointly learns both neural transformations and the encoder by optimizing the Deterministic Contrastive Loss (DCL). The DCL encourages similarity between each sample and its transformed views (positive pairs) while enforcing diversity among the views (negative pairs), promoting informative and complementary representations of normality. Despite its relevance, NeuTraL AD optimizes a single proxy task and does not consider the integration of learnable neural transformations with multiple complementary proxy tasks.

\section{Proposed Method: NeuCoReClass AD} \label{sec:method}

\begin{figure*}[ht]
    \centering

    \begin{subfigure}{0.9\textwidth}
        \centering
        \begin{tcolorbox}[colframe=black!30, colback=white, boxrule=0.3pt, arc=0pt, left=2pt, right=2pt, top=2pt, bottom=2pt]
            \includegraphics[width=\linewidth]{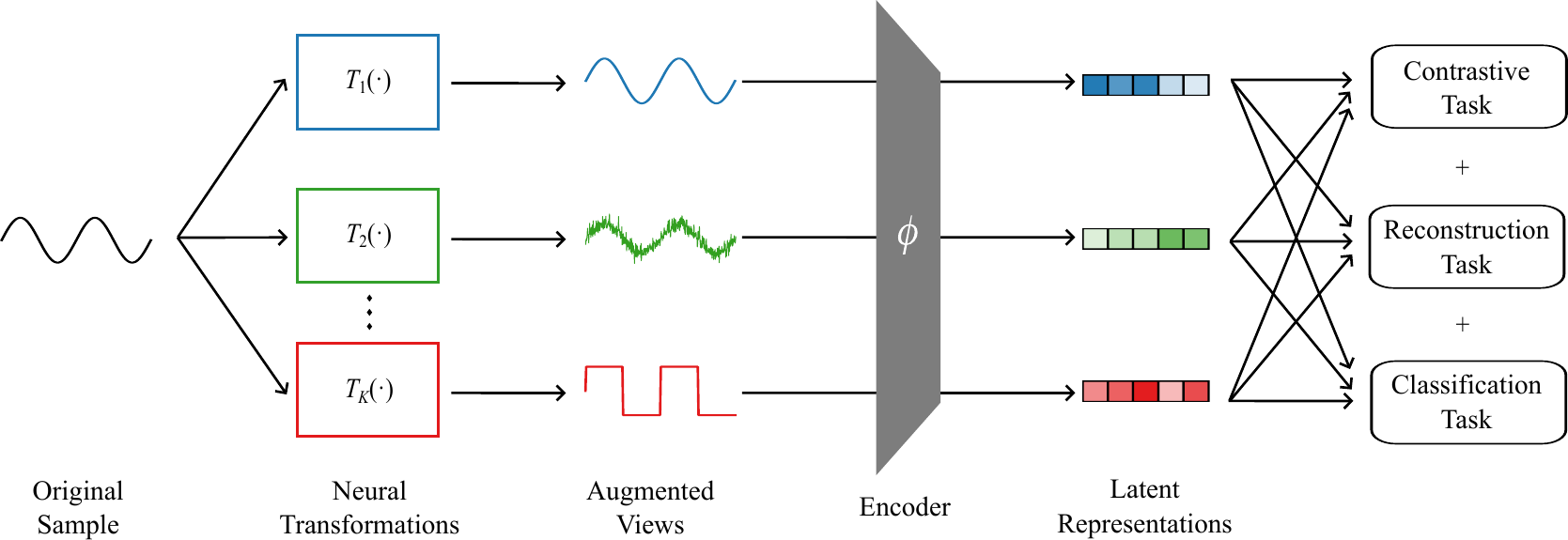}
        \end{tcolorbox}
        \caption{Pipeline of NeuCoReClass AD. Each sequence is processed by a set of $K$ neural transformations to generate diverse augmented views, with the first one representing the identity. These views are then encoded into latent representations, which serve as the input for three self-supervised proxy tasks: contrastive learning, reconstruction, and classification.}
        \label{fig:pipeline}
    \end{subfigure}

    \vspace{1em}

    \begin{subfigure}{0.32\textwidth}
        \centering
        \begin{tcolorbox}[colframe=black!30, colback=white, boxrule=0.3pt, arc=0pt, left=2pt, right=2pt, top=1.5pt, bottom=2pt, height=5.2cm, valign=center]
            \includegraphics[width=\linewidth]{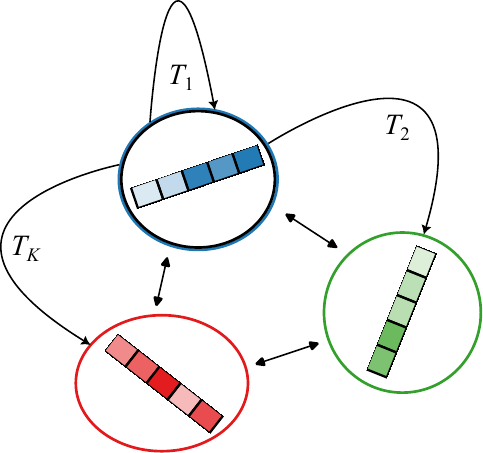}
        \end{tcolorbox}
        \caption{Contrastive task. The representations of the positive pairs are pulled together, while pushing apart the rest.}
        \label{fig:contrastive}
    \end{subfigure}
    \hfill
    \begin{subfigure}{0.32\textwidth}
        \centering
        \begin{tcolorbox}[colframe=black!30, colback=white, boxrule=0.3pt, arc=0pt, left=2pt, right=2pt, top=2pt, bottom=2pt, height=5.2cm, valign=center]
            \includegraphics[width=\linewidth]{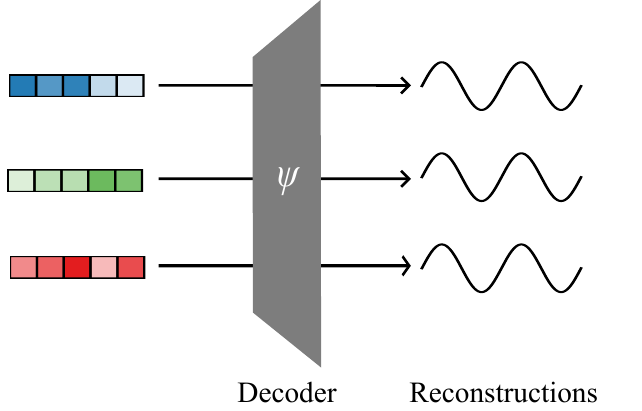}
        \end{tcolorbox}
        \caption{Reconstruction task. The latent representations of each augmented view are used to reconstruct the original input sequence.}
        \label{fig:reconstruction}
    \end{subfigure}
    \hfill
    \begin{subfigure}{0.32\textwidth}
        \centering
        \begin{tcolorbox}[colframe=black!30, colback=white, boxrule=0.3pt, arc=0pt, left=2pt, right=2pt, top=2pt, bottom=2pt, height=5.2cm, valign=center]
            \includegraphics[width=\linewidth]{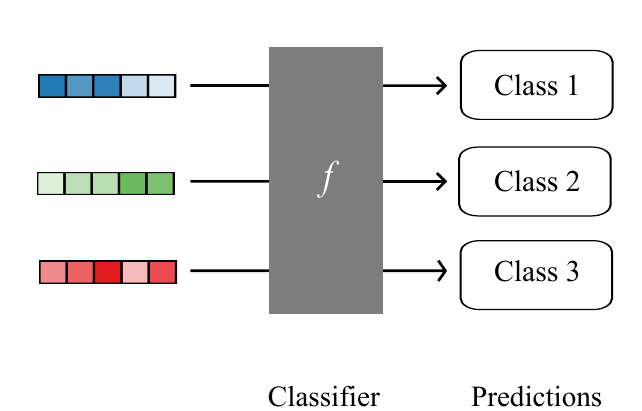}
        \end{tcolorbox}
        \caption{Classification task. The latent representations of the augmented views are used to predict the transformation that generated them.}
        \label{fig:classification}
    \end{subfigure}
    \caption{Overview of NeuCoReClass AD. The top panel shows the generation process of the augmented views and their encoding into the latent space. The bottom row illustrates the three self-supervised learning tasks used to guide the training of the method. Different colors denote different transformations.}
    \label{fig:method}
\end{figure*}

In this paper, we propose NeuCoReClass AD, a self-supervised TSAD method that integrates learnable neural transformations within a multi-task learning framework to generate augmented views for contrastive, reconstructive, and classification proxy tasks for TSAD. The pipeline and learning objectives of NeuCoReClass AD are depicted in Figure \ref{fig:method}.

In Section \ref{sec:architecture}, we describe the architecture of NeuCoReClass AD. Following, Section \ref{sec:revisiting} defines the desired properties that the generated augmented views should satisfy to be effective across the proxy tasks considered in our approach. Section \ref{sec:training} details the training process of NeuCoReClass AD. Finally, Section \ref{sec:detection} presents how the representations of the augmented views of new samples can be used to compute their anomaly scores.

\subsection{Architecture of NeuCoReClass AD} \label{sec:architecture}

Our model’s pipeline consists of four key components: (i) a neural transformation module, (ii) a feature extraction module, (iii) a reconstruction module, and (iv) a linear classification module. These modules enable the model to learn to perform the three proxy tasks.

\subsubsection{Neural Transformation Module}

The neural transformation module comprises a set of $K$ learnable neural transformations $\mathcal{T} = \lbrace T_1, ..., T_K \rbrace$. These are neural networks representing parameterized functions whose parameters $\theta_{\mathcal{T}} = \lbrace \theta_{T_1}, ..., \theta_{T_K} \rbrace$ are learned via gradient-based optimization. When applied to an input sample $x_i$, these transformations generate a set of augmented views associated with the original sample $\mathcal{T}(x_i) = \lbrace T_1(x_i), ..., T_K(x_i) \rbrace = \lbrace x_i^1, ..., x_i^K \rbrace$, where $x_i^k$ denotes the augmented view that is generated by applying the $k$-th transformation $T_k(\cdot)$ to that sample. Following the existing literature on self-supervised classification for anomaly detection \citep{sanchez2025review}, the first transformation $T_1(\cdot)$ in the neural transformation module is the identity function, i.e., $T_1(x_i) = x_i$.

\subsubsection{Feature Extraction Module}

The feature extraction module consists of an encoder $ \phi(\cdot) $ that maps the augmented views, generated by the neural transformation module, into a more compact and abstract latent space. We denote the parameters of the encoder $ \phi(\cdot) $ as $ \theta_\phi$. After generating the augmented views of a sample $x_i$, their corresponding latent representations are computed using the encoder $\phi(\cdot)$, resulting in $\phi(\mathcal{T}(x_i)) = \lbrace z_i^1, ..., z_i^K \rbrace$.

\subsubsection{Reconstruction Module}
The reconstruction module is based on a decoder $\psi(\cdot)$, which is added atop the encoder $\phi(\cdot)$. The parameters of the decoder are denoted as $\theta_{\psi}$. The decoder reconstructs the original sample $x_i$ from each latent view $z_i^k$, such that ideally $\psi(z_i^k) = x_i$, for $1 \leq k \leq K$.

\subsubsection{Linear Classification Module}
Finally, the linear classification module employs a linear classifier $f(\cdot)$, whose parameters are denoted as $\theta_f$. The classifier takes the latent representations of the augmented views associated with a sample $\{z_i^1, \dots, z_i^K\}$ as inputs and performs self-supervised classification. In other words, it predicts which transformation was applied to the original input sample for generating each of its augmented views, i.e., ideally \(f(z_i^k) = k\) for \(1 \leq k \leq K\).

\subsection{Revisiting Neural Transformation Learning for TSAD} \label{sec:revisiting}

The conditions that the augmented views of the samples must satisfy in the latent space are explicitly derived from those established in previous works for each of the respective proxy tasks \citep{sanchez2025review}.

We assume that each transformation induces a coherent latent subspace, where the representations of its associated augmented views are positioned close to each other. In addition, we require the latent representations of the augmented views to fulfill the following rules:

\begin{enumerate}
    \item \textbf{Disruption}. The augmented views must represent ``corrupted'' versions of their corresponding original samples, thereby disrupting the normal structure of the data.
    \item \textbf{Diversity}. They must also reflect diverse variations, such that the augmented views capture complementary information and are distinguishable from each other.
    \item \textbf{Preservation}. The augmented views must preserve relevant semantic information from their corresponding original samples.
\end{enumerate}

The disruption rule is derived from prior work in self-supervised classification \citep{yoo2022data}. The diversity rule is a shared requirement across reconstruction- and classification-based proxy tasks. Finally, the preservation rule is inherited from the neural transformation learning framework proposed by \citet{qiu2021neural} and is enforced across all considered proxy tasks. Figure \ref{fig:contrastive} shows an example latent space structure that meets these criteria.

\subsection{Training NeuCoReClass AD}\label{sec:training}

We consider three learning objectives that enable NeuCoReClass AD to jointly optimize the contrastive, reconstruction, and classification proxy tasks. These objectives are designed to enforce the rules introduced in Section \ref{sec:revisiting}: the contrastive and classification losses promote the disruption and diversity rules, while the reconstruction loss supports the preservation rule.

During the forward pass, a minibatch of $B$ normal samples is randomly sampled from the training set. For each sample $x_i$, the neural transformation module and the encoder are used to generate latent representations of its $K$ augmented views, denoted as $\lbrace z_i^1, ..., z_i^K \rbrace$, for $1 \leq i \leq B$ (see Figure~\ref{fig:pipeline}).
These latent representations serve as the basis for learning the three proxy tasks. The losses corresponding to the three objectives are then computed and combined in a multi-task learning fashion. Finally, during the backward pass, gradients are propagated and model parameters are updated using the mean loss over the minibatch.

\subsubsection{Contrastive Loss} 

To encourage coherent representations of augmented views, the embedding of an augmented view $z_i^k$ should be close to the embeddings of other samples' augmented views $z_j^k$ that are generated by the same transformation $T_k(\cdot)$, with $j \neq i$ and $1 \leq j \leq B$. Simultaneously, to promote diversity, each representation $z_i^k$ is pushed away from the embeddings of augmented views from all the samples in the minibatch that are generated using different transformations $z_j^q$, with $q \neq k$. In our framework, the inclusion of the identity transformation ensures that this separation is also defined with respect to normal samples, so that the contrastive objective promotes not only diversity but also disruption at the representation level (see Figure \ref{fig:contrastive}).

Following prior work on contrastive learning by \citet{chen2020simple}, we define the similarity score between two augmented views in the latent space as follows: 

\begin{equation}
    h(x_i^k, x_j^k) = \exp\left({\textnormal{sim}(z_i^k, z_j^k)} / \tau \right),
\end{equation}

\noindent where $\textnormal{sim}(\cdot,\cdot)$ denotes the cosine similarity measure, and $\tau$ is a temperature hyperparameter that controls the sharpness of the contrastive loss. Lower values of $\tau$ amplify the distinction between positive and negative samples, while higher values produce a smoother separation.

To enforce that the latent space of our model is structured according to both the disruption and diversity criteria, we define the following contrastive loss for each sample $x_i$ in the minibatch:

\begin{equation}\label{eq:contrastive}
\ell_{\text{con}}(x_i) = -\frac{1}{K} \sum_{k=1}^K \log \left( \frac{\textnormal{Pos}_k}{\textnormal{Pos}_k + \textnormal{Neg}_k} \right),
\end{equation}

\noindent
where

\begin{equation}
\textnormal{Pos}_k = \sum_{j \neq i}^B h(x_i^k, x_j^k), \quad
\textnormal{Neg}_k = \sum_{{q \neq k}}^K \sum_{j=1}^B h(x_i^k, x_j^q).
\end{equation}

Here, $\textnormal{Pos}_k$ aggregates the similarities between $x_i^k$ and the augmented views of all other samples in the minibatch associated with the $k$-th transformation (the positive pairs). Conversely, $\textnormal{Neg}_k$ aggregates the similarities between $x_i^k$ and the augmented views of the samples in the minibatch generated by transformations $q \neq k$ (negative pairs). During the backward pass, this loss is used to update the parameters of the neural transformation module and the feature extraction module $\theta_\mathcal{T}, \theta_\phi$.

\subsubsection{Reconstruction Loss}

The proposed contrastive loss encourages that the latent representations of the augmented views meet the rules of disruption and diversity in the latent space generated by the encoder $\phi(\cdot)$. However, these representations may converge to a trivial solution as all augmented views associated with the same transformation may be mapped to a constant embedding. In this situation, the resulting representations of the augmented views lose all the information they obtained from the original sample, which violates the third rule about preservation proposed in Section \ref{sec:revisiting}.

To avoid this, our decoder reconstructs the original sample $x_i$ from each latent view $z_i^k$, such that ideally $\psi(z_i^k) = x_i$, for $1 \leq k \leq K$ (see Figure \ref{fig:reconstruction}). This is enforced by the reconstruction loss:

\begin{equation}\label{eq:reconstruction}
\ell_{\text{rec}}(x_i) = \dfrac{1}{K} \sum_{k=1}^K (x_i - \psi(z_i^k))^2.
\end{equation}

In the backward pass, this loss is used to update the parameters of the neural transformation module, the feature extraction module, and the reconstruction module, $\theta_\mathcal{T}, \theta_\phi, \theta_{\psi}$.

\subsubsection{Classification Loss}

Finally, the linear classification module $f(\cdot)$ is trained to perform the self-supervised classification (see Figure \ref{fig:classification}).
This is accomplished by minimizing the following cross-entropy loss for every sample \(x_i\) in the minibatch:

\begin{equation}\label{eq:classification}
    \ell_{\text{class}}(x_i) = - \frac{1}{K} \sum_{k=1}^{K} y_i^k \log\bigl(f(z_i^k)\bigr),
\end{equation}

\noindent where \(y_i^k\) is a one-hot vector indicating the true class of \(z_i^k\), and \(f(z_i^k)\) outputs its associated predicted probability distribution over the \(K\) classes. In the backward pass,  this objective is used to update the parameters of the neural transformation module, the feature extractor and the classification module $\theta_\mathcal{T}, \theta_\phi, \theta_f$.

\subsubsection{Multi-task learning in NeuCoReClass AD}

We train our model concurrently on the three proxy tasks described above, combining their loss functions in a multi-task learning setting. To prevent the optimization process from being dominated by losses with larger magnitudes, potentially leading to the neglect of others, we adopt the uncertainty-based weighting scheme proposed by \citet{liebel2018auxiliary}. This method automatically adjusts the contribution of each task based on its predictive uncertainty, promoting a more balanced and adaptive optimization across tasks.

To implement this scheme, we introduce a learnable uncertainty parameter for each loss: $\sigma_{\text{con}}$ for the contrastive loss, $\sigma_{\text{rec}}$ for reconstruction, and $\sigma_{\text{class}}$ for classification. These parameters are learned together with the model’s other parameters and determine the contribution of each loss to the overall learning objective. Accordingly, we optimize the following general loss for every sample $x_i$ in a minibatch:

\begin{equation}
\begin{split}
\ell(x_i) &= \frac{1}{2\sigma_{\text{con}}^2}\,\ell_{\text{con}}(x_i)
 + \frac{1}{2\sigma_{\text{rec}}^2}\,\ell_{\text{rec}}(x_i)
 + \frac{1}{2\sigma_{\text{class}}^2}\,\ell_{\text{class}}(x_i) \\
&\quad + \ln(1+\sigma_{\text{con}})
 + \ln(1+\sigma_{\text{rec}})
 + \ln(1+\sigma_{\text{class}}) \,.
\end{split}
\end{equation}

\subsection{Anomaly Detection in NeuCoReClass AD}\label{sec:detection}

During inference, NeuCoReClass AD assumes that the latent-space properties learned from normal training samples through the proxy tasks should generalize more consistently to unseen normal samples than to anomalous ones. Accordingly, the anomaly score of a new sample is computed from the contrastive, reconstruction, and classification proxy-task objectives.

Specifically, each proxy task provides an anomaly score. For a new sample $x_{\text{new}}$, the scores from the reconstruction $\textnormal{AS}_{\text{rec}}(x_{\text{new}})$ and classification $\textnormal{AS}_{\text{class}}(x_{\text{new}})$ tasks are the same as their respective losses (see Equations \ref{eq:reconstruction} and \ref{eq:classification}). However, the original contrastive loss relies on cross-sample relations within minibatches to learn coherent transformation-induced subspaces. Directly applying this formulation would make the anomaly score dependent on the composition of the evaluation minibatch and the augmented views of other samples. Since the anomaly score should ideally depend only on the evaluated sample and its augmented views, we define a modified contrastive anomaly score $\textnormal{AS}_{\text{con}}(x_{\text{new}})$ based exclusively on within-sample relations:

\begin{equation}
    \text{AS}_{\text{con}}(x_{\textnormal{new}}) =  - \dfrac{1}{K}\sum_{k=1}^{K}  \log \frac{1}{ 1 + \sum_{{q \neq k}}^K h(x_{\mathrm{new}\mathstrut}^{k\mathstrut},  x_{\mathrm{new}\mathstrut}^{q\mathstrut})}.
\end{equation}

Rather than reproducing the original contrastive loss exactly, this score measures whether the augmented views of the new sample preserve the diversity and disruption properties enforced for normal samples during training.

The overall anomaly score for a new sample, $\textnormal{AS}(x_{\text{new}})$, combines the three task-specific anomaly scores. We weight each score by the uncertainty parameters learned during training. Thus, the overall anomaly score for a new sample is computed as:

\begin{equation}
\label{eq:final_anomaly_score}
\begin{aligned}
\textnormal{AS}(x_{\text{new}}) 
    &= \frac{1}{2\sigma_{\text{con}}^2}\,\text{AS}_{\text{con}}(x_{\text{new}}) \\
    &\quad {}+ \frac{1}{2\sigma_{\text{rec}}^2}\,\text{AS}_{\text{rec}}(x_{\text{new}}) \\
    &\quad {}+ \frac{1}{2\sigma_{\text{class}}^2}\,\text{AS}_{\text{class}}(x_{\text{new}}).
\end{aligned}
\end{equation}

\section{Experimental Setup}\label{sec:experimentalsetup}

In this section, we describe the experimental setup used to evaluate the performance of NeuCoReClass AD. We assess the method across a diverse set of benchmark datasets and compare it against both shallow and deep anomaly detection approaches under different evaluation settings.

\subsection{Datasets}

For the experimentation, we consider datasets from the UCR Time Series Classification Archive \citep{dau2019ucr}, encompassing both univariate and multivariate time series from various domains. We adopt a one-class classification framework: for each dataset with $N$ classes, $n$ classes are designated as normal, and the remaining $N-n$ are treated as anomalies. Following prior work by \citet{qiu2021neural}, we evaluate performance under two settings: \textit{one-vs-rest}, where one class is considered normal and the remaining $(N-1)$ classes are treated as anomalies; and \textit{$(N-1)$-vs-rest}, where $(N-1)$ classes are considered normal and the remaining one is treated as anomalous. This latter scenario is more challenging, as it requires models to handle greater intra-class variability within the normal category.

Considering these two settings, we derive a total of $2N$ anomaly detection problems from each dataset with $N$ classes. Since the UCR archive contains a large number of benchmarks, we select the first 10 univariate and 10 multivariate datasets that satisfy the following criteria:

\begin{itemize}
    \item The dataset contains at most 10 classes.
    \item Each class includes at least 30 sequences.
    \item All sequences in the dataset are of equal length.
    \item No sequence contains missing values.
\end{itemize}

On this basis, we conduct experiments on a total of 144 TSAD problems. The characteristics of the datasets considered in our experimentation are presented in Appendix \ref{app:datasets}.

\subsection{Baselines for Time Series Anomaly Detection}

We evaluate NeuCoReClass AD by comparing it against a comprehensive set of baselines, ranging from classical shallow methods to large-scale transformer-based models. These are categorized into four main families based on their architectural nature and learning paradigm.

\textbf{Shallow Methods.} We select three popular shallow anomaly detection baselines widely used in the literature: One-Class SVM (OCSVM) \citep{scholkopf1999support}, which learns a compact decision boundary around the normal data; Isolation Forest (IF) \citep{liu2008isolation}, which detects anomalies by recursively partitioning the feature space and isolating outliers; and Local Outlier Factor (LOF), which flags samples with substantially lower local density than their neighbors \citep{breunig2000lof}.

\textbf{Standard Deep Learning Methods.} Among common deep baselines, we include an autoencoder (AutoAD) \citep{chen2018autoencoder}, which computes anomaly scores based on reconstruction errors; Deep SVDD, a deep variant of One-Class SVM that operates in the latent space of an autoencoder \citep{ruff2018deep}; and DAGMM, which combines a deep autoencoder with a Gaussian mixture model to estimate density in the latent space of the model \citep{zong2018deep}. 

\textbf{Self-Supervised Methods.} We consider deep self-supervised methods including NeuTraL AD \citep{qiu2021neural}, the first neural transformation learning approach for anomaly detection, and FixedTS, a self-supervised classification scheme for TSAD based on fixed transformations, inspired by \citet{wang2019effective}. We also include TimesURL, a self-supervised method based on contrastive pre-training for dataset-specific time series representation learning \citep{liu2024timesurl}.

\textbf{Large-Scale Transformer-Based Models.} We include two recent large-scale transformer-based methods: MOMENT, a pre-trained foundation model for time series \citep{goswami2024moment}, and TSINR, a GPT-2-based implicit neural representation framework \citep{li2025tsinr}. Both perform time series anomaly detection by reconstructing the input time series.

\subsection{Implementation and Experimental Details}

Here we describe the implementation choices and training protocols used for NeuCoReClass AD and all baseline methods.

\subsubsection{Implementation of Methods}

Shallow baselines are implemented using the scikit-learn library, adopting their predefined default hyperparameters without further tuning.

For NeuCoReClass AD and the remaining baselines except for MOMENT and TSINR, we adopt the encoder architecture proposed in TS2Vec \citep{yue2022ts2vec}, which is based on a dilated convolutional network for time series representation learning. For methods that incorporate reconstruction objectives (e.g., NeuCoReClass AD, DAGMM, and AutoAD), the decoder is implemented symmetrically to the encoder.

For NeuCoReClass AD and NeuTraL AD, learnable transformations are modeled following the architecture proposed in NeuTraL AD \citep{qiu2021neural}, which is based on convolutional neural networks.

For MOMENT, we use the official implementation provided by the authors, which is based on a transformer architecture pre-trained on large-scale time series data. Similarly, for TimesURL and TSINR, we adopt the official implementation released by the authors.

\subsubsection{Training and Validation Protocol}

For shallow models, the entire set of normal samples in the training set is used for model fitting. In the remaining methods, we split the normal samples into training (90\%) and validation (10\%) subsets. Models are trained on the training set for up to 10,000 epochs, and the validation set is used to monitor overfitting and adjust the learning rate.

Specifically, we employ a \texttt{ReduceLROnPlateau} scheduler that decreases the learning rate when the validation anomaly score plateaus (patience of 10 epochs), and apply early stopping when the learning rate drops below $10^{-6}$.

For TimesURL, we follow a two-stage procedure commonly adopted in the literature. First, the model is pre-trained on the training set to learn time series representations. Then, an OCSVM is fit on the representations of the training samples generated by the pre-trained encoder, following the protocol established in prior work \citep{pretrainocsvm}.

For MOMENT, we follow the protocol described in the original work. Specifically, we use the pre-trained foundation model and fine-tune the reconstruction head on the training data, adapting the model to the anomaly detection task while preserving the pre-trained representations, as suggested in the original paper that presents this method \citep{goswami2024moment}.

We use a batch size of 32 for all datasets, except for \textit{Motor Imagery} and \textit{Ethanol Concentration}, where we reduce it to 16 to accommodate the longer sequences and higher dimensionality of these datasets, which increases memory requirements.

\subsubsection{Additional Hyperparameters}

Certain hyperparameters specific to the considered deep methods must be fixed for all experiments. Below, we describe these hyperparameters along with the values we have chosen for them.

The key hyperparameter of DAGMM is the number of Gaussian components in its mixture-density network, that is, the number of GMM mixtures modeling the latent-space energy distribution. Based on \citet{zong2018deep}, we fix this to 3 components.

Following the methodology employed in the experimentation of NeuTraL AD \citep{qiu2021neural}, we set the fixed transformations in FixedTS to be all the possible combinations of flipping along the time axis (true/false), flipping along the channel axis (true/false), and shifting along the time axis by 0.25 of its time length (forward/backward/none). This leads to a total of 12 transformations. For consistency, we also use 12 transformations in NeuTraL AD and NeuCoReClass AD.

Finally, the temperature parameter for the contrastive loss in NeuTraL AD and NeuCoReClass AD is fixed at $0.1$, a standard choice in prior work in contrastive learning \citep{chen2020simple}.

\subsection{Evaluation Protocol}

For each problem considered, each method is trained exclusively on the normal samples from its corresponding training set and then used to compute anomaly scores on the unseen instances of the evaluation set.

Following prior work, model performance is assessed using two widely adopted metrics in anomaly detection: the area under the receiver operating characteristic curve (AUROC) and the area under the precision–recall curve (AUPR) \citep{huang2022efficient}. The AUROC evaluates the model’s ability to distinguish between normal and anomalous samples across varying thresholds, while the AUPR is particularly suited to imbalanced scenarios, summarizing the trade-off between precision and recall. High scores in both metrics indicate that the model effectively prioritizes anomalous instances over normal ones, while maintaining precision in its detections.

\section{Experimental Results}\label{sec:results}

In this section, we present the experimental evaluation of NeuCoReClass AD. Section \ref{sec:discussion} reports the quantitative performance of all methods across all the anomaly detection problems considered under both \textit{one-vs-rest} and \textit{(N-1)-vs-rest} settings. Section \ref{sec:illustration} illustrates how the generated augmented views satisfy the disruption, diversity, and preservation criteria. Finally, Section \ref{sec:characterization} shows how per-transformation contributions enable unsupervised characterization of distinct anomaly types.

\subsection{Results and Discussion}\label{sec:discussion}

We evaluate each method in each problem over five runs with different random seeds to mitigate random variability. Tables \ref{tab:onevsrest} and \ref{tab:n1vsrest} summarize the results under the one-vs-rest and (N-1)-vs-rest settings, respectively.

Each table summarizes  the mean AUROC and AUPR across all datasets for each method; the signed difference ($\Delta$) between each method’s mean and NeuCoReClass AD’s mean for each metric; and its average rank (lower is better). The values are computed by averaging first over the random seeds, and then across the problems. In the $\Delta$ columns, a negative value indicates the method performs worse than NeuCoReClass AD, while a positive value indicates it performs better. The best results in each column are underlined and bolded, and NeuCoReClass AD’s entries are shaded in dark gray whenever it is not the top-performing method. The complete results of each method in each problem considered for the experimentation are presented in Appendix \ref{app:results}. 

\begin{table*}[ht]
\centering
\scriptsize
\caption{Average AUROC and AUPR scores, $\Delta$ vs NeuCoReClass AD, and rankings over all problems under the \textit{one-vs-rest} setting.}
\label{tab:onevsrest}
{
\resizebox{\textwidth}{!}{%
  \begin{tabular}{l  c  c  c  c  c  c}
  \toprule
    Method & Mean AUROC (\%) & $\Delta$ AUROC (\%) & Rank & Mean AUPR (\%) & $\Delta$ AUPR (\%) & Rank \\
  \midrule
    IF & $66.32$ & $-6.89$ & $8.04$ & $78.53$ & $-2.89$ & $7.86$ \\
    LOF & $70.19$ & $-3.02$ & $6.34$ & $80.34$ & $-1.07$ & $6.43$ \\
    OCSVM & $70.63$ & $-2.59$ & $5.93$ & $80.32$ & $-1.10$ & $5.93$ \\
    AutoAD & $71.28$ & $-1.93$ & $6.03$ & $80.58$ & $-0.84$ & $6.03$ \\
    DAGMM & $53.94$ & $-19.27$ & $9.64$ & $77.56$ & $-3.86$ & $8.62$ \\
    DeepSVDD & $\underline{\mathbf{73.62}}$ & $\mathbf{+0.40}$ & $\underline{\mathbf{4.56}}$ & $\underline{\mathbf{81.65}}$ & $\mathbf{+0.23}$ & $\underline{\mathbf{4.69}}$ \\
    FixedTS & $66.59$ & $-6.63$ & $7.85$ & $78.38$ & $-3.03$ & $7.99$ \\
    NeuTraL AD & $70.20$ & $-3.01$ & $6.88$ & $80.04$ & $-1.38$ & $7.10$ \\
    TimesURL & $72.19$ & $-1.03$ & $6.01$ & $81.03$ & $-0.39$ & $6.08$ \\
    MOMENT & $67.61	$ & $-5.60$ & $7.17$ & $79.18$ & $-2.23$ & $7.24$ \\
    TSINR & $72.25$ & $-0.97$ & $4.83$ & $81.04$ & $-0.38$ & $5.05$ \\
    \textcolor{darkgray}{NeuCoReClass AD} & \textcolor{darkgray}{$73.22$} & \textcolor{darkgray}{$-$} & \textcolor{darkgray}{$4.74$} & \textcolor{darkgray}{$81.42$} & \textcolor{darkgray}{$-$} & \textcolor{darkgray}{$4.99$}\\
  \bottomrule
  \end{tabular}
}
}
\end{table*}

\begin{table*}[ht]
\centering
\scriptsize
\caption{Average AUROC and AUPR scores, $\Delta$ vs NeuCoReClass AD, and rankings over all problems under the \textit{(N-1)-vs-rest} setting.}
\label{tab:n1vsrest}
{
\resizebox{\textwidth}{!}{%
  \begin{tabular}{l  c  c  c  c  c  c}
  \toprule
    Method & Mean AUROC (\%) & $\Delta$ AUROC (\%) & Rank & Mean AUPR (\%) & $\Delta$ AUPR (\%) & Rank \\
  \midrule
    IF & $57.55$ & $-9.08$ & $7.47$ & $36.60$ & $-10.18$ & $7.78$ \\
    LOF & $62.62$ & $-4.01$ & $6.10$ & $45.26$ & $-1.52$ & $6.35$ \\
    OCSVM & $58.97$ & $-7.66$ & $7.06$ & $38.34$ & $-8.43$ & $7.09$ \\
    AutoAD & $59.87$ & $-6.76$ & $6.69$ & $37.91$ & $-8.86$ & $6.74$ \\
    DAGMM & $55.29$ & $-11.34$ & $7.94$ & $39.90$ & $-6.88$ & $5.74$ \\
    DeepSVDD & $60.41$ & $-6.22$ & $6.47$ & $38.22$ & $-8.56$ & $6.51$ \\
    FixedTS & $63.33$ & $-3.30$ & $5.90$ & $44.87$ & $-1.90$ & $6.25$ \\
    NeuTraL AD & $61.08$ & $-5.55$ & $6.36$ & $37.97$ & $-8.81$ & $6.35$ \\
    TimesURL & $60.64$ & $-5.99$ & $6.56$ & $37.65$ & $-9.12$ & $6.82$ \\
    MOMENT & $57.99$ & $-8.64$ & $7.15$ & $36.41$ & $-10.35$ & $7.49$ \\
    TSINR & $64.19$ & $-2.44$ &$5.43$ & $44.90$ & $-1.87$ & $5.69$ \\
    NeuCoReClass AD & $\underline{\mathbf{66.63}}$ & $-$ & $\underline{\mathbf{4.86}}$ & $\underline{\mathbf{46.77}}$ & $-$ & $\underline{\mathbf{5.21}}$ \\
  \bottomrule
  \end{tabular}
}
}
\end{table*}

Under the \textit{one-vs-rest} setting (see Table \ref{tab:onevsrest}), NeuCoReClass AD achieves a mean AUROC of 73.22\% and a mean AUPR of 81.42\%, placing it among the best-performing methods. In this setting, NeuCoReClass AD outperforms all considered baselines except DeepSVDD, which achieves slightly higher AUROC and AUPR values. Overall, the differences among the leading methods remain relatively small. This suggests that, when normality is defined by a single class, various approaches can effectively capture the underlying structure.

However, a markedly different behavior is observed under the more challenging \textit{(N-1)-vs-rest} setting (see Table \ref{tab:n1vsrest}), where the normal class is composed of multiple heterogeneous patterns. In this scenario, NeuCoReClass AD clearly outperforms all competing methods, achieving a mean AUROC of 66.63\% and a mean AUPR of 46.77\%. In particular, it surpasses both recent large-scale and self-supervised approaches such as MOMENT, TSINR and TimesURL, as well as strong deep baselines like DeepSVDD, with substantial margins (e.g., +6.22 points in AUROC and +8.56 points in AUPR over DeepSVDD, the strongest method in the one-vs-rest setting).

Overall, NeuCoReClass AD achieves a favorable balance across both evaluation settings, remaining generally competitive in simpler scenarios while providing substantial gains in more challenging and practically relevant conditions. Importantly, the \textit{(N-1)-vs-rest} setting better reflects real-world anomaly detection scenarios, where normal data typically comprises multiple variations rather than a single homogeneous pattern. In such cases, the proposed framework shows a stronger ability to capture this diversity, resulting in a more robust anomaly detection performance. Appendix~\ref{app:results} provides the complete results of the experiments for each problem and method. Appendix~\ref{app:statistical} provides a detailed statistical assessment of the results, showing ranking-based comparisons across methods.

In addition to detection performance, we analyze the model complexity and computational cost of the proposed method in Appendix~\ref{app:cost}. NeuCoReClass AD introduces a moderate increase in computational cost compared to simpler deep learning baselines, mainly due to the inclusion of learnable neural transformations. However, it exhibits much lower computational cost compared to recent large-scale methods such as MOMENT and TimesURL, while achieving superior detection performance across evaluation settings. Notably, these improvements are not solely explained by model size, as NeuCoReClass AD requires substantially fewer parameters than large-scale models such as MOMENT and TSINR, despite delivering stronger results. These findings highlight a favorable trade-off between effectiveness and computational efficiency, supporting its practical applicability.

Appendices~\ref{app:ablation} and~\ref{app:sensitivity} report the results of the ablation and sensitivity studies, respectively. These analyze the impact of the training objectives and the effect of the key hyperparameters on the performance of NeuCoReClass AD.

\subsection{Illustration of the Properties of the Augmented Views}\label{sec:illustration}

We analyze how jointly training on the proposed learning objectives encourages the augmented views to satisfy the principles of disruption, diversity, and preservation, as defined in Section~\ref{sec:revisiting}.

This analysis is conducted using the \textit{Epilepsy} dataset, which includes four human action recognition activities: \textit{sawing}, \textit{walking}, \textit{running} and \textit{epilepsy}. We train NeuCoReClass AD to characterize the \textit{sawing} class as the normal class, while the remaining three classes are treated as anomalous. The posterior analysis is conducted using all the samples in the evaluation set.

\subsubsection{Evidence of Disruption and Diversity}

The contrastive and classification objectives are designed to guide the model to organize the latent space into separated subspaces, each associated with a specific transformation. To assess whether this structure emerges in practice, we apply t-SNE \citep{van2008visualizing} to visualize the latent representations of the augmented views generated from both normal and anomalous samples in the evaluation set.

\begin{figure*}[ht]
    \centering
    \begin{subfigure}{0.32\textwidth}
        \centering
        \includegraphics[width=\linewidth]{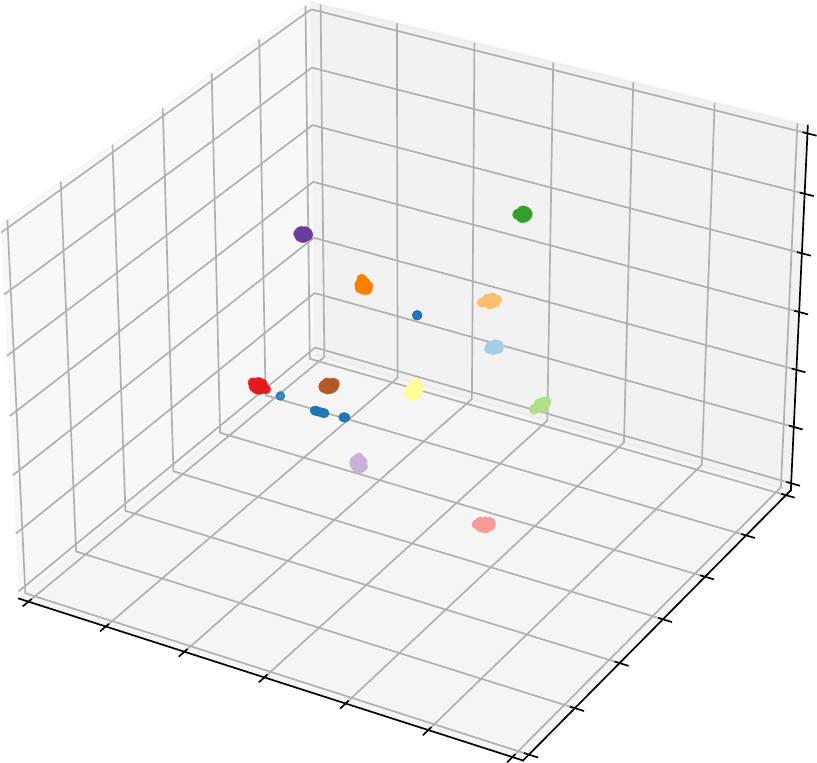}
        \caption{Latent representations of the augmented views of normal samples.}
        \label{fig:latent_normals}
    \end{subfigure}
    \hspace{2.5cm}  
    \begin{subfigure}{0.32\textwidth}
        \centering
        \includegraphics[width=\linewidth]{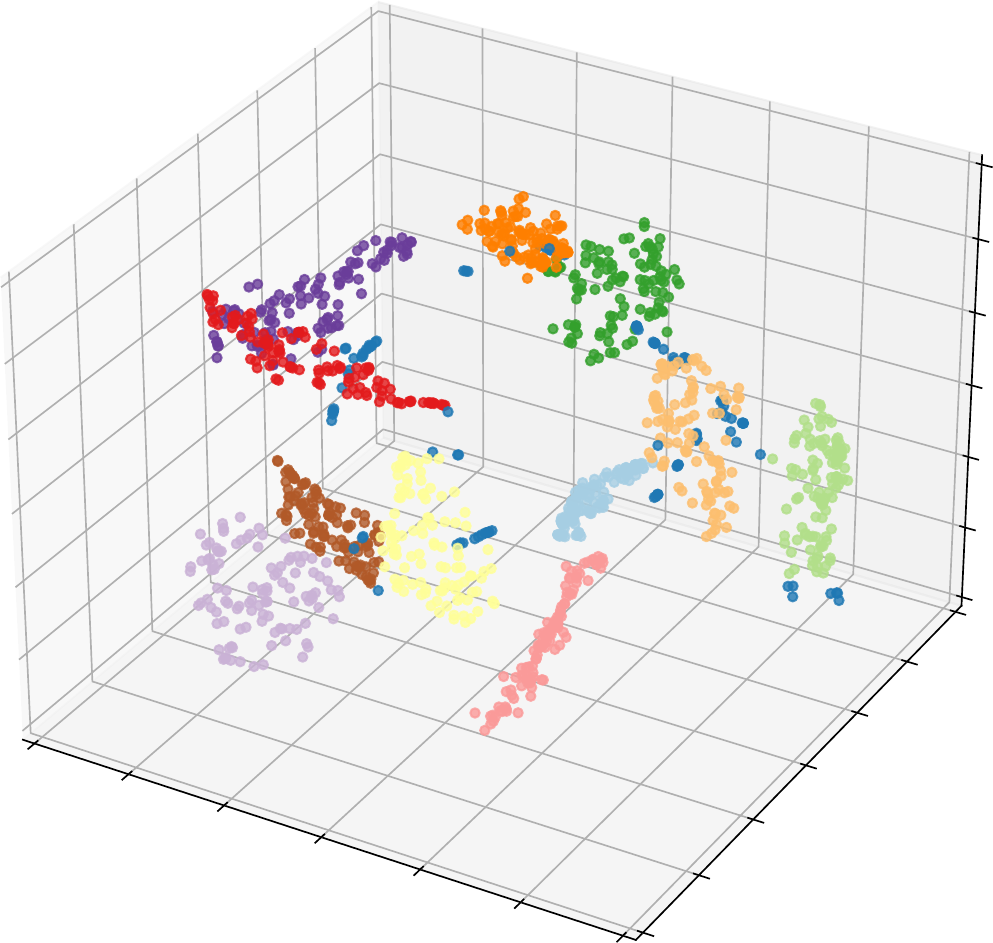}
        \caption{Latent representations of the augmented views of anomalous samples.}
        \label{fig:latent_anomalies}
    \end{subfigure}

    \caption{
        Visualization of the latent space induced by the joint training with reconstruction, contrastive, and classification objectives. Different colors represent different classes, where dark blue points correspond to the augmented views of the identity transformation.}
    \label{fig:latent_space_tsne}
\end{figure*}

As shown in Figure \ref{fig:latent_space_tsne}, the proxy tasks guide the encoder to produce a structured latent space in which the augmented views of normal samples form well-separated clusters, each corresponding to a different transformation-induced class. This behavior reflects the fulfillment of the diversity principle. In addition, the representations associated with the identity transformation (dark blue points) capture the characteristics of normality, as the proposed framework is trained exclusively using samples considered normal within the one-class learning setting. As these representations are separated from the remaining augmented views, this suggests that disruption is achieved at the latent level for the augmented views of normal samples. Conversely, the augmented anomalous samples result in latent representations that are more scattered and entangled, suggesting a failure to align with the learned transformation structure and thereby exposing their atypical nature.

\subsubsection{Evidence of Semantic Preservation}

To assess whether the augmented views retain meaningful information from their original time series, we analyze the model’s ability to reconstruct the original samples from their augmented representations. Specifically, we evaluate how well the decoder reconstructs unseen normal and anomalous samples by computing the reconstruction error for each input dimension. These errors are then visualized using box plots, allowing us to observe whether normal samples exhibit systematically lower reconstruction errors—a key indicator that semantic information is preserved through the transformations.

\begin{figure*}[t]
    \centering
    \begin{subfigure}[b]{0.42\linewidth}
        \centering
        \includegraphics[width=\linewidth]{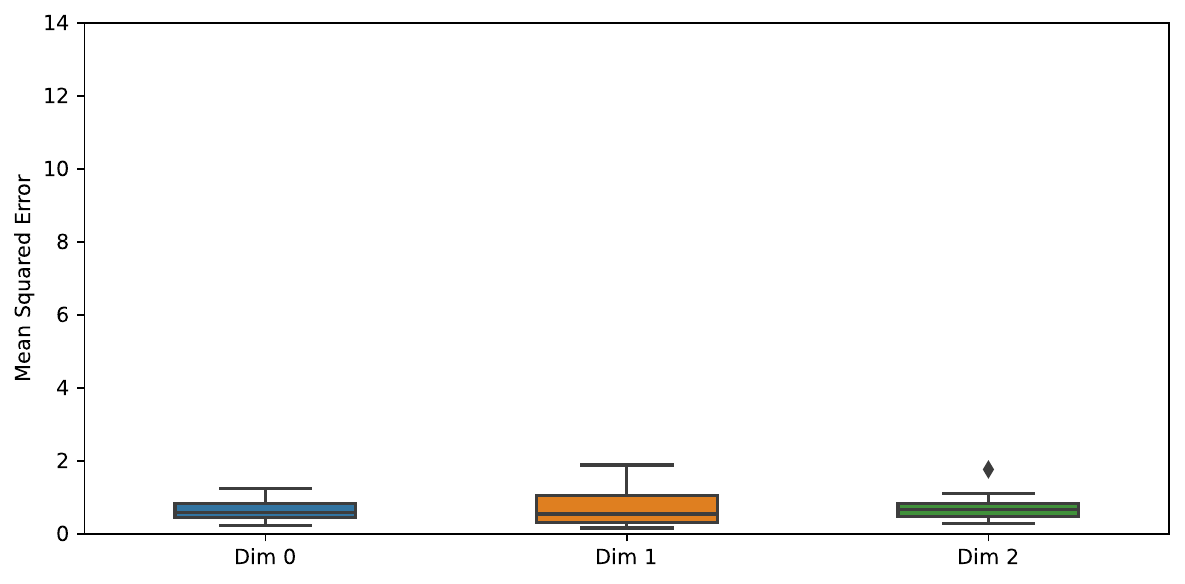}
        \caption{Reconstruction error per dimension for normal samples.}
        \label{fig:reconstruction_normal}
    \end{subfigure}
    \hfill
    \begin{subfigure}[b]{0.42\linewidth}
        \centering
        \includegraphics[width=\linewidth]{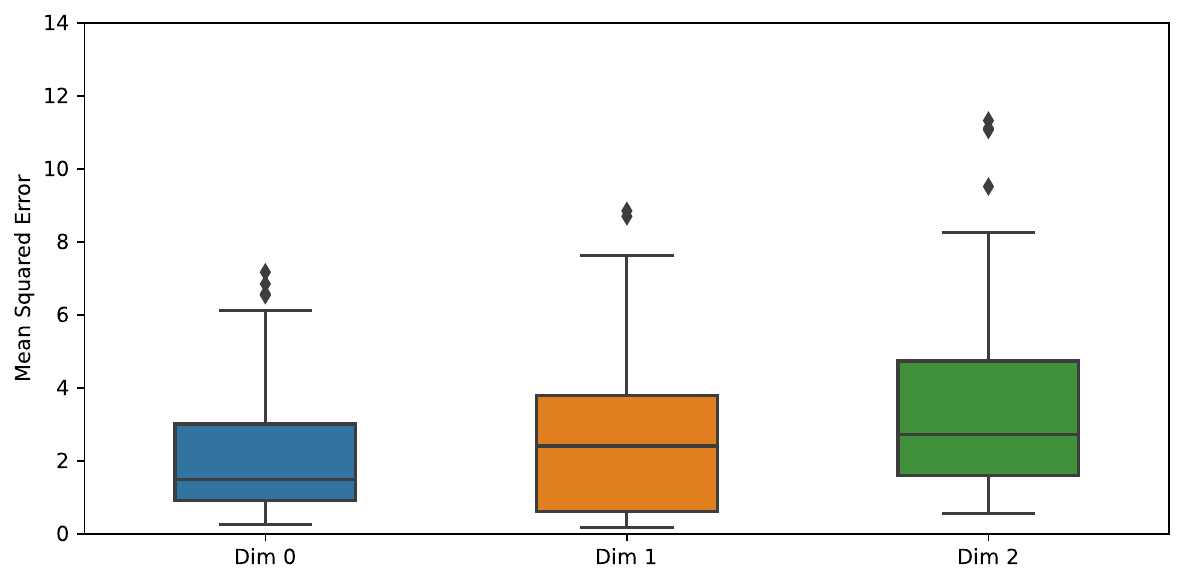}
        \caption{Reconstruction error per dimension for anomalous samples.}
        \label{fig:reconstruction_anomalies}
    \end{subfigure}
    \caption{Comparison of the reconstruction errors per input dimension between normal and anomalous test samples. Different colors represent different dimensions.}
    \label{fig:reconstruction_comparison}
\end{figure*}

As shown in Figure~\ref{fig:reconstruction_comparison}, the reconstruction error for normal samples is consistently low across all input dimensions, suggesting that the semantic content of the original time series is effectively preserved through the augmented views. In contrast, anomalous samples yield substantially higher reconstruction errors, reflecting their misalignment with the semantic structure learned from normal data.

\subsection{Characterization of Different Anomaly Types}\label{sec:characterization}

While the previous analysis focused on how NeuCoReClass AD distinguishes between normal and anomalous samples, a complementary question emerges: can the model also capture different types of anomalies based on the contribution of each transformation to the anomaly score? To explore this, we analyze the individual contribution of each transformation to the anomaly score for every sample in the evaluation set of the \textit{Epilepsy} dataset.

We compare the structure of the samples in the evaluation set in two different spaces: the original input space and the transformation-contribution space, where each sample is represented by a vector of $K$ elements, and each element in that vector denotes the anomaly score of the sample derived from each transformation. In both cases, we use t-SNE to project the high-dimensional data into two dimensions for visualization.

\begin{figure*}[ht]
    \centering
    \begin{subfigure}{0.45\textwidth}
        \centering
        \includegraphics[width=\linewidth]{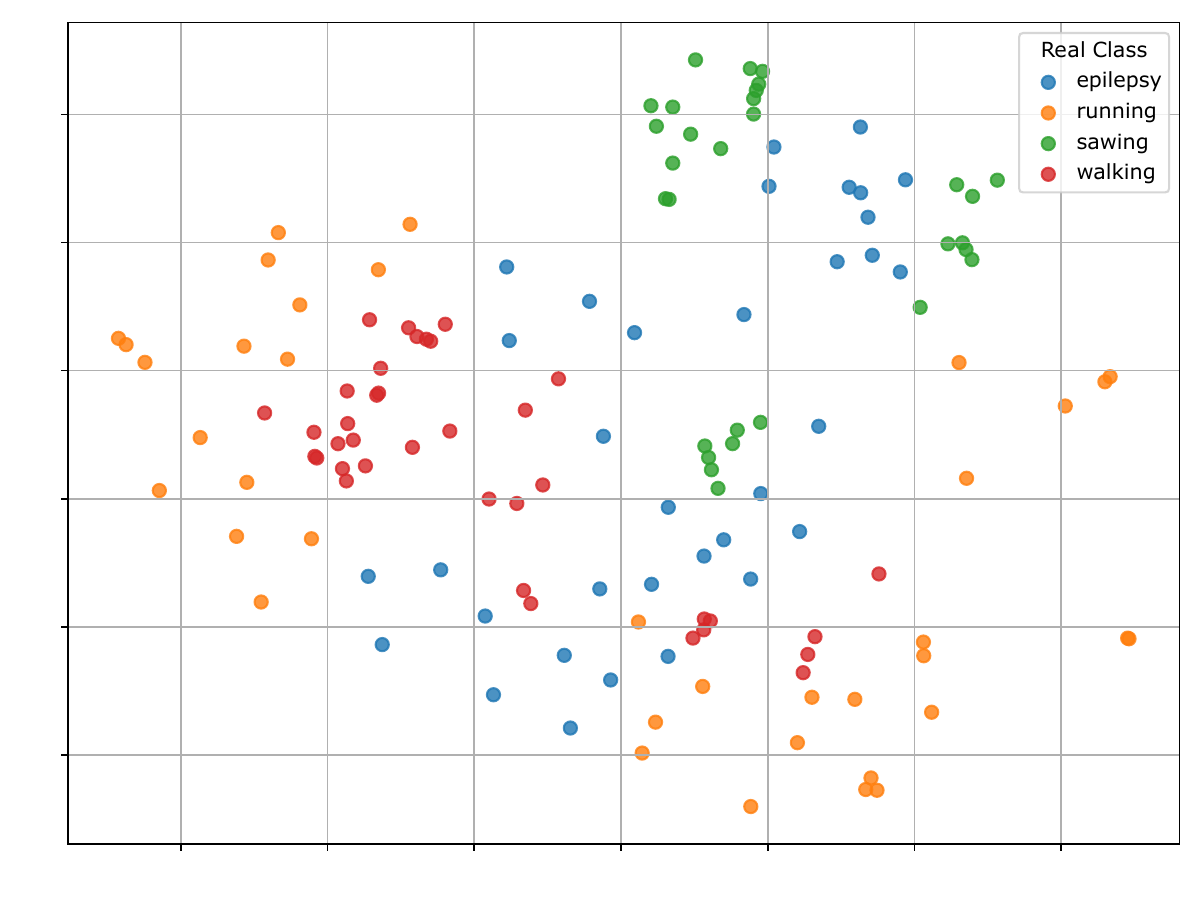}
        \caption{Dimensionality reduction over the original samples in the evaluation set.}
        \label{fig:original_space}
    \end{subfigure}
    \hfill
    \begin{subfigure}{0.45\textwidth}
        \centering
        \includegraphics[width=\linewidth]{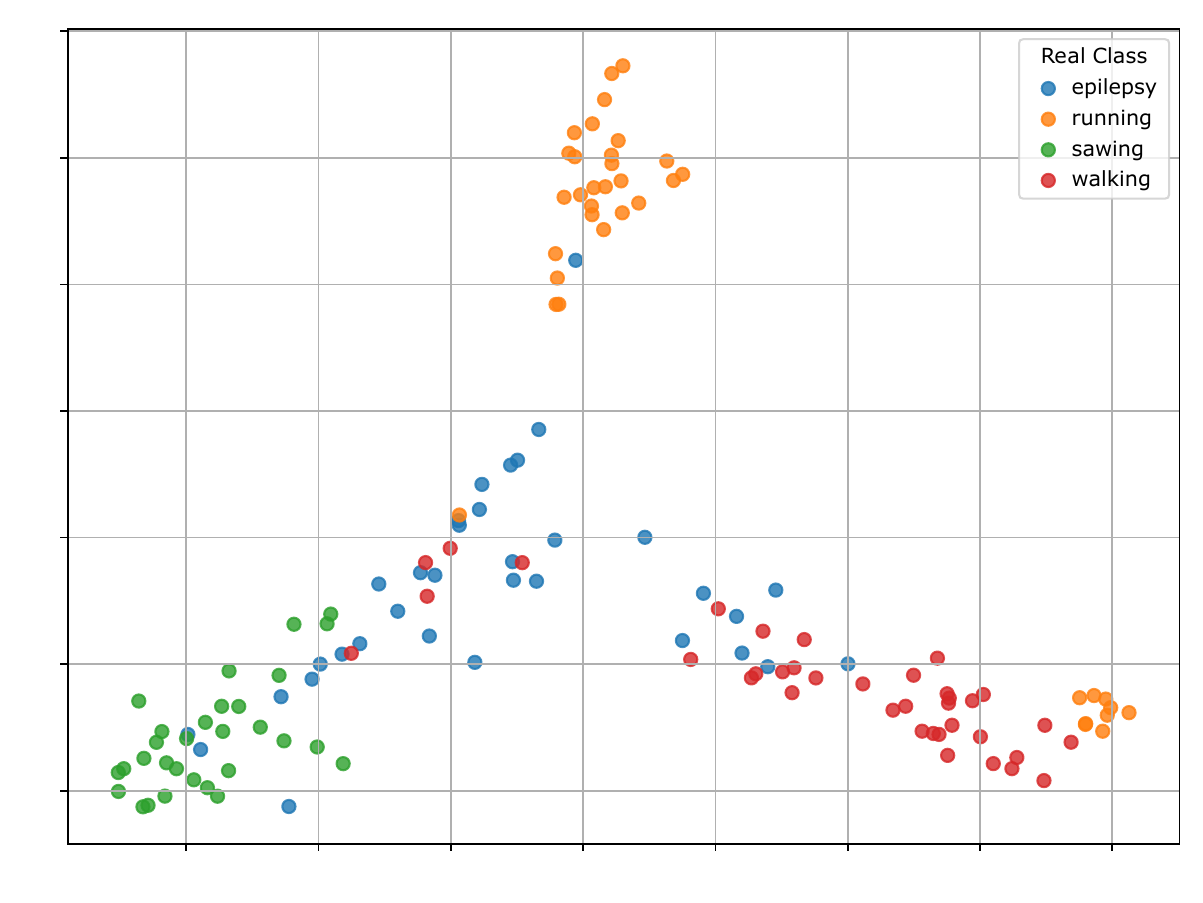}
        \caption{Dimensionality reduction over the transformation-wise anomaly contributions.}
        \label{fig:score_matrix_space}
    \end{subfigure}
    
    \caption{Class separability in the \textit{Epilepsy} dataset across different feature spaces. Each point corresponds to a sample from the evaluation set, and each color indicates one of the ground-truth classes.}
    \label{fig:score_matrix_vs_original}
\end{figure*}

As shown in Figure~\ref{fig:score_matrix_vs_original}, the projection based on the original features (Figure~\ref{fig:original_space}) exhibits no clear structure or clustering that corresponds to the true anomaly classes. In contrast, the projection based on the transformation-wise anomaly contributions (Figure~\ref{fig:score_matrix_space}) reveals well-separated clusters that align with the different types of anomalies. This highlights the capability of NeuCoReClass AD not only to detect anomalous behavior, but also to uncover structure within the anomaly space. Therefore, the contribution patterns of the transformations act as interpretable descriptors, enabling the model to differentiate between types of anomalies in a completely unsupervised manner.

\section{Conclusions and Future Work} \label{sec:conclusions}

This paper presents NeuCoReClass AD, a self-supervised time series anomaly detection method that integrates contrastive, reconstructive, and classification proxy tasks within a unified multi-task framework. This framework generates representations of augmented views that are informative, diverse, and semantically coherent for anomaly detection in time series. Across extensive experiments, NeuCoReClass AD demonstrates strong performance across both evaluation settings. In particular, under the \textit{one-vs-rest} setting, it outperforms all baselines except DeepSVDD, which achieves slightly better results, while showing clear advantages under the more demanding \textit{(N-1)-vs-rest} scenario. This latter setting better reflects real-world anomaly detection conditions, where normal data typically comprises multiple heterogeneous patterns rather than a single homogeneous class. The superior performance of NeuCoReClass AD in this scenario highlights its ability to effectively model diverse notions of normality, resulting in more robust and practically relevant anomaly detection performance.

Latent-space visualizations and reconstruction-error analyses confirm that the representations of augmented views fulfill the disruption, diversity, and preservation criteria, enabling both clear cluster separation and semantic retention. Additionally, analyzing transformation-specific anomaly scores reveals NeuCoReClass AD’s ability to identify and differentiate distinct types of anomalies in a fully unsupervised manner.

Nevertheless, despite the strong performance of the proposed approach across a wide range of datasets, various limitations should be acknowledged. First, the current transformation module relies on a single family of transformations (convolution-based), which may limit the diversity of the generated augmented views. This can lead to transformations capturing similar patterns, potentially restricting the ability to model heterogeneous anomaly profiles. Exploring hybrid transformation modules, such as attention- or frequency-based designs, could improve the diversity and expressiveness of the learned representations.

Second, the number of transformations is fixed a priori and shared across all datasets. However, the optimal number of transformations is likely to depend on the complexity and characteristics of each dataset. Developing fully unsupervised mechanisms to adapt this number dynamically could improve both efficiency and performance.

Third, although the proposed framework enforces disruption with respect to normality, the generated augmented views are not explicitly constrained to resemble realistic anomalies. As a result, disruption is primarily achieved at a structural level, without direct control over the semantic meaning of the generated variations. Incorporating mechanisms that leverage anomalies discovered during inference to progressively specialize transformations, potentially in an online manner, could enable the generation of more meaningful and representative anomaly patterns, thereby improving anomaly detection and characterization.

Finally, our experimental results indicate that no single method consistently outperforms all others across all datasets. In particular, classical or alternative approaches remain competitive in certain scenarios, especially when the underlying data structure can be effectively captured by simpler models. This observation aligns with the \textit{No Free Lunch Theorem} and highlights the need for a deeper understanding of the conditions under which different families of methods are most effective in time series anomaly detection.

\section*{Acknowledgments}

Authors thank financial support of the grant [PID2022-137442NB-I00] funded by the MICIU / AEI / 10.13039 / 501100011033, and the Basque Government [IT1504-22 and KK-2024/00030 from ELKARTEK Programme].  
A. S. F. thanks financial support of Departamento de Educación of the Basque Government under the grant PRE\_2022\_1\_0103. 
JA. L. thanks financial support of the Basque Government through the BERC 2022-2025 program and
by the Ministry of Science and Innovation: BCAM Severo Ochoa accreditation CEX2021-001142-S / MICIU / AEI / 10.13039 / 501100011033.

\bibliographystyle{cas-model2-names}

\bibliography{cas-refs}

\clearpage
\newpage
\appendix

\section{Datasets}\label{app:datasets}

We summarize the key characteristics of each dataset used in our experiments. Table \ref{tab:datasets} provides an overview of each dataset, including the number of classes, train/test sample counts, fixed sequence length, dimensionality, and data source. The source column denotes the origin or nature of the data. Note that the train size refers to the number of samples in the original training set of the time series classification problem. In each experiment, we only consider samples from the class(es) designated as normal for models' training.

\begin{table*}[h!]
\centering
\caption{Overview of the 20 UCR time series datasets used in the experiments.}
\begin{tabular}{lcccccl}
\toprule
Dataset & Classes & Train & Test & Length & Dim & Source \\
\midrule
ChlorineConcentration & 3 & 467 & 3840 & 166 & 1 & Simulated \\
Computers & 2 & 250 & 250 & 720 & 1 & Device \\
DistalPhalanxOutlineCorrect & 2 & 600 & 276 & 80 & 1 & Image \\
DistalPhalanxOutlineAgeGroup & 3 & 400 & 139 & 80 & 1 & Image \\
Earthquakes & 2 & 322 & 139 & 512 & 1 & Sensor \\
ECG200 & 2 & 100 & 100 & 96 & 1 & ECG \\
ElectricDevices & 7 & 8926 & 7711 & 96 & 1 & Device \\
EthanolLevel & 4 & 504 & 500 & 1751 & 1 & Spectro \\
FordA & 2 & 3601 & 1320 & 500 & 1 & Sensor \\
FordB & 2 & 3636 & 810 & 500 & 1 & Sensor \\
Epilepsy & 4 & 137 & 138 & 206 & 3 & HAR \\
EthanolConcentration & 4 & 261 & 263 & 1751 & 3 & Spectro \\
FaceDetection & 2 & 5890 & 3524 & 62 & 144 & EEG \\
FingerMovements & 2 & 316 & 100 & 50 & 28 & EEG \\
HandMovementDirection & 4 & 160 & 74 & 400 & 10 & EEG \\
Heartbeat & 2 & 204 & 205 & 405 & 61 & Audio \\
MotorImagery & 2 & 278 & 100 & 3000 & 64 & EEG \\
NATOPS & 6 & 180 & 180 & 51 & 24 & HAR \\
PEMS-SF & 7 & 267 & 173 & 144 & 963 & Other \\
PenDigits & 10 & 7494 & 3498 & 8 & 2 & Motion \\
\bottomrule
\end{tabular}
    \label{tab:datasets}
\end{table*}

\section{Complete Results}\label{app:results}

This appendix presents the complete results of the experiments conducted for each problem and method. We report the mean AUROC and AUPR performance scores for all problems considered under both evaluation settings.

Specifically, we include four tables: (i) Table~\ref{tab:datasets_roc_rev0} shows the mean AUROC scores under the \textit{one-vs-rest} setting; (ii) Table~\ref{tab:datasets_pr_rev0} presents the corresponding AUPR results for the same setting; (iii) Table~\ref{tab:datasets_roc_rev1} displays the AUROC scores under the \textit{(N-1)-vs-rest} setting; and (iv) Table~\ref{tab:datasets_pr_rev1} contains the corresponding AUPR results. Each value represents the average over five independent runs with different seeds. These detailed results complement the summary statistics reported in the main paper and provide a comprehensive view of the performance of each method across all problems.

\begin{table*}[h!]
\centering
\caption{Mean AUROC scores and rankings for all problems under the \textit{one-vs-rest} setting.}
\label{tab:datasets_roc_rev0}
{

\resizebox{\linewidth}{!}{%
  \begin{tabular}{lcccccccccccc}
  \toprule
    Dataset\_Normality & IF & LOF & OCSVM & AutoAD & DAGMM & DeepSVDD & FixedTS & NeuTraL AD & TimesURL & MOMENT & TSINR & NeuCoReClass AD \\
    \midrule
    ChlorineConcentration\_0 & $41.10\,(10)$ & $38.47\,(11)$ & $36.76\,(12)$ & $41.75\,(8)$ & $50.83\,(4)$ & $41.22\,(9)$ & $46.07\,(6)$ & $52.38\,(3)$ & $46.77\,(5)$ & $44.80\,(7)$ & $\underline{\mathbf{69.91\,(1)}}$ & $58.35\,(2)$ \\
    ChlorineConcentration\_1 & $53.26\,(9)$ & $52.17\,(11)$ & $56.91\,(5)$ & $54.88\,(7)$ & $53.74\,(8)$ & $52.33\,(10)$ & $51.64\,(12)$ & $59.74\,(3)$ & $55.11\,(6)$ & $58.27\,(4)$ & $\underline{\mathbf{70.24\,(1)}}$ & $66.92\,(2)$ \\
    ChlorineConcentration\_2 & $60.04\,(8)$ & $56.34\,(12)$ & $66.39\,(3)$ & $58.98\,(9)$ & $56.78\,(11)$ & $61.78\,(7)$ & $57.07\,(10)$ & $61.87\,(6)$ & $62.60\,(5)$ & $64.89\,(4)$ & $\underline{\mathbf{81.91\,(1)}}$ & $68.54\,(2)$ \\
    Computers\_0 & $30.28\,(12)$ & $62.00\,(2)$ & $54.21\,(8)$ & $\underline{\mathbf{63.26\,(1)}}$ & $45.52\,(10)$ & $50.42\,(9)$ & $35.40\,(11)$ & $57.24\,(6)$ & $56.34\,(7)$ & $61.32\,(4)$ & $58.14\,(5)$ & $61.91\,(3)$ \\
    Computers\_1 & $\underline{\mathbf{69.17\,(1)}}$ & $49.56\,(9)$ & $54.59\,(5)$ & $52.17\,(8)$ & $48.79\,(11)$ & $55.90\,(4)$ & $61.43\,(2)$ & $59.27\,(3)$ & $54.58\,(6)$ & $49.34\,(10)$ & $47.48\,(12)$ & $54.41\,(7)$ \\
    DistalPhalanxOutlineAgeGroup\_0 & $68.32\,(9)$ & $\underline{\mathbf{79.09\,(1)}}$ & $73.39\,(7)$ & $76.44\,(4)$ & $75.97\,(5)$ & $78.86\,(2)$ & $49.24\,(12)$ & $71.29\,(8)$ & $77.25\,(3)$ & $65.33\,(11)$ & $67.55\,(10)$ & $75.02\,(6)$ \\
    DistalPhalanxOutlineAgeGroup\_1 & $\underline{\mathbf{75.57\,(1)}}$ & $74.92\,(2)$ & $73.79\,(3)$ & $73.68\,(4)$ & $31.10\,(11)$ & $61.01\,(6)$ & $26.44\,(12)$ & $34.34\,(10)$ & $73.39\,(5)$ & $41.10\,(8)$ & $38.18\,(9)$ & $41.12\,(7)$ \\
    DistalPhalanxOutlineAgeGroup\_2 & $81.45\,(5)$ & $68.48\,(11)$ & $79.63\,(8)$ & $78.20\,(9)$ & $53.59\,(12)$ & $82.17\,(4)$ & $80.13\,(7)$ & $73.95\,(10)$ & $80.30\,(6)$ & $82.60\,(3)$ & $\underline{\mathbf{83.78\,(1)}}$ & $83.56\,(2)$ \\
    DistalPhalanxOutlineCorrect\_0 & $50.29\,(3)$ & $40.74\,(11)$ & $46.97\,(5)$ & $52.48\,(2)$ & $46.25\,(8)$ & $47.87\,(4)$ & $32.81\,(12)$ & $46.83\,(6)$ & $\underline{\mathbf{54.16\,(1)}}$ & $46.60\,(7)$ & $45.51\,(9)$ & $43.29\,(10)$ \\
    DistalPhalanxOutlineCorrect\_1 & $71.85\,(6)$ & $\underline{\mathbf{83.08\,(1)}}$ & $71.96\,(5)$ & $67.27\,(9)$ & $48.97\,(12)$ & $70.80\,(7)$ & $72.27\,(3)$ & $73.73\,(2)$ & $65.95\,(11)$ & $67.27\,(10)$ & $67.85\,(8)$ & $72.11\,(4)$ \\
    ECG200\_0 & $52.16\,(8)$ & $22.68\,(12)$ & $56.51\,(4)$ & $51.39\,(9)$ & $42.77\,(11)$ & $54.02\,(6)$ & $59.67\,(2)$ & $52.40\,(7)$ & $44.52\,(10)$ & $\underline{\mathbf{66.28\,(1)}}$ & $57.10\,(3)$ & $56.48\,(5)$ \\
    ECG200\_1 & $86.80\,(7)$ & $89.11\,(4)$ & $87.41\,(6)$ & $85.11\,(8)$ & $89.68\,(2)$ & $\underline{\mathbf{89.91\,(1)}}$ & $68.63\,(12)$ & $73.63\,(10)$ & $83.02\,(9)$ & $72.45\,(11)$ & $89.37\,(3)$ & $87.53\,(5)$ \\
    Earthquakes\_0 & $65.55\,(4)$ & $\underline{\mathbf{66.18\,(1)}}$ & $64.59\,(5)$ & $41.52\,(11)$ & $50.39\,(8)$ & $47.71\,(10)$ & $65.96\,(2)$ & $48.27\,(9)$ & $52.35\,(7)$ & $40.97\,(12)$ & $64.05\,(6)$ & $65.55\,(3)$ \\
    Earthquakes\_1 & $35.95\,(12)$ & $58.26\,(8)$ & $53.93\,(10)$ & $67.78\,(5)$ & $61.86\,(6)$ & $70.71\,(3)$ & $48.33\,(11)$ & $68.17\,(4)$ & $71.33\,(2)$ & $\underline{\mathbf{73.16\,(1)}}$ & $55.00\,(9)$ & $61.31\,(7)$ \\
    ElectricDevices\_0 & $\underline{\mathbf{75.95\,(1)}}$ & $67.06\,(10)$ & $72.03\,(4)$ & $70.61\,(5)$ & $41.90\,(12)$ & $69.55\,(6)$ & $73.52\,(2)$ & $66.72\,(11)$ & $72.23\,(3)$ & $67.46\,(9)$ & $68.79\,(7)$ & $67.51\,(8)$ \\
    ElectricDevices\_1 & $13.36\,(12)$ & $83.67\,(8)$ & $37.90\,(11)$ & $84.49\,(7)$ & $62.37\,(10)$ & $\underline{\mathbf{96.34\,(1)}}$ & $76.71\,(9)$ & $93.61\,(3)$ & $93.62\,(2)$ & $90.09\,(5)$ & $88.94\,(6)$ & $93.51\,(4)$ \\
    ElectricDevices\_2 & $75.05\,(10)$ & $74.24\,(11)$ & $78.88\,(8)$ & $81.39\,(7)$ & $52.21\,(12)$ & $88.63\,(4)$ & $81.73\,(6)$ & $\underline{\mathbf{93.49\,(1)}}$ & $89.61\,(3)$ & $83.09\,(5)$ & $76.79\,(9)$ & $90.20\,(2)$ \\
    ElectricDevices\_3 & $\underline{\mathbf{89.05\,(1)}}$ & $85.08\,(7)$ & $85.31\,(6)$ & $70.86\,(11)$ & $47.19\,(12)$ & $89.04\,(2)$ & $82.66\,(10)$ & $83.62\,(8)$ & $82.74\,(9)$ & $86.72\,(4)$ & $86.71\,(5)$ & $89.04\,(3)$ \\
    ElectricDevices\_4 & $48.16\,(12)$ & $67.99\,(7)$ & $65.77\,(8)$ & $\underline{\mathbf{80.69\,(1)}}$ & $48.43\,(11)$ & $71.18\,(5)$ & $55.23\,(10)$ & $61.91\,(9)$ & $70.99\,(6)$ & $71.85\,(4)$ & $75.28\,(2)$ & $73.13\,(3)$ \\
    ElectricDevices\_5 & $83.61\,(5)$ & $78.53\,(11)$ & $81.29\,(8)$ & $80.58\,(9)$ & $64.41\,(12)$ & $\underline{\mathbf{86.65\,(1)}}$ & $84.38\,(3)$ & $85.76\,(2)$ & $84.13\,(4)$ & $83.14\,(7)$ & $79.94\,(10)$ & $83.57\,(6)$ \\
    ElectricDevices\_6 & $44.86\,(12)$ & $69.78\,(3)$ & $62.65\,(7)$ & $\underline{\mathbf{77.30\,(1)}}$ & $55.61\,(10)$ & $60.64\,(9)$ & $48.62\,(11)$ & $66.84\,(4)$ & $69.85\,(2)$ & $65.78\,(5)$ & $64.38\,(6)$ & $61.45\,(8)$ \\
    Epilepsy\_0 & $75.53\,(10)$ & $90.38\,(6)$ & $87.19\,(7)$ & $93.48\,(4)$ & $91.83\,(5)$ & $\underline{\mathbf{97.43\,(1)}}$ & $61.29\,(12)$ & $81.62\,(8)$ & $94.80\,(2)$ & $67.07\,(11)$ & $76.28\,(9)$ & $93.84\,(3)$ \\
    Epilepsy\_1 & $21.70\,(11)$ & $25.21\,(9)$ & $20.63\,(12)$ & $23.69\,(10)$ & $88.63\,(3)$ & $\underline{\mathbf{92.59\,(1)}}$ & $80.98\,(5)$ & $89.99\,(2)$ & $81.11\,(4)$ & $29.44\,(7)$ & $27.54\,(8)$ & $74.71\,(6)$ \\
    Epilepsy\_2 & $75.96\,(11)$ & $84.89\,(10)$ & $92.35\,(7)$ & $96.14\,(5)$ & $94.79\,(6)$ & $\underline{\mathbf{99.22\,(1)}}$ & $69.68\,(12)$ & $99.02\,(3)$ & $98.02\,(4)$ & $89.27\,(9)$ & $90.42\,(8)$ & $99.09\,(2)$ \\
    Epilepsy\_3 & $98.07\,(6)$ & $98.58\,(4)$ & $98.77\,(3)$ & $99.03\,(2)$ & $50.00\,(12)$ & $\underline{\mathbf{99.56\,(1)}}$ & $77.80\,(11)$ & $85.70\,(10)$ & $97.73\,(7)$ & $95.86\,(9)$ & $97.62\,(8)$ & $98.58\,(5)$ \\
    EthanolConcentration\_0 & $61.98\,(6)$ & $62.66\,(5)$ & $65.31\,(2)$ & $63.85\,(3)$ & $54.54\,(10)$ & $60.02\,(8)$ & $48.64\,(11)$ & $45.00\,(12)$ & $62.75\,(4)$ & $\underline{\mathbf{66.71\,(1)}}$ & $61.62\,(7)$ & $58.60\,(9)$ \\
    EthanolConcentration\_1 & $61.26\,(2)$ & $\underline{\mathbf{62.04\,(1)}}$ & $58.54\,(4)$ & $52.01\,(8)$ & $49.95\,(10)$ & $53.05\,(7)$ & $50.09\,(9)$ & $45.82\,(12)$ & $47.89\,(11)$ & $59.93\,(3)$ & $56.25\,(5)$ & $55.32\,(6)$ \\
    EthanolConcentration\_2 & $52.10\,(7)$ & $53.50\,(4)$ & $52.27\,(6)$ & $53.63\,(3)$ & $48.49\,(11)$ & $51.13\,(8)$ & $48.47\,(12)$ & $50.23\,(9)$ & $54.98\,(2)$ & $\underline{\mathbf{55.14\,(1)}}$ & $53.21\,(5)$ & $49.94\,(10)$ \\
    EthanolConcentration\_3 & $42.10\,(11)$ & $48.35\,(8)$ & $41.20\,(12)$ & $43.11\,(10)$ & $49.67\,(6)$ & $48.49\,(7)$ & $51.96\,(5)$ & $\underline{\mathbf{56.71\,(1)}}$ & $52.52\,(3)$ & $45.65\,(9)$ & $53.60\,(2)$ & $52.02\,(4)$ \\
    EthanolLevel\_0 & $48.47\,(11)$ & $49.04\,(10)$ & $51.98\,(7)$ & $54.13\,(5)$ & $49.34\,(9)$ & $45.40\,(12)$ & $\underline{\mathbf{66.90\,(1)}}$ & $57.94\,(4)$ & $53.26\,(6)$ & $51.03\,(8)$ & $64.28\,(2)$ & $59.77\,(3)$ \\
    EthanolLevel\_1 & $56.11\,(6)$ & $53.39\,(9)$ & $56.22\,(5)$ & $52.58\,(10)$ & $51.28\,(11)$ & $56.37\,(4)$ & $\underline{\mathbf{60.82\,(1)}}$ & $53.53\,(8)$ & $50.74\,(12)$ & $54.37\,(7)$ & $58.22\,(3)$ & $59.66\,(2)$ \\
    EthanolLevel\_2 & $56.13\,(4)$ & $54.94\,(7)$ & $56.25\,(3)$ & $56.00\,(5)$ & $50.05\,(11)$ & $51.06\,(8)$ & $50.26\,(10)$ & $49.01\,(12)$ & $50.62\,(9)$ & $57.03\,(2)$ & $\underline{\mathbf{57.81\,(1)}}$ & $55.91\,(6)$ \\
    EthanolLevel\_3 & $52.91\,(9)$ & $52.80\,(10)$ & $55.39\,(4)$ & $54.14\,(5)$ & $48.06\,(12)$ & $52.98\,(8)$ & $56.77\,(2)$ & $53.69\,(6)$ & $48.88\,(11)$ & $55.73\,(3)$ & $\underline{\mathbf{66.70\,(1)}}$ & $53.62\,(7)$ \\
    FaceDetection\_0 & $51.64\,(5)$ & $\underline{\mathbf{52.25\,(1)}}$ & $52.08\,(3)$ & $50.90\,(9)$ & $49.44\,(12)$ & $51.30\,(6)$ & $50.31\,(11)$ & $50.64\,(10)$ & $50.98\,(7)$ & $50.91\,(8)$ & $52.21\,(2)$ & $52.04\,(4)$ \\
    FaceDetection\_1 & $48.91\,(7)$ & $48.38\,(11)$ & $48.67\,(9)$ & $49.33\,(5)$ & $49.82\,(2)$ & $48.05\,(12)$ & $49.43\,(4)$ & $\underline{\mathbf{49.84\,(1)}}$ & $49.45\,(3)$ & $49.17\,(6)$ & $48.45\,(10)$ & $48.70\,(8)$ \\
    FingerMovements\_0 & $62.67\,(2)$ & $\underline{\mathbf{63.63\,(1)}}$ & $61.78\,(4)$ & $61.37\,(5)$ & $54.79\,(10)$ & $57.72\,(8)$ & $45.30\,(12)$ & $46.06\,(11)$ & $59.40\,(7)$ & $60.55\,(6)$ & $61.79\,(3)$ & $57.04\,(9)$ \\
    FingerMovements\_1 & $38.70\,(10)$ & $39.98\,(8)$ & $37.49\,(12)$ & $41.55\,(6)$ & $\underline{\mathbf{50.50\,(1)}}$ & $37.71\,(11)$ & $49.62\,(2)$ & $48.27\,(3)$ & $39.24\,(9)$ & $40.58\,(7)$ & $43.64\,(4)$ & $41.89\,(5)$ \\
    FordA\_0 & $52.11\,(5)$ & $56.96\,(3)$ & $56.99\,(2)$ & $38.31\,(12)$ & $51.34\,(6)$ & $\underline{\mathbf{62.05\,(1)}}$ & $46.31\,(8)$ & $52.71\,(4)$ & $44.15\,(9)$ & $43.82\,(10)$ & $49.79\,(7)$ & $40.37\,(11)$ \\
    FordA\_1 & $52.80\,(11)$ & $59.80\,(9)$ & $58.02\,(10)$ & $69.71\,(6)$ & $49.18\,(12)$ & $74.64\,(2)$ & $73.51\,(3)$ & $71.52\,(4)$ & $\underline{\mathbf{77.42\,(1)}}$ & $68.70\,(7)$ & $65.97\,(8)$ & $70.85\,(5)$ \\
    FordB\_0 & $50.31\,(5)$ & $\underline{\mathbf{58.66\,(1)}}$ & $57.57\,(3)$ & $44.78\,(11)$ & $49.43\,(6)$ & $50.83\,(4)$ & $47.74\,(8)$ & $47.27\,(9)$ & $40.28\,(12)$ & $49.11\,(7)$ & $57.91\,(2)$ & $47.00\,(10)$ \\
    FordB\_1 & $49.65\,(11)$ & $52.41\,(10)$ & $54.38\,(8)$ & $62.27\,(6)$ & $48.87\,(12)$ & $64.09\,(4)$ & $64.41\,(3)$ & $66.97\,(2)$ & $\underline{\mathbf{72.02\,(1)}}$ & $60.25\,(7)$ & $53.33\,(9)$ & $63.31\,(5)$ \\
    HandMovementDirection\_0 & $53.76\,(7)$ & $49.04\,(10)$ & $53.67\,(8)$ & $\underline{\mathbf{60.75\,(1)}}$ & $48.97\,(11)$ & $57.60\,(3)$ & $46.42\,(12)$ & $54.12\,(6)$ & $58.80\,(2)$ & $55.93\,(4)$ & $52.34\,(9)$ & $55.75\,(5)$ \\
    HandMovementDirection\_1 & $57.56\,(9)$ & $63.86\,(3)$ & $62.12\,(5)$ & $\underline{\mathbf{68.76\,(1)}}$ & $50.21\,(12)$ & $61.58\,(6)$ & $53.95\,(10)$ & $53.83\,(11)$ & $61.30\,(7)$ & $58.20\,(8)$ & $62.94\,(4)$ & $66.70\,(2)$ \\
    HandMovementDirection\_2 & $46.13\,(10)$ & $50.17\,(4)$ & $49.60\,(5)$ & $46.98\,(8)$ & $48.14\,(7)$ & $\underline{\mathbf{56.20\,(1)}}$ & $53.22\,(2)$ & $49.20\,(6)$ & $44.04\,(11)$ & $40.68\,(12)$ & $50.17\,(4)$ & $46.73\,(9)$ \\
    HandMovementDirection\_3 & $48.12\,(8)$ & $48.57\,(6)$ & $45.71\,(12)$ & $50.00\,(3)$ & $48.38\,(7)$ & $\underline{\mathbf{58.36\,(1)}}$ & $45.98\,(11)$ & $48.98\,(5)$ & $54.93\,(2)$ & $46.05\,(9)$ & $46.05\,(10)$ & $49.93\,(4)$ \\
    Heartbeat\_0 & $63.25\,(6)$ & $64.91\,(2)$ & $\underline{\mathbf{65.55\,(1)}}$ & $60.84\,(7)$ & $44.44\,(12)$ & $57.49\,(9)$ & $51.47\,(11)$ & $53.99\,(10)$ & $64.90\,(3)$ & $64.02\,(5)$ & $64.35\,(4)$ & $60.36\,(8)$ \\
    Heartbeat\_1 & $39.67\,(7)$ & $37.34\,(11)$ & $36.96\,(12)$ & $40.34\,(6)$ & $52.60\,(2)$ & $52.15\,(3)$ & $\underline{\mathbf{57.97\,(1)}}$ & $47.99\,(4)$ & $38.13\,(10)$ & $38.21\,(9)$ & $39.11\,(8)$ & $41.66\,(5)$ \\
    MotorImagery\_0 & $40.65\,(10)$ & $41.76\,(8)$ & $40.04\,(11)$ & $43.50\,(6)$ & $50.00\,(3)$ & $43.50\,(5)$ & $54.77\,(2)$ & $\underline{\mathbf{55.83\,(1)}}$ & $44.81\,(4)$ & $40.83\,(9)$ & $42.92\,(7)$ & $39.18\,(12)$ \\
    MotorImagery\_1 & $58.16\,(6)$ & $58.16\,(7)$ & $60.70\,(2)$ & $58.40\,(5)$ & $52.10\,(10)$ & $57.07\,(9)$ & $44.18\,(11)$ & $44.02\,(12)$ & $57.09\,(8)$ & $59.74\,(3)$ & $58.63\,(4)$ & $\underline{\mathbf{60.90\,(1)}}$ \\
    NATOPS\_0 & $76.77\,(10)$ & $77.29\,(8)$ & $79.93\,(6)$ & $82.29\,(2)$ & $64.44\,(12)$ & $\underline{\mathbf{82.44\,(1)}}$ & $74.27\,(11)$ & $81.09\,(4)$ & $80.90\,(5)$ & $76.79\,(9)$ & $79.01\,(7)$ & $81.46\,(3)$ \\
    NATOPS\_1 & $80.24\,(10)$ & $81.83\,(9)$ & $84.00\,(6)$ & $83.35\,(7)$ & $67.83\,(11)$ & $85.46\,(2)$ & $60.38\,(12)$ & $84.42\,(5)$ & $\underline{\mathbf{85.65\,(1)}}$ & $82.49\,(8)$ & $85.38\,(3)$ & $84.77\,(4)$ \\
    NATOPS\_2 & $84.15\,(10)$ & $88.57\,(7)$ & $88.36\,(8)$ & $89.80\,(3)$ & $68.14\,(11)$ & $89.31\,(5)$ & $49.92\,(12)$ & $85.37\,(9)$ & $\underline{\mathbf{90.41\,(1)}}$ & $89.01\,(6)$ & $90.35\,(2)$ & $89.66\,(4)$ \\
    NATOPS\_3 & $85.24\,(9)$ & $87.69\,(7)$ & $92.31\,(4)$ & $95.02\,(3)$ & $74.74\,(11)$ & $96.67\,(2)$ & $80.75\,(10)$ & $88.12\,(6)$ & $87.43\,(8)$ & $33.67\,(12)$ & $88.68\,(5)$ & $\underline{\mathbf{97.52\,(1)}}$ \\
    NATOPS\_4 & $85.07\,(11)$ & $91.76\,(8)$ & $98.24\,(2)$ & $97.97\,(4)$ & $86.40\,(10)$ & $98.18\,(3)$ & $89.54\,(9)$ & $91.89\,(7)$ & $97.16\,(5)$ & $34.90\,(12)$ & $95.12\,(6)$ & $\underline{\mathbf{98.91\,(1)}}$ \\
    NATOPS\_5 & $89.49\,(9)$ & $96.27\,(6)$ & $98.78\,(2)$ & $96.17\,(7)$ & $79.95\,(10)$ & $96.54\,(5)$ & $54.90\,(12)$ & $89.91\,(8)$ & $\underline{\mathbf{99.48\,(1)}}$ & $68.82\,(11)$ & $96.80\,(4)$ & $98.39\,(3)$ \\
    PEMS-SF\_0 & $92.80\,(10)$ & $97.86\,(5)$ & $\underline{\mathbf{98.09\,(1)}}$ & $97.01\,(7)$ & $64.67\,(12)$ & $97.28\,(6)$ & $94.26\,(9)$ & $89.54\,(11)$ & $94.31\,(8)$ & $97.94\,(4)$ & $97.97\,(2)$ & $97.95\,(3)$ \\
    PEMS-SF\_1 & $68.73\,(12)$ & $87.32\,(8)$ & $90.95\,(2)$ & $89.57\,(6)$ & $73.60\,(11)$ & $\underline{\mathbf{92.24\,(1)}}$ & $81.38\,(10)$ & $90.36\,(5)$ & $87.89\,(7)$ & $86.94\,(9)$ & $90.79\,(4)$ & $90.80\,(3)$ \\
    PEMS-SF\_2 & $67.17\,(11)$ & $79.57\,(6)$ & $82.63\,(3)$ & $77.92\,(7)$ & $43.78\,(12)$ & $\underline{\mathbf{86.67\,(1)}}$ & $76.44\,(8)$ & $79.67\,(5)$ & $72.98\,(9)$ & $70.01\,(10)$ & $82.86\,(2)$ & $82.58\,(4)$ \\
    PEMS-SF\_3 & $62.28\,(11)$ & $80.06\,(7)$ & $\underline{\mathbf{86.61\,(1)}}$ & $83.70\,(5)$ & $37.33\,(12)$ & $86.32\,(4)$ & $79.58\,(8)$ & $81.33\,(6)$ & $78.88\,(9)$ & $73.04\,(10)$ & $86.59\,(2)$ & $86.47\,(3)$ \\
    PEMS-SF\_4 & $63.55\,(11)$ & $73.83\,(9)$ & $79.56\,(6)$ & $75.27\,(7)$ & $49.37\,(12)$ & $\underline{\mathbf{85.47\,(1)}}$ & $82.07\,(3)$ & $82.29\,(2)$ & $74.83\,(8)$ & $72.56\,(10)$ & $81.53\,(4)$ & $80.43\,(5)$ \\
    PEMS-SF\_5 & $62.20\,(12)$ & $75.14\,(10)$ & $84.32\,(5)$ & $81.18\,(7)$ & $75.82\,(9)$ & $\underline{\mathbf{91.59\,(1)}}$ & $86.91\,(3)$ & $87.47\,(2)$ & $79.37\,(8)$ & $74.53\,(11)$ & $86.16\,(4)$ & $83.53\,(6)$ \\
    PEMS-SF\_6 & $93.18\,(11)$ & $98.20\,(5)$ & $\underline{\mathbf{99.35\,(1)}}$ & $97.84\,(7)$ & $18.77\,(12)$ & $98.90\,(3)$ & $96.73\,(9)$ & $98.08\,(6)$ & $97.22\,(8)$ & $96.21\,(10)$ & $99.14\,(2)$ & $98.78\,(4)$ \\
    PenDigits\_0 & $94.83\,(11)$ & $98.19\,(5)$ & $97.72\,(7)$ & $98.17\,(6)$ & $39.20\,(12)$ & $98.72\,(2)$ & $97.01\,(9)$ & $95.61\,(10)$ & $97.38\,(8)$ & $\underline{\mathbf{99.07\,(1)}}$ & $98.23\,(4)$ & $98.69\,(3)$ \\
    PenDigits\_1 & $95.37\,(8)$ & $97.31\,(7)$ & $95.21\,(9)$ & $97.60\,(6)$ & $39.77\,(12)$ & $98.56\,(3)$ & $97.76\,(4)$ & $92.93\,(11)$ & $94.97\,(10)$ & $97.74\,(5)$ & $99.16\,(2)$ & $\underline{\mathbf{99.29\,(1)}}$ \\
    PenDigits\_2 & $98.36\,(10)$ & $99.40\,(3)$ & $98.79\,(8)$ & $99.24\,(6)$ & $62.40\,(12)$ & $99.37\,(4)$ & $99.02\,(7)$ & $92.52\,(11)$ & $98.57\,(9)$ & $99.44\,(2)$ & $\underline{\mathbf{99.65\,(1)}}$ & $99.37\,(5)$ \\
    PenDigits\_3 & $99.17\,(7)$ & $\underline{\mathbf{99.57\,(1)}}$ & $99.25\,(6)$ & $99.38\,(4)$ & $26.02\,(12)$ & $99.43\,(3)$ & $95.23\,(10)$ & $88.45\,(11)$ & $99.07\,(8)$ & $98.82\,(9)$ & $99.45\,(2)$ & $99.31\,(5)$ \\
    PenDigits\_4 & $98.64\,(9)$ & $99.88\,(6)$ & $99.95\,(5)$ & $99.97\,(4)$ & $44.19\,(12)$ & $99.99\,(2)$ & $97.96\,(10)$ & $95.41\,(11)$ & $99.97\,(3)$ & $99.78\,(8)$ & $\underline{\mathbf{99.99\,(1)}}$ & $99.84\,(7)$ \\
    PenDigits\_5 & $99.14\,(6)$ & $\underline{\mathbf{99.86\,(1)}}$ & $99.32\,(5)$ & $98.71\,(7)$ & $68.44\,(12)$ & $99.33\,(4)$ & $98.62\,(8)$ & $97.98\,(10)$ & $98.14\,(9)$ & $97.69\,(11)$ & $99.79\,(3)$ & $99.80\,(2)$ \\
    PenDigits\_6 & $99.72\,(8)$ & $\underline{\mathbf{100.00\,(1)}}$ & $99.91\,(7)$ & $99.92\,(6)$ & $48.85\,(12)$ & $99.97\,(2)$ & $98.38\,(10)$ & $92.85\,(11)$ & $99.96\,(4)$ & $99.65\,(9)$ & $99.95\,(5)$ & $99.97\,(3)$ \\
    PenDigits\_7 & $94.39\,(8)$ & $96.03\,(4)$ & $95.70\,(7)$ & $95.71\,(6)$ & $41.97\,(12)$ & $96.08\,(3)$ & $90.42\,(9)$ & $88.46\,(11)$ & $95.79\,(5)$ & $90.11\,(10)$ & $96.56\,(2)$ & $\underline{\mathbf{97.07\,(1)}}$ \\
    PenDigits\_8 & $99.09\,(7)$ & $99.95\,(2)$ & $99.53\,(5)$ & $98.19\,(9)$ & $28.78\,(12)$ & $99.65\,(4)$ & $99.43\,(6)$ & $94.81\,(11)$ & $98.59\,(8)$ & $96.51\,(10)$ & $\underline{\mathbf{99.98\,(1)}}$ & $99.81\,(3)$ \\
    PenDigits\_9 & $97.45\,(7)$ & $\underline{\mathbf{99.25\,(1)}}$ & $98.52\,(4)$ & $97.89\,(6)$ & $26.10\,(12)$ & $97.91\,(5)$ & $97.01\,(9)$ & $91.52\,(11)$ & $97.39\,(8)$ & $95.80\,(10)$ & $99.08\,(3)$ & $99.14\,(2)$ \\
  \bottomrule
  \end{tabular}
}
}
\end{table*}

\begin{table*}[h!]
\centering
\caption
{Mean AUPR scores and rankings for all problems under the \textit{one-vs-rest} setting.}\label{tab:datasets_pr_rev0}
{
\resizebox{\linewidth}{!}{%
  \begin{tabular}{lcccccccccccc}
  \toprule
    Dataset\_Normality & IF & LOF & OCSVM & AutoAD & DAGMM & DeepSVDD & FixedTS & NeuTraL AD & TimesURL & MOMENT & TSINR & NeuCoReClass AD \\
    \midrule
ChlorineConcentration\_0 & $69.38\,(10)$ & $68.20\,(12)$ & $68.42\,(11)$ & $70.64\,(8)$ & $77.92\,(2)$ & $69.77\,(9)$ & $73.44\,(5)$ & $76.35\,(4)$ & $72.85\,(6)$ & $71.86\,(7)$ & $\underline{\mathbf{82.45\,(1)}}$ & $77.28\,(3)$ \\
    ChlorineConcentration\_1 & $79.85\,(10)$ & $79.36\,(11)$ & $81.85\,(5)$ & $80.77\,(8)$ & $80.80\,(7)$ & $80.07\,(9)$ & $78.62\,(12)$ & $83.21\,(3)$ & $81.56\,(6)$ & $82.77\,(4)$ & $\underline{\mathbf{86.58\,(1)}}$ & $85.41\,(2)$ \\
    ChlorineConcentration\_2 & $60.93\,(9)$ & $58.49\,(12)$ & $67.77\,(3)$ & $61.28\,(8)$ & $58.58\,(11)$ & $63.42\,(7)$ & $59.40\,(10)$ & $63.88\,(6)$ & $63.89\,(5)$ & $66.34\,(4)$ & $\underline{\mathbf{78.40\,(1)}}$ & $69.26\,(2)$ \\
    Computers\_0 & $37.67\,(12)$ & $57.70\,(6)$ & $54.30\,(8)$ & $\underline{\mathbf{64.09\,(1)}}$ & $51.65\,(10)$ & $52.43\,(9)$ & $40.45\,(11)$ & $58.01\,(5)$ & $61.76\,(2)$ & $61.02\,(3)$ & $55.18\,(7)$ & $59.60\,(4)$ \\
    Computers\_1 & $\underline{\mathbf{74.13\,(1)}}$ & $49.09\,(11)$ & $51.46\,(8)$ & $51.57\,(7)$ & $56.87\,(5)$ & $56.88\,(4)$ & $61.37\,(2)$ & $58.81\,(3)$ & $51.02\,(9)$ & $49.31\,(10)$ & $47.37\,(12)$ & $53.36\,(6)$ \\
    DistalPhalanxOutlineAgeGroup\_0 & $92.35\,(11)$ & $95.22\,(4)$ & $94.29\,(8)$ & $94.98\,(6)$ & $\underline{\mathbf{96.44\,(1)}}$ & $95.75\,(2)$ & $89.15\,(12)$ & $95.21\,(5)$ & $95.37\,(3)$ & $93.02\,(10)$ & $93.25\,(9)$ & $94.66\,(7)$ \\
    DistalPhalanxOutlineAgeGroup\_1 & $\underline{\mathbf{80.42\,(1)}}$ & $78.81\,(2)$ & $74.57\,(5)$ & $75.77\,(4)$ & $48.05\,(11)$ & $64.96\,(6)$ & $44.84\,(12)$ & $48.63\,(10)$ & $76.28\,(3)$ & $55.65\,(7)$ & $54.87\,(9)$ & $55.52\,(8)$ \\
    DistalPhalanxOutlineAgeGroup\_2 & $76.85\,(6)$ & $72.48\,(11)$ & $75.71\,(8)$ & $73.93\,(9)$ & $71.11\,(12)$ & $78.76\,(5)$ & $\underline{\mathbf{82.28\,(1)}}$ & $73.75\,(10)$ & $76.36\,(7)$ & $79.59\,(4)$ & $81.48\,(2)$ & $80.70\,(3)$ \\
    DistalPhalanxOutlineCorrect\_0 & $55.74\,(3)$ & $49.14\,(11)$ & $52.83\,(8)$ & $\underline{\mathbf{57.81\,(1)}}$ & $55.70\,(4)$ & $53.58\,(6)$ & $46.23\,(12)$ & $53.60\,(5)$ & $56.21\,(2)$ & $53.44\,(7)$ & $52.52\,(9)$ & $50.95\,(10)$ \\
    DistalPhalanxOutlineCorrect\_1 & $64.17\,(6)$ & $\underline{\mathbf{78.23\,(1)}}$ & $67.39\,(3)$ & $63.16\,(9)$ & $46.82\,(12)$ & $66.69\,(4)$ & $61.13\,(11)$ & $65.52\,(5)$ & $61.77\,(10)$ & $64.10\,(7)$ & $63.81\,(8)$ & $68.30\,(2)$ \\
    ECG200\_0 & $60.18\,(8)$ & $47.97\,(12)$ & $61.89\,(5)$ & $58.79\,(9)$ & $57.17\,(10)$ & $60.76\,(7)$ & $67.69\,(2)$ & $62.56\,(3)$ & $56.93\,(11)$ & $\underline{\mathbf{73.51\,(1)}}$ & $61.99\,(4)$ & $61.68\,(6)$ \\
    ECG200\_1 & $79.70\,(7)$ & $82.24\,(4)$ & $81.23\,(5)$ & $77.34\,(8)$ & $83.43\,(2)$ & $\underline{\mathbf{84.10\,(1)}}$ & $62.48\,(11)$ & $63.25\,(10)$ & $76.19\,(9)$ & $56.20\,(12)$ & $82.94\,(3)$ & $80.32\,(6)$ \\
    Earthquakes\_0 & $33.07\,(5)$ & $34.70\,(2)$ & $34.49\,(3)$ & $20.86\,(11)$ & $28.91\,(7)$ & $23.81\,(9)$ & $\underline{\mathbf{37.36\,(1)}}$ & $22.60\,(10)$ & $26.15\,(8)$ & $20.36\,(12)$ & $33.27\,(4)$ & $33.00\,(6)$ \\
    Earthquakes\_1 & $70.40\,(12)$ & $81.50\,(8)$ & $78.18\,(10)$ & $86.29\,(5)$ & $83.40\,(7)$ & $87.66\,(2)$ & $76.97\,(11)$ & $87.42\,(3)$ & $86.68\,(4)$ & $\underline{\mathbf{88.95\,(1)}}$ & $79.29\,(9)$ & $84.69\,(6)$ \\
    ElectricDevices\_0 & $\underline{\mathbf{96.53\,(1)}}$ & $94.76\,(6)$ & $95.99\,(3)$ & $94.91\,(5)$ & $90.61\,(12)$ & $94.38\,(8)$ & $95.69\,(4)$ & $93.55\,(10)$ & $96.28\,(2)$ & $92.05\,(11)$ & $94.58\,(7)$ & $93.86\,(9)$ \\
    ElectricDevices\_1 & $57.12\,(12)$ & $93.70\,(8)$ & $73.78\,(11)$ & $94.89\,(7)$ & $85.42\,(9)$ & $\underline{\mathbf{98.68\,(1)}}$ & $81.73\,(10)$ & $97.31\,(3)$ & $97.71\,(2)$ & $96.54\,(5)$ & $95.80\,(6)$ & $97.07\,(4)$ \\
    ElectricDevices\_2 & $97.00\,(9)$ & $96.48\,(11)$ & $97.29\,(8)$ & $97.65\,(7)$ & $91.33\,(12)$ & $98.71\,(4)$ & $97.79\,(6)$ & $\underline{\mathbf{99.27\,(1)}}$ & $98.78\,(3)$ & $97.93\,(5)$ & $96.99\,(10)$ & $98.82\,(2)$ \\
    ElectricDevices\_3 & $\underline{\mathbf{97.89\,(1)}}$ & $96.35\,(4)$ & $94.60\,(9)$ & $89.77\,(11)$ & $87.27\,(12)$ & $97.43\,(2)$ & $96.14\,(5)$ & $95.37\,(8)$ & $93.85\,(10)$ & $95.99\,(6)$ & $95.96\,(7)$ & $96.82\,(3)$ \\
    ElectricDevices\_4 & $77.21\,(11)$ & $86.35\,(6)$ & $82.96\,(8)$ & $\underline{\mathbf{91.48\,(1)}}$ & $75.36\,(12)$ & $86.91\,(4)$ & $79.22\,(10)$ & $82.70\,(9)$ & $85.35\,(7)$ & $87.17\,(3)$ & $89.00\,(2)$ & $86.62\,(5)$ \\
    ElectricDevices\_5 & $98.02\,(4)$ & $94.64\,(12)$ & $96.89\,(7)$ & $96.68\,(8)$ & $94.81\,(11)$ & $\underline{\mathbf{98.22\,(1)}}$ & $98.04\,(3)$ & $98.12\,(2)$ & $97.80\,(5)$ & $97.04\,(6)$ & $94.82\,(10)$ & $95.65\,(9)$ \\
    ElectricDevices\_6 & $94.12\,(11)$ & $96.93\,(3)$ & $95.83\,(9)$ & $\underline{\mathbf{97.46\,(1)}}$ & $95.13\,(10)$ & $96.12\,(7)$ & $93.69\,(12)$ & $96.78\,(4)$ & $97.08\,(2)$ & $96.58\,(5)$ & $96.21\,(6)$ & $96.04\,(8)$ \\
    Epilepsy\_0 & $90.44\,(10)$ & $96.94\,(6)$ & $95.75\,(7)$ & $98.02\,(4)$ & $96.97\,(5)$ & $\underline{\mathbf{99.29\,(1)}}$ & $83.77\,(12)$ & $91.88\,(8)$ & $98.49\,(2)$ & $89.38\,(11)$ & $91.54\,(9)$ & $98.15\,(3)$ \\
    Epilepsy\_1 & $59.71\,(10)$ & $59.98\,(9)$ & $58.87\,(11)$ & $58.24\,(12)$ & $95.71\,(3)$ & $\underline{\mathbf{96.72\,(1)}}$ & $91.78\,(4)$ & $96.04\,(2)$ & $89.75\,(6)$ & $67.86\,(7)$ & $62.64\,(8)$ & $90.76\,(5)$ \\
    Epilepsy\_2 & $93.01\,(11)$ & $95.97\,(10)$ & $98.11\,(7)$ & $98.97\,(5)$ & $98.56\,(6)$ & $\underline{\mathbf{99.80\,(1)}}$ & $89.63\,(12)$ & $99.75\,(3)$ & $99.43\,(4)$ & $97.16\,(9)$ & $97.55\,(8)$ & $99.76\,(2)$ \\
    Epilepsy\_3 & $99.26\,(6)$ & $99.56\,(4)$ & $99.57\,(3)$ & $99.67\,(2)$ & $86.59\,(12)$ & $\underline{\mathbf{99.86\,(1)}}$ & $89.93\,(11)$ & $94.01\,(10)$ & $99.18\,(7)$ & $98.45\,(9)$ & $99.11\,(8)$ & $99.54\,(5)$ \\
    EthanolConcentration\_0 & $84.62\,(4)$ & $86.09\,(2)$ & $84.92\,(3)$ & $83.58\,(5)$ & $80.24\,(10)$ & $80.40\,(9)$ & $73.45\,(11)$ & $72.90\,(12)$ & $83.36\,(6)$ & $\underline{\mathbf{87.51\,(1)}}$ & $83.25\,(7)$ & $80.89\,(8)$ \\
    EthanolConcentration\_1 & $80.49\,(2)$ & $\underline{\mathbf{81.69\,(1)}}$ & $80.11\,(3)$ & $78.63\,(7)$ & $77.70\,(10)$ & $78.83\,(6)$ & $77.73\,(9)$ & $74.01\,(12)$ & $76.41\,(11)$ & $79.85\,(4)$ & $79.06\,(5)$ & $77.98\,(8)$ \\
    EthanolConcentration\_2 & $77.27\,(7)$ & $\underline{\mathbf{78.92\,(1)}}$ & $77.46\,(6)$ & $76.65\,(8)$ & $77.51\,(5)$ & $76.52\,(9)$ & $72.10\,(12)$ & $74.73\,(11)$ & $78.32\,(2)$ & $77.88\,(4)$ & $78.01\,(3)$ & $76.43\,(10)$ \\
    EthanolConcentration\_3 & $72.84\,(9)$ & $74.14\,(8)$ & $72.04\,(11)$ & $72.49\,(10)$ & $78.20\,(2)$ & $74.14\,(8)$ & $77.33\,(3)$ & $\underline{\mathbf{80.34\,(1)}}$ & $75.22\,(6)$ & $71.80\,(12)$ & $75.54\,(5)$ & $76.91\,(4)$ \\
    EthanolLevel\_0 & $72.83\,(11)$ & $74.02\,(10)$ & $75.06\,(9)$ & $76.62\,(6)$ & $75.21\,(8)$ & $72.76\,(12)$ & $\underline{\mathbf{85.50\,(1)}}$ & $79.59\,(3)$ & $76.99\,(5)$ & $75.84\,(7)$ & $80.99\,(2)$ & $78.77\,(4)$ \\
    EthanolLevel\_1 & $79.20\,(4)$ & $76.73\,(9)$ & $78.45\,(5)$ & $77.96\,(7)$ & $76.65\,(10)$ & $78.38\,(6)$ & $\underline{\mathbf{81.18\,(1)}}$ & $76.27\,(11)$ & $76.15\,(12)$ & $76.85\,(8)$ & $79.57\,(3)$ & $79.62\,(2)$ \\
    EthanolLevel\_2 & $\underline{\mathbf{78.32\,(1)}}$ & $76.78\,(7)$ & $77.74\,(3)$ & $77.46\,(4)$ & $75.57\,(9)$ & $75.68\,(8)$ & $73.80\,(11)$ & $73.57\,(12)$ & $75.55\,(10)$ & $77.39\,(6)$ & $77.44\,(5)$ & $77.83\,(2)$ \\
    EthanolLevel\_3 & $77.25\,(7)$ & $78.04\,(3)$ & $77.61\,(5)$ & $77.64\,(4)$ & $75.12\,(11)$ & $76.45\,(10)$ & $76.76\,(9)$ & $77.26\,(6)$ & $74.06\,(12)$ & $78.39\,(2)$ & $\underline{\mathbf{83.20\,(1)}}$ & $77.02\,(8)$ \\
    FaceDetection\_0 & $51.64\,(4)$ & $51.86\,(2)$ & $51.67\,(3)$ & $50.89\,(9)$ & $49.88\,(12)$ & $51.07\,(8)$ & $50.24\,(11)$ & $50.43\,(10)$ & $51.34\,(6)$ & $51.11\,(7)$ & $\underline{\mathbf{51.91\,(1)}}$ & $51.50\,(5)$ \\
    FaceDetection\_1 & $49.02\,(7)$ & $48.74\,(11)$ & $48.92\,(9)$ & $49.43\,(3)$ & $\underline{\mathbf{50.34\,(1)}}$ & $48.60\,(12)$ & $49.25\,(5)$ & $49.96\,(2)$ & $49.30\,(4)$ & $49.16\,(6)$ & $48.74\,(10)$ & $48.98\,(8)$ \\
    FingerMovements\_0 & $\underline{\mathbf{64.68\,(1)}}$ & $61.86\,(8)$ & $63.03\,(4)$ & $62.83\,(5)$ & $62.43\,(7)$ & $59.83\,(9)$ & $50.17\,(11)$ & $48.86\,(12)$ & $62.49\,(6)$ & $64.62\,(2)$ & $63.14\,(3)$ & $58.07\,(10)$ \\
    FingerMovements\_1 & $40.72\,(10)$ & $41.64\,(6)$ & $40.39\,(12)$ & $41.44\,(7)$ & $\underline{\mathbf{55.67\,(1)}}$ & $40.53\,(11)$ & $48.41\,(3)$ & $48.57\,(2)$ & $40.80\,(9)$ & $41.29\,(8)$ & $43.80\,(4)$ & $42.76\,(5)$ \\
    FordA\_0 & $49.28\,(3)$ & $48.01\,(6)$ & $48.28\,(5)$ & $38.80\,(12)$ & $48.66\,(4)$ & $\underline{\mathbf{57.84\,(1)}}$ & $43.66\,(7)$ & $49.34\,(2)$ & $43.15\,(9)$ & $40.82\,(10)$ & $43.46\,(8)$ & $39.53\,(11)$ \\
    FordA\_1 & $53.62\,(12)$ & $70.87\,(8)$ & $68.95\,(9)$ & $78.76\,(4)$ & $55.19\,(11)$ & $\underline{\mathbf{81.76\,(1)}}$ & $68.69\,(10)$ & $76.15\,(7)$ & $78.27\,(5)$ & $78.98\,(3)$ & $77.60\,(6)$ & $81.14\,(2)$ \\
    FordB\_0 & $51.37\,(5)$ & $53.80\,(2)$ & $\underline{\mathbf{54.87\,(1)}}$ & $44.33\,(11)$ & $52.35\,(4)$ & $49.61\,(6)$ & $48.07\,(7)$ & $46.56\,(9)$ & $42.57\,(12)$ & $47.04\,(8)$ & $52.75\,(3)$ & $45.15\,(10)$ \\
    FordB\_1 & $49.93\,(12)$ & $58.69\,(9)$ & $59.90\,(7)$ & $66.79\,(5)$ & $52.74\,(11)$ & $68.67\,(4)$ & $56.74\,(10)$ & $69.30\,(3)$ & $69.42\,(2)$ & $66.08\,(6)$ & $59.51\,(8)$ & $\underline{\mathbf{69.63\,(1)}}$ \\
    HandMovementDirection\_0 & $83.13\,(9)$ & $83.77\,(6)$ & $84.15\,(4)$ & $\underline{\mathbf{86.18\,(1)}}$ & $80.46\,(11)$ & $84.08\,(5)$ & $78.31\,(12)$ & $80.83\,(10)$ & $84.66\,(2)$ & $83.30\,(8)$ & $83.49\,(7)$ & $84.26\,(3)$ \\
    HandMovementDirection\_1 & $66.22\,(9)$ & $72.93\,(3)$ & $72.27\,(5)$ & $\underline{\mathbf{76.83\,(1)}}$ & $67.88\,(7)$ & $64.67\,(10)$ & $62.17\,(12)$ & $64.45\,(11)$ & $71.68\,(6)$ & $67.25\,(8)$ & $72.55\,(4)$ & $75.28\,(2)$ \\
    HandMovementDirection\_2 & $76.35\,(11)$ & $76.61\,(10)$ & $76.68\,(9)$ & $79.82\,(4)$ & $80.15\,(3)$ & $\underline{\mathbf{85.08\,(1)}}$ & $82.78\,(2)$ & $78.81\,(5)$ & $77.90\,(6)$ & $76.13\,(12)$ & $76.85\,(8)$ & $77.32\,(7)$ \\
    HandMovementDirection\_3 & $81.52\,(10)$ & $83.19\,(4)$ & $82.25\,(7)$ & $84.00\,(3)$ & $82.28\,(6)$ & $\underline{\mathbf{86.12\,(1)}}$ & $79.49\,(12)$ & $82.86\,(5)$ & $86.06\,(2)$ & $79.99\,(11)$ & $82.13\,(8)$ & $81.53\,(9)$ \\
    Heartbeat\_0 & $39.30\,(3)$ & $38.94\,(5)$ & $\underline{\mathbf{39.91\,(1)}}$ & $35.84\,(8)$ & $28.60\,(12)$ & $32.78\,(9)$ & $29.19\,(10)$ & $28.89\,(11)$ & $37.76\,(6)$ & $39.13\,(4)$ & $39.65\,(2)$ & $36.25\,(7)$ \\
    Heartbeat\_1 & $66.64\,(9)$ & $66.17\,(11)$ & $66.20\,(10)$ & $67.29\,(6)$ & $\underline{\mathbf{81.36\,(1)}}$ & $74.87\,(3)$ & $77.49\,(2)$ & $71.32\,(4)$ & $67.04\,(7)$ & $65.93\,(12)$ & $66.66\,(8)$ & $67.60\,(5)$ \\
    MotorImagery\_0 & $48.21\,(5)$ & $47.16\,(8)$ & $46.45\,(11)$ & $46.89\,(10)$ & $\underline{\mathbf{75.00\,(1)}}$ & $47.66\,(7)$ & $52.63\,(3)$ & $56.32\,(2)$ & $49.93\,(4)$ & $47.16\,(9)$ & $47.73\,(6)$ & $46.27\,(12)$ \\
    MotorImagery\_1 & $52.51\,(10)$ & $53.82\,(9)$ & $56.46\,(3)$ & $55.66\,(4)$ & $\underline{\mathbf{67.99\,(1)}}$ & $54.45\,(7)$ & $46.22\,(12)$ & $47.56\,(11)$ & $56.72\,(2)$ & $54.65\,(5)$ & $54.30\,(8)$ & $54.60\,(6)$ \\
    NATOPS\_0 & $94.62\,(10)$ & $95.31\,(7)$ & $95.79\,(5)$ & $96.28\,(2)$ & $93.19\,(11)$ & $\underline{\mathbf{96.32\,(1)}}$ & $92.61\,(12)$ & $95.04\,(9)$ & $96.03\,(4)$ & $95.20\,(8)$ & $95.64\,(6)$ & $96.10\,(3)$ \\
    NATOPS\_1 & $95.21\,(10)$ & $96.36\,(8)$ & $96.69\,(5)$ & $96.51\,(6)$ & $93.67\,(11)$ & $97.05\,(2)$ & $88.99\,(12)$ & $96.04\,(9)$ & $\underline{\mathbf{97.07\,(1)}}$ & $96.39\,(7)$ & $96.96\,(3)$ & $96.87\,(4)$ \\
    NATOPS\_2 & $96.66\,(10)$ & $97.69\,(8)$ & $97.70\,(7)$ & $97.96\,(3)$ & $93.80\,(11)$ & $97.82\,(5)$ & $86.44\,(12)$ & $96.97\,(9)$ & $\underline{\mathbf{98.16\,(1)}}$ & $97.81\,(6)$ & $98.09\,(2)$ & $97.91\,(4)$ \\
    NATOPS\_3 & $94.49\,(11)$ & $95.67\,(8)$ & $97.93\,(4)$ & $98.95\,(3)$ & $94.80\,(10)$ & $99.37\,(2)$ & $95.50\,(9)$ & $97.75\,(5)$ & $96.35\,(7)$ & $74.99\,(12)$ & $96.74\,(6)$ & $\underline{\mathbf{99.50\,(1)}}$ \\
    NATOPS\_4 & $95.71\,(11)$ & $98.21\,(8)$ & $99.64\,(3)$ & $99.58\,(4)$ & $96.91\,(10)$ & $99.64\,(2)$ & $97.29\,(9)$ & $98.39\,(7)$ & $99.38\,(5)$ & $76.59\,(12)$ & $98.80\,(6)$ & $\underline{\mathbf{99.77\,(1)}}$ \\
    NATOPS\_5 & $97.38\,(9)$ & $99.21\,(4)$ & $99.75\,(2)$ & $99.02\,(7)$ & $96.00\,(10)$ & $99.17\,(6)$ & $86.30\,(12)$ & $97.58\,(8)$ & $\underline{\mathbf{99.89\,(1)}}$ & $92.55\,(11)$ & $99.20\,(5)$ & $99.64\,(3)$ \\
    PEMS-SF\_0 & $98.34\,(9)$ & $99.55\,(5)$ & $\underline{\mathbf{99.59\,(1)}}$ & $99.37\,(7)$ & $93.55\,(12)$ & $99.42\,(6)$ & $98.10\,(10)$ & $94.37\,(11)$ & $98.78\,(8)$ & $99.59\,(2)$ & $99.56\,(3)$ & $99.56\,(4)$ \\
    PEMS-SF\_1 & $89.49\,(12)$ & $96.58\,(7)$ & $97.26\,(4)$ & $96.61\,(6)$ & $95.91\,(10)$ & $\underline{\mathbf{97.61\,(1)}}$ & $96.32\,(9)$ & $97.35\,(2)$ & $96.43\,(8)$ & $95.87\,(11)$ & $97.28\,(3)$ & $97.24\,(5)$ \\
    PEMS-SF\_2 & $89.10\,(11)$ & $94.92\,(7)$ & $95.55\,(4)$ & $93.61\,(8)$ & $83.15\,(12)$ & $\underline{\mathbf{97.18\,(1)}}$ & $95.45\,(5)$ & $95.14\,(6)$ & $91.36\,(9)$ & $89.12\,(10)$ & $95.75\,(2)$ & $95.63\,(3)$ \\
    PEMS-SF\_3 & $90.27\,(11)$ & $95.84\,(7)$ & $96.60\,(2)$ & $95.95\,(6)$ & $82.67\,(12)$ & $96.41\,(5)$ & $96.44\,(4)$ & $95.52\,(8)$ & $94.81\,(9)$ & $92.57\,(10)$ & $\underline{\mathbf{96.60\,(1)}}$ & $96.59\,(3)$ \\
    PEMS-SF\_4 & $91.68\,(11)$ & $94.89\,(7)$ & $95.72\,(6)$ & $94.19\,(8)$ & $90.13\,(12)$ & $\underline{\mathbf{97.04\,(1)}}$ & $96.88\,(2)$ & $96.46\,(3)$ & $93.29\,(9)$ & $92.75\,(10)$ & $96.30\,(4)$ & $95.96\,(5)$ \\
    PEMS-SF\_5 & $85.85\,(12)$ & $93.23\,(10)$ & $95.71\,(6)$ & $94.20\,(8)$ & $95.88\,(5)$ & $\underline{\mathbf{97.78\,(1)}}$ & $97.22\,(2)$ & $96.86\,(3)$ & $94.19\,(9)$ & $91.70\,(11)$ & $96.44\,(4)$ & $95.57\,(7)$ \\
    PEMS-SF\_6 & $97.17\,(11)$ & $99.77\,(5)$ & $\underline{\mathbf{99.91\,(1)}}$ & $99.72\,(7)$ & $80.26\,(12)$ & $99.85\,(3)$ & $99.57\,(9)$ & $99.72\,(6)$ & $99.63\,(8)$ & $99.50\,(10)$ & $99.89\,(2)$ & $99.84\,(4)$ \\
    PenDigits\_0 & $98.28\,(11)$ & $99.69\,(4)$ & $99.38\,(8)$ & $99.66\,(5)$ & $88.34\,(12)$ & $99.79\,(2)$ & $99.57\,(6)$ & $99.25\,(10)$ & $99.29\,(9)$ & $\underline{\mathbf{99.87\,(1)}}$ & $99.52\,(7)$ & $99.77\,(3)$ \\
    PenDigits\_1 & $99.43\,(8)$ & $99.66\,(7)$ & $99.41\,(9)$ & $99.71\,(5)$ & $87.97\,(12)$ & $99.83\,(3)$ & $99.71\,(6)$ & $99.06\,(11)$ & $99.39\,(10)$ & $99.72\,(4)$ & $99.89\,(2)$ & $\underline{\mathbf{99.91\,(1)}}$ \\
    PenDigits\_2 & $99.81\,(10)$ & $99.93\,(3)$ & $99.86\,(8)$ & $99.91\,(6)$ & $94.10\,(12)$ & $99.93\,(4)$ & $99.88\,(7)$ & $98.85\,(11)$ & $99.84\,(9)$ & $99.93\,(2)$ & $\underline{\mathbf{99.96\,(1)}}$ & $99.91\,(6)$ \\
    PenDigits\_3 & $99.89\,(3)$ & $99.87\,(4)$ & $99.87\,(4)$ & $99.89\,(2)$ & $85.99\,(12)$ & $\underline{\mathbf{99.91\,(1)}}$ & $99.48\,(10)$ & $97.87\,(11)$ & $99.87\,(6)$ & $99.83\,(8)$ & $99.84\,(7)$ & $99.80\,(9)$ \\
    PenDigits\_4 & $99.83\,(9)$ & $99.99\,(6)$ & $99.99\,(6)$ & $100.00\,(4)$ & $88.30\,(12)$ & $\underline{\mathbf{100.00\,(2)}}$ & $99.75\,(10)$ & $99.34\,(11)$ & $100.00\,(4)$ & $99.97\,(8)$ & $\underline{\mathbf{100.00\,(2)}}$ & $99.98\,(7)$ \\
    PenDigits\_5 & $99.91\,(6)$ & $\underline{\mathbf{99.99\,(1)}}$ & $99.92\,(5)$ & $99.86\,(7)$ & $96.76\,(12)$ & $99.93\,(4)$ & $99.85\,(8)$ & $99.71\,(11)$ & $99.79\,(9)$ & $99.73\,(10)$ & $99.98\,(2)$ & $99.96\,(3)$ \\
    PenDigits\_6 & $99.97\,(8)$ & $\underline{\mathbf{100.00\,(2)}}$ & $99.99\,(6)$ & $99.99\,(7)$ & $91.14\,(12)$ & $\underline{\mathbf{100.00\,(2)}}$ & $99.82\,(10)$ & $98.92\,(11)$ & $100.00\,(4)$ & $99.96\,(9)$ & $99.99\,(5)$ & $100.00\,(3)$ \\
    PenDigits\_7 & $99.10\,(8)$ & $99.25\,(7)$ & $99.34\,(4)$ & $99.30\,(5)$ & $90.20\,(12)$ & $99.37\,(3)$ & $98.74\,(9)$ & $97.92\,(11)$ & $99.30\,(6)$ & $98.11\,(10)$ & $99.47\,(2)$ & $\underline{\mathbf{99.54\,(1)}}$ \\
    PenDigits\_8 & $99.88\,(7)$ & $99.99\,(2)$ & $99.94\,(5)$ & $99.80\,(9)$ & $85.30\,(12)$ & $99.96\,(3)$ & $99.92\,(6)$ & $99.04\,(11)$ & $99.84\,(8)$ & $99.60\,(10)$ & $\underline{\mathbf{100.00\,(1)}}$ & $99.95\,(4)$ \\
    PenDigits\_9 & $99.67\,(6)$ & $99.83\,(2)$ & $99.77\,(4)$ & $99.64\,(9)$ & $84.56\,(12)$ & $99.70\,(5)$ & $99.64\,(8)$ & $98.92\,(11)$ & $99.66\,(7)$ & $99.41\,(10)$ & $99.79\,(3)$ & $\underline{\mathbf{99.88\,(1)}}$ \\
  \bottomrule
  \end{tabular}
}
}
\end{table*}

\begin{table*}[h!]
\centering
\scriptsize
\caption{Mean AUROC scores and rankings for all problems under the \textit{(N-1)-vs-rest} setting.}
\label{tab:datasets_roc_rev1}
{
\resizebox{\linewidth}{!}{%
  \begin{tabular}{lcccccccccccc}
  \toprule
    Dataset\_Normality & IF & LOF & OCSVM & AutoAD & DAGMM & DeepSVDD & FixedTS & NeuTraL AD & TimesURL & MOMENT & TSINR & NeuCoReClass AD \\
    \midrule
ChlorineConcentration\_0 & $64.54\,(9)$ & $63.35\,(10)$ & $73.80\,(2)$ & $64.59\,(8)$ & $55.81\,(12)$ & $65.15\,(7)$ & $60.74\,(11)$ & $66.97\,(5)$ & $65.60\,(6)$ & $68.85\,(4)$ & $\underline{\mathbf{81.78\,(1)}}$ & $71.67\,(3)$ \\
    ChlorineConcentration\_1 & $47.88\,(9)$ & $45.99\,(12)$ & $46.50\,(11)$ & $46.54\,(10)$ & $49.81\,(6)$ & $50.18\,(4)$ & $49.86\,(5)$ & $52.16\,(3)$ & $49.78\,(7)$ & $49.39\,(8)$ & $\underline{\mathbf{73.45\,(1)}}$ & $57.24\,(2)$ \\
    ChlorineConcentration\_2 & $45.50\,(9)$ & $45.32\,(10)$ & $42.26\,(12)$ & $46.41\,(7)$ & $47.78\,(6)$ & $43.03\,(11)$ & $47.86\,(5)$ & $51.02\,(3)$ & $45.93\,(8)$ & $49.47\,(4)$ & $\underline{\mathbf{76.60\,(1)}}$ & $60.82\,(2)$ \\
    Computers\_0 & $\underline{\mathbf{69.17\,(1)}}$ & $49.56\,(9)$ & $54.59\,(6)$ & $52.46\,(8)$ & $44.28\,(12)$ & $55.90\,(4)$ & $60.97\,(2)$ & $59.27\,(3)$ & $54.57\,(7)$ & $49.34\,(10)$ & $47.48\,(11)$ & $54.81\,(5)$ \\
    Computers\_1 & $30.28\,(12)$ & $62.00\,(2)$ & $54.21\,(8)$ & $\underline{\mathbf{63.51\,(1)}}$ & $45.68\,(10)$ & $50.42\,(9)$ & $35.72\,(11)$ & $57.24\,(6)$ & $56.34\,(7)$ & $61.32\,(3)$ & $58.14\,(5)$ & $60.49\,(4)$ \\
    DistalPhalanxOutlineAgeGroup\_0 & $83.76\,(7)$ & $94.25\,(2)$ & $79.73\,(9)$ & $71.87\,(11)$ & $59.89\,(12)$ & $83.13\,(8)$ & $\underline{\mathbf{94.40\,(1)}}$ & $92.45\,(4)$ & $78.31\,(10)$ & $86.67\,(6)$ & $89.29\,(5)$ & $93.46\,(3)$ \\
    DistalPhalanxOutlineAgeGroup\_1 & $69.87\,(6)$ & $47.25\,(12)$ & $70.68\,(4)$ & $\underline{\mathbf{71.51\,(1)}}$ & $55.11\,(10)$ & $71.25\,(2)$ & $55.09\,(11)$ & $61.12\,(9)$ & $70.77\,(3)$ & $70.04\,(5)$ & $69.13\,(8)$ & $69.25\,(7)$ \\
    DistalPhalanxOutlineAgeGroup\_2 & $72.58\,(2)$ & $64.91\,(5)$ & $67.57\,(4)$ & $70.67\,(3)$ & $47.33\,(7)$ & $52.29\,(6)$ & $20.01\,(12)$ & $31.39\,(8)$ & $\underline{\mathbf{74.22\,(1)}}$ & $27.61\,(9)$ & $21.32\,(11)$ & $25.39\,(10)$ \\
    DistalPhalanxOutlineCorrect\_0 & $71.85\,(6)$ & $\underline{\mathbf{83.08\,(1)}}$ & $71.96\,(5)$ & $67.09\,(10)$ & $58.36\,(12)$ & $70.80\,(7)$ & $72.79\,(3)$ & $73.44\,(2)$ & $65.97\,(11)$ & $67.27\,(9)$ & $67.85\,(8)$ & $72.02\,(4)$ \\
    DistalPhalanxOutlineCorrect\_1 & $50.29\,(3)$ & $40.74\,(11)$ & $46.97\,(6)$ & $52.58\,(2)$ & $49.62\,(4)$ & $47.43\,(5)$ & $33.16\,(12)$ & $46.83\,(7)$ & $\underline{\mathbf{54.14\,(1)}}$ & $46.60\,(8)$ & $45.51\,(9)$ & $43.40\,(10)$ \\
    ECG200\_0 & $86.80\,(7)$ & $89.11\,(4)$ & $87.41\,(5)$ & $85.16\,(8)$ & $89.72\,(2)$ & $\underline{\mathbf{89.91\,(1)}}$ & $67.79\,(12)$ & $73.63\,(10)$ & $83.03\,(9)$ & $72.45\,(11)$ & $89.37\,(3)$ & $87.22\,(6)$ \\
    ECG200\_1 & $52.16\,(8)$ & $22.68\,(12)$ & $56.51\,(4)$ & $51.97\,(9)$ & $42.77\,(11)$ & $54.02\,(6)$ & $59.63\,(2)$ & $52.40\,(7)$ & $44.53\,(10)$ & $\underline{\mathbf{66.28\,(1)}}$ & $57.10\,(3)$ & $56.45\,(5)$ \\
    Earthquakes\_0 & $35.95\,(12)$ & $58.26\,(8)$ & $53.93\,(10)$ & $67.51\,(5)$ & $62.19\,(6)$ & $70.71\,(3)$ & $48.29\,(11)$ & $68.17\,(4)$ & $71.31\,(2)$ & $\underline{\mathbf{73.16\,(1)}}$ & $55.00\,(9)$ & $58.48\,(7)$ \\
    Earthquakes\_1 & $65.55\,(4)$ & $66.18\,(2)$ & $64.59\,(5)$ & $41.26\,(11)$ & $51.11\,(8)$ & $47.71\,(10)$ & $65.88\,(3)$ & $47.99\,(9)$ & $52.37\,(7)$ & $40.97\,(12)$ & $64.05\,(6)$ & $\underline{\mathbf{66.70\,(1)}}$ \\
    ElectricDevices\_0 & $35.11\,(12)$ & $56.19\,(4)$ & $50.06\,(9)$ & $63.75\,(3)$ & $52.57\,(7)$ & $54.60\,(6)$ & $37.72\,(11)$ & $55.89\,(5)$ & $\underline{\mathbf{66.55\,(1)}}$ & $50.13\,(8)$ & $46.01\,(10)$ & $64.79\,(2)$ \\
    ElectricDevices\_1 & $91.64\,(3)$ & $91.32\,(4)$ & $86.15\,(8)$ & $45.25\,(11)$ & $40.32\,(12)$ & $90.58\,(5)$ & $80.03\,(10)$ & $81.24\,(9)$ & $86.38\,(7)$ & $89.26\,(6)$ & $\underline{\mathbf{92.38\,(1)}}$ & $91.78\,(2)$ \\
    ElectricDevices\_2 & $28.98\,(11)$ & $67.35\,(3)$ & $44.76\,(7)$ & $\underline{\mathbf{79.92\,(1)}}$ & $53.76\,(5)$ & $43.36\,(8)$ & $38.63\,(9)$ & $36.55\,(10)$ & $27.75\,(12)$ & $72.78\,(2)$ & $53.16\,(6)$ & $56.24\,(4)$ \\
    ElectricDevices\_3 & $17.77\,(12)$ & $29.51\,(9)$ & $29.95\,(8)$ & $\underline{\mathbf{79.64\,(1)}}$ & $49.03\,(2)$ & $43.79\,(4)$ & $22.67\,(11)$ & $36.11\,(5)$ & $46.77\,(3)$ & $32.70\,(7)$ & $26.11\,(10)$ & $34.22\,(6)$ \\
    ElectricDevices\_4 & $63.35\,(6)$ & $60.22\,(7)$ & $53.73\,(10)$ & $27.60\,(12)$ & $58.96\,(9)$ & $70.45\,(5)$ & $\underline{\mathbf{79.17\,(1)}}$ & $72.43\,(3)$ & $71.10\,(4)$ & $51.66\,(11)$ & $59.84\,(8)$ & $74.21\,(2)$ \\
    ElectricDevices\_5 & $17.88\,(12)$ & $31.71\,(11)$ & $43.09\,(9)$ & $\underline{\mathbf{83.25\,(1)}}$ & $54.59\,(3)$ & $46.12\,(7)$ & $33.89\,(10)$ & $48.90\,(6)$ & $49.16\,(5)$ & $52.00\,(4)$ & $43.76\,(8)$ & $61.96\,(2)$ \\
    ElectricDevices\_6 & $60.54\,(6)$ & $62.76\,(3)$ & $46.61\,(8)$ & $33.68\,(12)$ & $50.93\,(7)$ & $45.59\,(9)$ & $\underline{\mathbf{84.90\,(1)}}$ & $62.54\,(4)$ & $42.85\,(10)$ & $37.09\,(11)$ & $62.49\,(5)$ & $65.48\,(2)$ \\
    Epilepsy\_0 & $44.22\,(11)$ & $45.56\,(10)$ & $46.01\,(9)$ & $61.20\,(6)$ & $66.81\,(5)$ & $77.01\,(2)$ & $76.25\,(3)$ & $59.15\,(7)$ & $26.44\,(12)$ & $\underline{\mathbf{79.18\,(1)}}$ & $56.13\,(8)$ & $70.50\,(4)$ \\
    Epilepsy\_1 & $96.68\,(9)$ & $99.25\,(6)$ & $99.57\,(5)$ & $99.87\,(3)$ & $95.24\,(10)$ & $99.90\,(2)$ & $57.61\,(12)$ & $79.66\,(11)$ & $\underline{\mathbf{99.96\,(1)}}$ & $98.34\,(8)$ & $98.67\,(7)$ & $99.65\,(4)$ \\
    Epilepsy\_2 & $60.49\,(8)$ & $49.32\,(10)$ & $57.07\,(9)$ & $62.96\,(7)$ & $66.01\,(6)$ & $80.34\,(3)$ & $\underline{\mathbf{83.95\,(1)}}$ & $78.93\,(4)$ & $75.72\,(5)$ & $27.94\,(12)$ & $44.91\,(11)$ & $82.88\,(2)$ \\
    Epilepsy\_3 & $22.50\,(9)$ & $35.32\,(7)$ & $11.75\,(12)$ & $22.29\,(10)$ & $80.47\,(3)$ & $69.16\,(6)$ & $\underline{\mathbf{92.76\,(1)}}$ & $83.62\,(2)$ & $75.13\,(5)$ & $23.62\,(8)$ & $18.18\,(11)$ & $79.36\,(4)$ \\
    EthanolConcentration\_0 & $45.95\,(11)$ & $50.53\,(6)$ & $46.71\,(10)$ & $44.62\,(12)$ & $47.34\,(8)$ & $50.31\,(7)$ & $46.74\,(9)$ & $\underline{\mathbf{56.15\,(1)}}$ & $54.77\,(2)$ & $51.22\,(4)$ & $52.13\,(3)$ & $50.72\,(5)$ \\
    EthanolConcentration\_1 & $42.08\,(12)$ & $50.78\,(4)$ & $46.19\,(9)$ & $45.97\,(10)$ & $47.43\,(8)$ & $47.88\,(5)$ & $52.54\,(2)$ & $\underline{\mathbf{52.85\,(1)}}$ & $51.79\,(3)$ & $47.52\,(7)$ & $47.63\,(6)$ & $44.15\,(11)$ \\
    EthanolConcentration\_2 & $\underline{\mathbf{55.00\,(1)}}$ & $48.98\,(8)$ & $52.88\,(2)$ & $50.33\,(3)$ & $49.82\,(5)$ & $49.41\,(6)$ & $50.06\,(4)$ & $48.63\,(9)$ & $49.12\,(7)$ & $45.82\,(12)$ & $45.94\,(11)$ & $48.24\,(10)$ \\
    EthanolConcentration\_3 & $65.65\,(2)$ & $45.10\,(11)$ & $60.88\,(5)$ & $63.65\,(3)$ & $49.69\,(9)$ & $56.27\,(7)$ & $53.86\,(8)$ & $41.57\,(12)$ & $48.78\,(10)$ & $\underline{\mathbf{66.23\,(1)}}$ & $61.90\,(4)$ & $57.64\,(6)$ \\
    EthanolLevel\_0 & $57.29\,(4)$ & $56.80\,(5)$ & $57.47\,(2)$ & $52.79\,(8)$ & $50.27\,(10)$ & $\underline{\mathbf{59.10\,(1)}}$ & $49.40\,(11)$ & $52.08\,(9)$ & $48.63\,(12)$ & $57.35\,(3)$ & $54.70\,(7)$ & $56.22\,(6)$ \\
    EthanolLevel\_1 & $47.40\,(8)$ & $47.22\,(9)$ & $47.63\,(5)$ & $\underline{\mathbf{50.51\,(1)}}$ & $48.66\,(3)$ & $45.71\,(12)$ & $45.71\,(11)$ & $47.41\,(7)$ & $48.91\,(2)$ & $48.25\,(4)$ & $47.42\,(6)$ & $46.87\,(10)$ \\
    EthanolLevel\_2 & $43.89\,(11)$ & $44.10\,(10)$ & $43.85\,(12)$ & $45.74\,(9)$ & $\underline{\mathbf{53.39\,(1)}}$ & $48.75\,(5)$ & $53.00\,(2)$ & $51.13\,(3)$ & $50.68\,(4)$ & $46.40\,(6)$ & $45.89\,(7)$ & $45.88\,(8)$ \\
    EthanolLevel\_3 & $57.85\,(4)$ & $57.26\,(7)$ & $57.66\,(6)$ & $57.66\,(5)$ & $48.10\,(12)$ & $48.78\,(11)$ & $63.28\,(3)$ & $56.40\,(8)$ & $52.79\,(10)$ & $54.72\,(9)$ & $\underline{\mathbf{70.16\,(1)}}$ & $66.17\,(2)$ \\
    FaceDetection\_0 & $48.91\,(7)$ & $48.38\,(11)$ & $48.67\,(9)$ & $49.36\,(4)$ & $\underline{\mathbf{51.02\,(1)}}$ & $48.05\,(12)$ & $49.28\,(5)$ & $49.92\,(2)$ & $49.81\,(3)$ & $49.17\,(6)$ & $48.45\,(10)$ & $48.90\,(8)$ \\
    FaceDetection\_1 & $51.64\,(5)$ & $\underline{\mathbf{52.25\,(1)}}$ & $52.08\,(3)$ & $50.97\,(7)$ & $49.10\,(12)$ & $51.30\,(6)$ & $50.15\,(11)$ & $50.45\,(10)$ & $50.80\,(9)$ & $50.91\,(8)$ & $52.21\,(2)$ & $51.92\,(4)$ \\
    FingerMovements\_0 & $38.70\,(10)$ & $39.98\,(8)$ & $37.49\,(12)$ & $41.52\,(6)$ & $\underline{\mathbf{51.15\,(1)}}$ & $37.57\,(11)$ & $49.51\,(3)$ & $49.85\,(2)$ & $39.20\,(9)$ & $40.58\,(7)$ & $43.64\,(4)$ & $42.47\,(5)$ \\
    FingerMovements\_1 & $62.67\,(2)$ & $\underline{\mathbf{63.63\,(1)}}$ & $61.78\,(4)$ & $61.38\,(5)$ & $54.72\,(10)$ & $58.02\,(8)$ & $46.97\,(11)$ & $45.96\,(12)$ & $59.40\,(7)$ & $60.55\,(6)$ & $61.79\,(3)$ & $57.37\,(9)$ \\
    FordA\_0 & $52.80\,(12)$ & $59.80\,(9)$ & $58.02\,(10)$ & $69.74\,(6)$ & $57.30\,(11)$ & $74.74\,(3)$ & $73.96\,(4)$ & $77.21\,(2)$ & $\underline{\mathbf{77.75\,(1)}}$ & $68.70\,(7)$ & $65.97\,(8)$ & $70.55\,(5)$ \\
    FordA\_1 & $52.11\,(5)$ & $56.96\,(3)$ & $56.99\,(2)$ & $38.14\,(12)$ & $49.91\,(6)$ & $\underline{\mathbf{62.05\,(1)}}$ & $47.83\,(8)$ & $52.93\,(4)$ & $44.48\,(9)$ & $43.82\,(10)$ & $49.79\,(7)$ & $42.62\,(11)$ \\
    FordB\_0 & $49.65\,(11)$ & $52.41\,(10)$ & $54.38\,(8)$ & $62.27\,(6)$ & $45.78\,(12)$ & $64.09\,(4)$ & $64.29\,(3)$ & $\underline{\mathbf{72.39\,(1)}}$ & $71.81\,(2)$ & $60.25\,(7)$ & $53.33\,(9)$ & $63.35\,(5)$ \\
    FordB\_1 & $50.31\,(5)$ & $\underline{\mathbf{58.66\,(1)}}$ & $57.57\,(3)$ & $44.69\,(11)$ & $48.90\,(9)$ & $51.28\,(4)$ & $50.27\,(6)$ & $47.15\,(10)$ & $40.12\,(12)$ & $49.11\,(8)$ & $57.91\,(2)$ & $49.26\,(7)$ \\
    HandMovementDirection\_0 & $50.67\,(9)$ & $55.37\,(2)$ & $53.11\,(5)$ & $52.61\,(6)$ & $45.67\,(12)$ & $52.05\,(7)$ & $54.64\,(3)$ & $49.65\,(11)$ & $51.93\,(8)$ & $49.90\,(10)$ & $\underline{\mathbf{56.38\,(1)}}$ & $53.29\,(4)$ \\
    HandMovementDirection\_1 & $43.26\,(7)$ & $45.00\,(5)$ & $43.03\,(9)$ & $39.49\,(12)$ & $49.67\,(3)$ & $52.15\,(2)$ & $46.24\,(4)$ & $\underline{\mathbf{56.95\,(1)}}$ & $43.45\,(6)$ & $43.03\,(10)$ & $43.14\,(8)$ & $41.65\,(11)$ \\
    HandMovementDirection\_2 & $56.23\,(8)$ & $56.38\,(7)$ & $57.85\,(4)$ & $\underline{\mathbf{66.08\,(1)}}$ & $53.97\,(9)$ & $52.54\,(10)$ & $35.95\,(12)$ & $40.63\,(11)$ & $60.88\,(3)$ & $62.17\,(2)$ & $57.72\,(5)$ & $56.90\,(6)$ \\
    HandMovementDirection\_3 & $52.98\,(8)$ & $51.55\,(10)$ & $54.40\,(6)$ & $\underline{\mathbf{59.93\,(1)}}$ & $53.17\,(7)$ & $54.67\,(5)$ & $51.52\,(11)$ & $51.17\,(12)$ & $56.48\,(3)$ & $51.64\,(9)$ & $55.62\,(4)$ & $59.17\,(2)$ \\
    Heartbeat\_0 & $39.67\,(7)$ & $37.34\,(11)$ & $36.96\,(12)$ & $40.33\,(6)$ & $52.36\,(2)$ & $52.15\,(3)$ & $\underline{\mathbf{57.38\,(1)}}$ & $47.80\,(4)$ & $38.13\,(10)$ & $38.21\,(9)$ & $39.11\,(8)$ & $41.55\,(5)$ \\
    Heartbeat\_1 & $63.25\,(6)$ & $64.91\,(3)$ & $\underline{\mathbf{65.55\,(1)}}$ & $60.90\,(7)$ & $44.83\,(12)$ & $57.49\,(9)$ & $51.59\,(11)$ & $53.78\,(10)$ & $64.91\,(2)$ & $64.02\,(5)$ & $64.35\,(4)$ & $60.82\,(8)$ \\
    MotorImagery\_0 & $58.16\,(6)$ & $58.16\,(7)$ & $60.70\,(2)$ & $58.54\,(5)$ & $50.00\,(10)$ & $57.18\,(8)$ & $43.96\,(11)$ & $43.78\,(12)$ & $57.08\,(9)$ & $59.74\,(3)$ & $58.63\,(4)$ & $\underline{\mathbf{60.89\,(1)}}$ \\
    MotorImagery\_1 & $40.65\,(10)$ & $41.76\,(8)$ & $40.04\,(11)$ & $43.10\,(6)$ & $49.24\,(3)$ & $44.19\,(5)$ & $52.23\,(2)$ & $\underline{\mathbf{56.02\,(1)}}$ & $44.81\,(4)$ & $40.82\,(9)$ & $42.92\,(7)$ & $39.17\,(12)$ \\
    NATOPS\_0 & $42.50\,(10)$ & $52.60\,(6)$ & $44.13\,(9)$ & $52.39\,(7)$ & $41.27\,(11)$ & $71.49\,(3)$ & $75.86\,(2)$ & $\underline{\mathbf{77.10\,(1)}}$ & $68.38\,(4)$ & $32.68\,(12)$ & $46.06\,(8)$ & $59.84\,(5)$ \\
    NATOPS\_1 & $24.59\,(10)$ & $27.58\,(9)$ & $30.49\,(8)$ & $33.35\,(7)$ & $48.36\,(5)$ & $51.95\,(3)$ & $\underline{\mathbf{53.74\,(1)}}$ & $53.63\,(2)$ & $49.75\,(4)$ & $20.81\,(12)$ & $23.84\,(11)$ & $37.71\,(6)$ \\
    NATOPS\_2 & $19.33\,(10)$ & $19.87\,(9)$ & $26.04\,(8)$ & $26.43\,(7)$ & $50.85\,(3)$ & $45.85\,(5)$ & $\underline{\mathbf{54.49\,(1)}}$ & $53.46\,(2)$ & $46.87\,(4)$ & $12.94\,(12)$ & $17.28\,(11)$ & $32.70\,(6)$ \\
    NATOPS\_3 & $94.82\,(7)$ & $98.93\,(3)$ & $97.69\,(4)$ & $95.06\,(5)$ & $44.08\,(12)$ & $92.12\,(8)$ & $94.84\,(6)$ & $64.17\,(11)$ & $91.64\,(10)$ & $91.67\,(9)$ & $\underline{\mathbf{100.00\,(1)}}$ & $99.21\,(2)$ \\
    NATOPS\_4 & $90.53\,(6)$ & $95.67\,(2)$ & $93.20\,(5)$ & $87.89\,(8)$ & $40.32\,(12)$ & $83.74\,(9)$ & $95.48\,(3)$ & $63.58\,(11)$ & $80.52\,(10)$ & $88.23\,(7)$ & $\underline{\mathbf{99.07\,(1)}}$ & $94.16\,(4)$ \\
    NATOPS\_5 & $60.71\,(8)$ & $80.29\,(5)$ & $54.62\,(10)$ & $73.75\,(6)$ & $46.69\,(12)$ & $73.23\,(7)$ & $\underline{\mathbf{98.20\,(1)}}$ & $87.69\,(3)$ & $49.52\,(11)$ & $59.82\,(9)$ & $83.15\,(4)$ & $89.38\,(2)$ \\
    PEMS-SF\_0 & $\underline{\mathbf{92.32\,(1)}}$ & $60.28\,(10)$ & $82.24\,(5)$ & $55.04\,(11)$ & $75.58\,(6)$ & $82.70\,(4)$ & $90.65\,(2)$ & $83.51\,(3)$ & $72.30\,(8)$ & $51.50\,(12)$ & $72.52\,(7)$ & $64.87\,(9)$ \\
    PEMS-SF\_1 & $43.20\,(12)$ & $51.57\,(11)$ & $53.95\,(10)$ & $61.30\,(8)$ & $67.54\,(5)$ & $72.43\,(3)$ & $\underline{\mathbf{82.99\,(1)}}$ & $77.61\,(2)$ & $61.22\,(9)$ & $67.57\,(4)$ & $63.88\,(6)$ & $63.04\,(7)$ \\
    PEMS-SF\_2 & $40.05\,(12)$ & $63.32\,(10)$ & $68.45\,(9)$ & $74.68\,(4)$ & $69.25\,(7)$ & $77.76\,(2)$ & $60.61\,(11)$ & $68.88\,(8)$ & $75.35\,(3)$ & $\underline{\mathbf{77.86\,(1)}}$ & $73.79\,(6)$ & $74.34\,(5)$ \\
    PEMS-SF\_3 & $39.91\,(12)$ & $61.45\,(8)$ & $64.00\,(7)$ & $74.45\,(3)$ & $57.84\,(10)$ & $72.54\,(6)$ & $46.43\,(11)$ & $60.43\,(9)$ & $75.73\,(2)$ & $\underline{\mathbf{81.32\,(1)}}$ & $73.52\,(4)$ & $73.34\,(5)$ \\
    PEMS-SF\_4 & $51.74\,(10)$ & $61.80\,(8)$ & $68.90\,(7)$ & $74.12\,(4)$ & $50.78\,(12)$ & $73.44\,(5)$ & $51.36\,(11)$ & $56.71\,(9)$ & $72.09\,(6)$ & $\underline{\mathbf{77.14\,(1)}}$ & $74.34\,(2)$ & $74.15\,(3)$ \\
    PEMS-SF\_5 & $74.60\,(10)$ & $89.12\,(4)$ & $88.76\,(5)$ & $88.13\,(6)$ & $59.39\,(12)$ & $84.81\,(9)$ & $62.89\,(11)$ & $86.26\,(7)$ & $85.82\,(8)$ & $90.05\,(2)$ & $\underline{\mathbf{90.98\,(1)}}$ & $89.61\,(3)$ \\
    PEMS-SF\_6 & $73.16\,(3)$ & $71.31\,(6)$ & $67.25\,(8)$ & $60.56\,(11)$ & $60.99\,(10)$ & $73.12\,(4)$ & $76.11\,(2)$ & $60.03\,(12)$ & $71.61\,(5)$ & $63.86\,(9)$ & $\underline{\mathbf{77.00\,(1)}}$ & $67.54\,(7)$ \\
    PenDigits\_0 & $97.16\,(5)$ & $99.71\,(2)$ & $97.74\,(4)$ & $87.27\,(8)$ & $62.81\,(11)$ & $70.01\,(9)$ & $96.91\,(6)$ & $67.96\,(10)$ & $92.70\,(7)$ & $60.53\,(12)$ & $99.18\,(3)$ & $\underline{\mathbf{99.83\,(1)}}$ \\
    PenDigits\_1 & $71.73\,(8)$ & $\underline{\mathbf{95.07\,(1)}}$ & $80.37\,(5)$ & $80.77\,(4)$ & $47.31\,(12)$ & $63.23\,(10)$ & $74.69\,(7)$ & $70.78\,(9)$ & $78.96\,(6)$ & $62.39\,(11)$ & $92.88\,(2)$ & $91.91\,(3)$ \\
    PenDigits\_2 & $71.59\,(5)$ & $87.95\,(3)$ & $60.44\,(8)$ & $50.84\,(10)$ & $54.47\,(9)$ & $43.35\,(11)$ & $72.63\,(4)$ & $61.54\,(7)$ & $66.54\,(6)$ & $33.42\,(12)$ & $88.71\,(2)$ & $\underline{\mathbf{90.01\,(1)}}$ \\
    PenDigits\_3 & $48.74\,(8)$ & $95.41\,(2)$ & $23.49\,(12)$ & $38.95\,(10)$ & $64.79\,(6)$ & $44.48\,(9)$ & $\underline{\mathbf{96.14\,(1)}}$ & $62.33\,(7)$ & $38.92\,(11)$ & $82.13\,(5)$ & $92.28\,(4)$ & $94.75\,(3)$ \\
    PenDigits\_4 & $79.27\,(6)$ & $97.30\,(4)$ & $58.15\,(11)$ & $54.54\,(12)$ & $66.42\,(8)$ & $65.91\,(9)$ & $\underline{\mathbf{98.74\,(1)}}$ & $87.11\,(5)$ & $58.68\,(10)$ & $74.97\,(7)$ & $97.74\,(3)$ & $98.58\,(2)$ \\
    PenDigits\_5 & $77.37\,(6)$ & $\underline{\mathbf{97.92\,(1)}}$ & $74.94\,(9)$ & $77.73\,(5)$ & $70.05\,(10)$ & $54.54\,(12)$ & $91.59\,(4)$ & $75.57\,(7)$ & $75.09\,(8)$ & $69.64\,(11)$ & $94.66\,(3)$ & $96.20\,(2)$ \\
    PenDigits\_6 & $91.41\,(5)$ & $\underline{\mathbf{99.42\,(1)}}$ & $86.48\,(7)$ & $82.11\,(8)$ & $78.10\,(10)$ & $58.77\,(11)$ & $97.73\,(3)$ & $88.32\,(6)$ & $81.20\,(9)$ & $55.79\,(12)$ & $96.99\,(4)$ & $98.58\,(2)$ \\
    PenDigits\_7 & $71.41\,(5)$ & $\underline{\mathbf{98.92\,(1)}}$ & $50.68\,(8)$ & $49.05\,(10)$ & $71.00\,(6)$ & $41.28\,(12)$ & $92.74\,(3)$ & $70.32\,(7)$ & $50.04\,(9)$ & $47.92\,(11)$ & $92.69\,(4)$ & $96.30\,(2)$ \\
    PenDigits\_8 & $95.71\,(5)$ & $\underline{\mathbf{99.75\,(1)}}$ & $94.06\,(6)$ & $84.25\,(8)$ & $63.55\,(10)$ & $52.16\,(12)$ & $99.15\,(2)$ & $58.43\,(11)$ & $79.72\,(9)$ & $86.63\,(7)$ & $98.56\,(4)$ & $99.09\,(3)$ \\
    PenDigits\_9 & $79.66\,(9)$ & $\underline{\mathbf{99.07\,(1)}}$ & $88.15\,(6)$ & $85.02\,(7)$ & $59.94\,(10)$ & $52.09\,(11)$ & $95.73\,(4)$ & $81.19\,(8)$ & $51.92\,(12)$ & $92.12\,(5)$ & $98.10\,(2)$ & $98.08\,(3)$ \\
  \bottomrule
  \end{tabular}
}
}
\end{table*}

\begin{table*}[h!]
\centering
\scriptsize
\centering
\caption{Mean AUPR scores and rankings for all problems under the \textit{(N-1)-vs-rest} setting.}
\label{tab:datasets_pr_rev1}
{
\resizebox{\linewidth}{!}{%
  \begin{tabular}{lcccccccccccc}
  \toprule
    Dataset\_Normality & IF & LOF & OCSVM & AutoAD & DAGMM & DeepSVDD & FixedTS & NeuTraL AD & TimesURL & MOMENT & TSINR & NeuCoReClass AD \\
    \midrule
 ChlorineConcentration\_0 & $48.39\,(10)$ & $48.57\,(9)$ & $57.80\,(2)$ & $49.26\,(8)$ & $34.27\,(12)$ & $51.66\,(5)$ & $41.70\,(11)$ & $50.45\,(7)$ & $50.78\,(6)$ & $55.20\,(4)$ & $\underline{\mathbf{62.68\,(1)}}$ & $57.32\,(3)$ \\
    ChlorineConcentration\_1 & $21.37\,(9)$ & $20.74\,(12)$ & $20.87\,(11)$ & $20.99\,(10)$ & $28.39\,(2)$ & $22.16\,(7)$ & $24.48\,(4)$ & $24.38\,(5)$ & $22.31\,(6)$ & $21.92\,(8)$ & $\underline{\mathbf{37.34\,(1)}}$ & $26.29\,(3)$ \\
    ChlorineConcentration\_2 & $47.75\,(9)$ & $47.67\,(10)$ & $46.89\,(11)$ & $49.01\,(7)$ & $51.44\,(4)$ & $46.83\,(12)$ & $50.11\,(6)$ & $52.04\,(3)$ & $48.96\,(8)$ & $50.56\,(5)$ & $\underline{\mathbf{70.08\,(1)}}$ & $58.15\,(2)$ \\
    Computers\_0 & $\underline{\mathbf{74.13\,(1)}}$ & $49.09\,(11)$ & $51.46\,(8)$ & $51.75\,(7)$ & $52.38\,(6)$ & $56.88\,(4)$ & $61.09\,(2)$ & $58.81\,(3)$ & $51.01\,(9)$ & $49.31\,(10)$ & $47.37\,(12)$ & $54.05\,(5)$ \\
    Computers\_1 & $37.67\,(12)$ & $57.70\,(6)$ & $54.30\,(8)$ & $\underline{\mathbf{64.18\,(1)}}$ & $51.94\,(10)$ & $52.43\,(9)$ & $40.34\,(11)$ & $58.01\,(5)$ & $61.76\,(2)$ & $61.02\,(3)$ & $55.18\,(7)$ & $59.46\,(4)$ \\
    DistalPhalanxOutlineAgeGroup\_0 & $29.85\,(10)$ & $65.13\,(2)$ & $32.58\,(8)$ & $26.92\,(12)$ & $29.92\,(9)$ & $33.98\,(7)$ & $58.25\,(3)$ & $\underline{\mathbf{65.33\,(1)}}$ & $27.52\,(11)$ & $36.38\,(6)$ & $40.59\,(5)$ & $53.82\,(4)$ \\
    DistalPhalanxOutlineAgeGroup\_1 & $53.82\,(7)$ & $38.36\,(12)$ & $55.24\,(5)$ & $56.87\,(3)$ & $52.02\,(9)$ & $\underline{\mathbf{57.49\,(1)}}$ & $44.31\,(11)$ & $49.90\,(10)$ & $56.34\,(4)$ & $55.19\,(6)$ & $56.93\,(2)$ & $53.74\,(8)$ \\
    DistalPhalanxOutlineAgeGroup\_2 & $\underline{\mathbf{69.07\,(1)}}$ & $55.75\,(5)$ & $57.11\,(4)$ & $64.08\,(3)$ & $46.97\,(6)$ & $45.21\,(7)$ & $31.97\,(12)$ & $35.56\,(8)$ & $66.84\,(2)$ & $34.47\,(10)$ & $32.75\,(11)$ & $34.53\,(9)$ \\
    DistalPhalanxOutlineCorrect\_0 & $64.17\,(6)$ & $\underline{\mathbf{78.23\,(1)}}$ & $67.39\,(3)$ & $63.03\,(9)$ & $53.69\,(12)$ & $66.69\,(4)$ & $61.38\,(11)$ & $65.13\,(5)$ & $61.73\,(10)$ & $64.10\,(7)$ & $63.81\,(8)$ & $68.00\,(2)$ \\
    DistalPhalanxOutlineCorrect\_1 & $55.74\,(4)$ & $49.14\,(11)$ & $52.83\,(8)$ & $\underline{\mathbf{57.72\,(1)}}$ & $56.82\,(2)$ & $53.26\,(7)$ & $46.34\,(12)$ & $53.60\,(5)$ & $56.22\,(3)$ & $53.44\,(6)$ & $52.52\,(9)$ & $50.93\,(10)$ \\
    ECG200\_0 & $79.70\,(7)$ & $82.24\,(4)$ & $81.23\,(5)$ & $77.48\,(8)$ & $83.52\,(2)$ & $\underline{\mathbf{84.10\,(1)}}$ & $61.50\,(11)$ & $63.25\,(10)$ & $76.19\,(9)$ & $56.20\,(12)$ & $82.94\,(3)$ & $80.28\,(6)$ \\
    ECG200\_1 & $60.18\,(8)$ & $47.97\,(12)$ & $61.89\,(5)$ & $59.10\,(9)$ & $57.16\,(10)$ & $60.76\,(7)$ & $67.61\,(2)$ & $62.56\,(3)$ & $56.93\,(11)$ & $\underline{\mathbf{73.51\,(1)}}$ & $61.99\,(4)$ & $61.67\,(6)$ \\
    Earthquakes\_0 & $70.40\,(12)$ & $81.50\,(8)$ & $78.18\,(10)$ & $86.24\,(5)$ & $83.82\,(6)$ & $87.66\,(2)$ & $76.94\,(11)$ & $87.42\,(3)$ & $86.68\,(4)$ & $\underline{\mathbf{88.95\,(1)}}$ & $79.29\,(9)$ & $83.19\,(7)$ \\
    Earthquakes\_1 & $33.07\,(6)$ & $34.70\,(2)$ & $34.49\,(3)$ & $20.72\,(11)$ & $29.72\,(7)$ & $23.81\,(9)$ & $\underline{\mathbf{37.11\,(1)}}$ & $22.48\,(10)$ & $26.16\,(8)$ & $20.36\,(12)$ & $33.27\,(5)$ & $33.89\,(4)$ \\
    ElectricDevices\_0 & $8.56\,(10)$ & $10.09\,(7)$ & $9.99\,(8)$ & $14.38\,(2)$ & $12.04\,(6)$ & $12.26\,(5)$ & $6.37\,(12)$ & $14.07\,(3)$ & $\underline{\mathbf{18.77\,(1)}}$ & $9.44\,(9)$ & $7.93\,(11)$ & $13.36\,(4)$ \\
    ElectricDevices\_1 & $64.51\,(5)$ & $\underline{\mathbf{73.27\,(1)}}$ & $51.08\,(10)$ & $21.18\,(12)$ & $36.90\,(11)$ & $72.29\,(2)$ & $51.65\,(9)$ & $53.80\,(7)$ & $52.01\,(8)$ & $58.58\,(6)$ & $66.62\,(3)$ & $65.47\,(4)$ \\
    ElectricDevices\_2 & $6.54\,(11)$ & $13.53\,(4)$ & $8.76\,(7)$ & $21.94\,(2)$ & $\underline{\mathbf{23.71\,(1)}}$ & $7.92\,(8)$ & $7.25\,(9)$ & $7.18\,(10)$ & $6.31\,(12)$ & $17.56\,(3)$ & $10.37\,(6)$ & $10.37\,(5)$ \\
    ElectricDevices\_3 & $8.95\,(12)$ & $10.03\,(9)$ & $10.43\,(8)$ & $\underline{\mathbf{30.64\,(1)}}$ & $16.68\,(2)$ & $13.36\,(3)$ & $9.26\,(11)$ & $11.27\,(5)$ & $13.19\,(4)$ & $10.97\,(6)$ & $9.59\,(10)$ & $10.81\,(7)$ \\
    ElectricDevices\_4 & $28.85\,(8)$ & $27.18\,(10)$ & $25.57\,(11)$ & $16.32\,(12)$ & $38.32\,(4)$ & $38.14\,(5)$ & $\underline{\mathbf{49.60\,(1)}}$ & $39.57\,(2)$ & $35.16\,(6)$ & $27.83\,(9)$ & $29.07\,(7)$ & $39.24\,(3)$ \\
    ElectricDevices\_5 & $5.64\,(12)$ & $6.45\,(11)$ & $14.14\,(2)$ & $\underline{\mathbf{33.01\,(1)}}$ & $10.44\,(5)$ & $8.82\,(9)$ & $6.67\,(10)$ & $9.72\,(8)$ & $11.67\,(4)$ & $10.24\,(6)$ & $10.15\,(7)$ & $13.40\,(3)$ \\
    ElectricDevices\_6 & $8.12\,(7)$ & $8.82\,(5)$ & $6.16\,(9)$ & $4.94\,(12)$ & $8.99\,(3)$ & $6.47\,(8)$ & $\underline{\mathbf{23.66\,(1)}}$ & $8.84\,(4)$ & $5.97\,(10)$ & $5.22\,(11)$ & $8.72\,(6)$ & $9.80\,(2)$ \\
    Epilepsy\_0 & $20.37\,(11)$ & $21.33\,(9)$ & $21.29\,(10)$ & $27.28\,(7)$ & $38.02\,(5)$ & $\underline{\mathbf{63.88\,(1)}}$ & $49.08\,(3)$ & $37.78\,(6)$ & $16.30\,(12)$ & $56.72\,(2)$ & $25.05\,(8)$ & $43.52\,(4)$ \\
    Epilepsy\_1 & $92.05\,(9)$ & $98.27\,(6)$ & $99.02\,(5)$ & $99.68\,(3)$ & $91.79\,(10)$ & $99.74\,(2)$ & $47.95\,(12)$ & $59.59\,(11)$ & $\underline{\mathbf{99.90\,(1)}}$ & $95.25\,(8)$ & $96.37\,(7)$ & $99.23\,(4)$ \\
    Epilepsy\_2 & $23.54\,(8)$ & $19.37\,(10)$ & $22.31\,(9)$ & $24.53\,(7)$ & $35.26\,(5)$ & $45.47\,(3)$ & $\underline{\mathbf{53.69\,(1)}}$ & $44.74\,(4)$ & $33.28\,(6)$ & $14.93\,(12)$ & $18.47\,(11)$ & $53.28\,(2)$ \\
    Epilepsy\_3 & $17.18\,(10)$ & $19.92\,(7)$ & $15.73\,(12)$ & $17.19\,(9)$ & $65.13\,(3)$ & $40.78\,(6)$ & $\underline{\mathbf{88.40\,(1)}}$ & $72.53\,(2)$ & $45.06\,(5)$ & $17.33\,(8)$ & $16.74\,(11)$ & $58.62\,(4)$ \\
    EthanolConcentration\_0 & $22.70\,(12)$ & $25.65\,(6)$ & $23.83\,(11)$ & $23.88\,(10)$ & $25.08\,(8)$ & $26.48\,(4)$ & $24.66\,(9)$ & $\underline{\mathbf{28.75\,(1)}}$ & $28.22\,(2)$ & $25.08\,(7)$ & $27.13\,(3)$ & $26.14\,(5)$ \\
    EthanolConcentration\_1 & $23.99\,(6)$ & $\underline{\mathbf{28.51\,(1)}}$ & $23.83\,(7)$ & $22.25\,(11)$ & $23.38\,(9)$ & $23.04\,(10)$ & $25.18\,(3)$ & $27.63\,(2)$ & $24.83\,(5)$ & $24.91\,(4)$ & $23.46\,(8)$ & $21.71\,(12)$ \\
    EthanolConcentration\_2 & $\underline{\mathbf{31.06\,(1)}}$ & $26.73\,(6)$ & $27.92\,(2)$ & $27.57\,(4)$ & $26.86\,(5)$ & $26.41\,(7)$ & $27.62\,(3)$ & $25.84\,(9)$ & $24.94\,(10)$ & $23.01\,(12)$ & $23.66\,(11)$ & $26.12\,(8)$ \\
    EthanolConcentration\_3 & $33.70\,(3)$ & $24.29\,(11)$ & $32.03\,(5)$ & $33.62\,(4)$ & $28.73\,(9)$ & $29.28\,(8)$ & $29.73\,(7)$ & $20.62\,(12)$ & $27.41\,(10)$ & $\underline{\mathbf{35.98\,(1)}}$ & $34.36\,(2)$ & $31.67\,(6)$ \\
    EthanolLevel\_0 & $\underline{\mathbf{30.94\,(1)}}$ & $28.98\,(6)$ & $29.76\,(4)$ & $28.30\,(7)$ & $27.78\,(8)$ & $30.87\,(2)$ & $24.24\,(12)$ & $26.97\,(10)$ & $25.74\,(11)$ & $30.18\,(3)$ & $27.22\,(9)$ & $28.98\,(5)$ \\
    EthanolLevel\_1 & $23.41\,(8)$ & $\underline{\mathbf{25.07\,(1)}}$ & $23.82\,(5)$ & $24.01\,(4)$ & $24.40\,(2)$ & $23.29\,(9)$ & $22.84\,(12)$ & $23.05\,(10)$ & $23.41\,(7)$ & $23.71\,(6)$ & $24.33\,(3)$ & $23.03\,(11)$ \\
    EthanolLevel\_2 & $22.28\,(12)$ & $23.15\,(10)$ & $22.45\,(11)$ & $24.28\,(7)$ & $27.18\,(2)$ & $24.38\,(5)$ & $\underline{\mathbf{28.47\,(1)}}$ & $26.50\,(3)$ & $25.23\,(4)$ & $23.54\,(9)$ & $24.09\,(8)$ & $24.31\,(6)$ \\
    EthanolLevel\_3 & $29.21\,(4)$ & $27.70\,(8)$ & $28.72\,(5)$ & $28.53\,(6)$ & $24.18\,(11)$ & $23.97\,(12)$ & $34.27\,(2)$ & $27.81\,(7)$ & $26.63\,(10)$ & $26.70\,(9)$ & $\underline{\mathbf{36.49\,(1)}}$ & $33.95\,(3)$ \\
    FaceDetection\_0 & $49.02\,(8)$ & $48.74\,(11)$ & $48.92\,(9)$ & $49.40\,(4)$ & $\underline{\mathbf{51.65\,(1)}}$ & $48.60\,(12)$ & $49.16\,(5)$ & $50.15\,(2)$ & $49.43\,(3)$ & $49.16\,(6)$ & $48.74\,(10)$ & $49.15\,(7)$ \\
    FaceDetection\_1 & $51.64\,(4)$ & $51.86\,(2)$ & $51.67\,(3)$ & $50.94\,(9)$ & $49.95\,(12)$ & $51.07\,(8)$ & $50.01\,(11)$ & $50.37\,(10)$ & $51.28\,(6)$ & $51.11\,(7)$ & $\underline{\mathbf{51.91\,(1)}}$ & $51.42\,(5)$ \\
    FingerMovements\_0 & $40.72\,(10)$ & $41.64\,(6)$ & $40.39\,(11)$ & $41.43\,(7)$ & $\underline{\mathbf{56.70\,(1)}}$ & $40.37\,(12)$ & $48.71\,(3)$ & $50.21\,(2)$ & $40.79\,(9)$ & $41.29\,(8)$ & $43.80\,(4)$ & $42.84\,(5)$ \\
    FingerMovements\_1 & $\underline{\mathbf{64.68\,(1)}}$ & $61.86\,(8)$ & $63.03\,(4)$ & $62.92\,(5)$ & $62.37\,(7)$ & $59.90\,(9)$ & $50.73\,(11)$ & $48.72\,(12)$ & $62.49\,(6)$ & $64.62\,(2)$ & $63.14\,(3)$ & $58.09\,(10)$ \\
    FordA\_0 & $53.62\,(12)$ & $70.87\,(8)$ & $68.95\,(10)$ & $78.78\,(5)$ & $59.98\,(11)$ & $\underline{\mathbf{81.86\,(1)}}$ & $69.13\,(9)$ & $80.79\,(3)$ & $78.25\,(6)$ & $78.98\,(4)$ & $77.60\,(7)$ & $80.90\,(2)$ \\
    FordA\_1 & $49.28\,(3)$ & $48.01\,(5)$ & $48.28\,(4)$ & $38.68\,(12)$ & $47.11\,(6)$ & $\underline{\mathbf{57.84\,(1)}}$ & $44.84\,(7)$ & $49.85\,(2)$ & $43.34\,(9)$ & $40.82\,(10)$ & $43.46\,(8)$ & $40.64\,(11)$ \\
    FordB\_0 & $49.93\,(11)$ & $58.69\,(9)$ & $59.90\,(7)$ & $66.78\,(5)$ & $47.54\,(12)$ & $68.67\,(4)$ & $56.27\,(10)$ & $\underline{\mathbf{73.88\,(1)}}$ & $69.29\,(3)$ & $66.08\,(6)$ & $59.51\,(8)$ & $69.83\,(2)$ \\
    FordB\_1 & $51.37\,(5)$ & $53.80\,(3)$ & $\underline{\mathbf{54.87\,(1)}}$ & $44.34\,(11)$ & $53.97\,(2)$ & $49.87\,(7)$ & $49.98\,(6)$ & $47.12\,(8)$ & $42.55\,(12)$ & $47.04\,(9)$ & $52.75\,(4)$ & $45.90\,(10)$ \\
    HandMovementDirection\_0 & $20.33\,(11)$ & $21.32\,(8)$ & $20.43\,(10)$ & $22.09\,(5)$ & $\underline{\mathbf{36.90\,(1)}}$ & $23.40\,(4)$ & $24.82\,(3)$ & $24.83\,(2)$ & $20.83\,(9)$ & $19.39\,(12)$ & $22.01\,(6)$ & $21.66\,(7)$ \\
    HandMovementDirection\_1 & $35.85\,(12)$ & $38.32\,(6)$ & $38.33\,(5)$ & $37.14\,(9)$ & $\underline{\mathbf{57.37\,(1)}}$ & $45.89\,(3)$ & $37.88\,(7)$ & $47.82\,(2)$ & $38.63\,(4)$ & $36.11\,(11)$ & $37.77\,(8)$ & $36.79\,(10)$ \\
    HandMovementDirection\_2 & $31.37\,(7)$ & $32.28\,(5)$ & $32.83\,(4)$ & $37.83\,(2)$ & $\underline{\mathbf{41.98\,(1)}}$ & $21.13\,(10)$ & $17.41\,(11)$ & $16.62\,(12)$ & $33.98\,(3)$ & $31.83\,(6)$ & $29.47\,(8)$ & $23.16\,(9)$ \\
    HandMovementDirection\_3 & $20.70\,(6)$ & $18.81\,(12)$ & $19.42\,(10)$ & $22.05\,(4)$ & $\underline{\mathbf{42.51\,(1)}}$ & $20.66\,(7)$ & $20.21\,(8)$ & $19.34\,(11)$ & $21.20\,(5)$ & $26.53\,(2)$ & $20.03\,(9)$ & $24.68\,(3)$ \\
    Heartbeat\_0 & $66.64\,(9)$ & $66.17\,(11)$ & $66.20\,(10)$ & $67.29\,(6)$ & $\underline{\mathbf{81.50\,(1)}}$ & $74.87\,(3)$ & $77.30\,(2)$ & $71.18\,(4)$ & $67.04\,(7)$ & $65.93\,(12)$ & $66.66\,(8)$ & $67.54\,(5)$ \\
    Heartbeat\_1 & $39.30\,(3)$ & $38.94\,(5)$ & $\underline{\mathbf{39.91\,(1)}}$ & $35.93\,(8)$ & $34.43\,(9)$ & $32.78\,(10)$ & $29.57\,(11)$ & $28.80\,(12)$ & $37.77\,(6)$ & $39.13\,(4)$ & $39.65\,(2)$ & $36.69\,(7)$ \\
    MotorImagery\_0 & $52.51\,(10)$ & $53.82\,(9)$ & $56.46\,(3)$ & $55.90\,(4)$ & $\underline{\mathbf{75.00\,(1)}}$ & $54.56\,(7)$ & $45.41\,(12)$ & $47.28\,(11)$ & $56.71\,(2)$ & $54.65\,(5)$ & $54.30\,(8)$ & $54.62\,(6)$ \\
    MotorImagery\_1 & $48.21\,(5)$ & $47.16\,(9)$ & $46.45\,(11)$ & $46.65\,(10)$ & $\underline{\mathbf{69.29\,(1)}}$ & $47.84\,(6)$ & $52.02\,(3)$ & $56.41\,(2)$ & $49.93\,(4)$ & $47.17\,(8)$ & $47.73\,(7)$ & $46.29\,(12)$ \\
    NATOPS\_0 & $13.31\,(11)$ & $16.79\,(7)$ & $13.68\,(10)$ & $16.09\,(8)$ & $22.72\,(5)$ & $32.02\,(3)$ & $42.06\,(2)$ & $\underline{\mathbf{45.85\,(1)}}$ & $24.22\,(4)$ & $11.73\,(12)$ & $13.96\,(9)$ & $19.28\,(6)$ \\
    NATOPS\_1 & $10.48\,(11)$ & $11.11\,(9)$ & $11.21\,(8)$ & $15.69\,(6)$ & $18.10\,(3)$ & $\underline{\mathbf{20.34\,(1)}}$ & $17.79\,(4)$ & $20.19\,(2)$ & $16.01\,(5)$ & $10.16\,(12)$ & $10.49\,(10)$ & $14.94\,(7)$ \\
    NATOPS\_2 & $10.00\,(10)$ & $10.01\,(9)$ & $10.70\,(8)$ & $10.70\,(7)$ & $17.42\,(2)$ & $14.41\,(4)$ & $17.36\,(3)$ & $\underline{\mathbf{17.55\,(1)}}$ & $14.23\,(5)$ & $9.48\,(12)$ & $10.00\,(10)$ & $11.69\,(6)$ \\
    NATOPS\_3 & $68.30\,(6)$ & $94.31\,(3)$ & $87.42\,(4)$ & $74.50\,(5)$ & $34.09\,(11)$ & $67.25\,(7)$ & $66.48\,(8)$ & $25.73\,(12)$ & $62.92\,(9)$ & $59.16\,(10)$ & $\underline{\mathbf{99.98\,(1)}}$ & $96.18\,(2)$ \\
    NATOPS\_4 & $54.26\,(6)$ & $80.84\,(3)$ & $65.76\,(5)$ & $52.86\,(7)$ & $22.79\,(11)$ & $46.38\,(9)$ & $72.70\,(4)$ & $21.16\,(12)$ & $42.44\,(10)$ & $46.62\,(8)$ & $\underline{\mathbf{95.34\,(1)}}$ & $81.59\,(2)$ \\
    NATOPS\_5 & $18.27\,(9)$ & $30.70\,(5)$ & $16.23\,(11)$ & $25.64\,(7)$ & $18.80\,(8)$ & $26.56\,(6)$ & $\underline{\mathbf{92.25\,(1)}}$ & $49.81\,(2)$ & $15.02\,(12)$ & $17.91\,(10)$ & $33.93\,(4)$ & $49.33\,(3)$ \\
    PEMS-SF\_0 & $\underline{\mathbf{61.52\,(1)}}$ & $23.60\,(11)$ & $43.31\,(5)$ & $25.13\,(10)$ & $59.52\,(3)$ & $42.86\,(6)$ & $61.24\,(2)$ & $44.07\,(4)$ & $30.12\,(8)$ & $17.63\,(12)$ & $33.42\,(7)$ & $29.54\,(9)$ \\
    PEMS-SF\_1 & $12.36\,(12)$ & $19.50\,(9)$ & $18.82\,(11)$ & $24.44\,(6)$ & $51.51\,(2)$ & $30.92\,(4)$ & $\underline{\mathbf{51.91\,(1)}}$ & $35.22\,(3)$ & $18.98\,(10)$ & $22.57\,(8)$ & $25.57\,(5)$ & $24.11\,(7)$ \\
    PEMS-SF\_2 & $12.46\,(12)$ & $33.68\,(10)$ & $41.48\,(8)$ & $45.28\,(2)$ & $\underline{\mathbf{55.99\,(1)}}$ & $42.59\,(5)$ & $31.16\,(11)$ & $34.60\,(9)$ & $42.41\,(6)$ & $42.36\,(7)$ & $43.61\,(4)$ & $44.49\,(3)$ \\
    PEMS-SF\_3 & $10.55\,(12)$ & $23.06\,(10)$ & $24.95\,(8)$ & $33.26\,(4)$ & $\underline{\mathbf{51.85\,(1)}}$ & $28.23\,(7)$ & $13.32\,(11)$ & $23.21\,(9)$ & $37.67\,(2)$ & $36.01\,(3)$ & $30.27\,(6)$ & $32.38\,(5)$ \\
    PEMS-SF\_4 & $12.45\,(12)$ & $26.29\,(9)$ & $30.52\,(7)$ & $33.36\,(4)$ & $38.22\,(2)$ & $29.69\,(8)$ & $18.34\,(11)$ & $22.79\,(10)$ & $35.72\,(3)$ & $\underline{\mathbf{38.52\,(1)}}$ & $32.78\,(6)$ & $32.90\,(5)$ \\
    PEMS-SF\_5 & $37.18\,(11)$ & $57.37\,(3)$ & $57.74\,(2)$ & $53.99\,(6)$ & $49.90\,(7)$ & $43.97\,(9)$ & $31.45\,(12)$ & $42.40\,(10)$ & $48.78\,(8)$ & $56.43\,(4)$ & $\underline{\mathbf{59.02\,(1)}}$ & $56.10\,(5)$ \\
    PEMS-SF\_6 & $19.31\,(6)$ & $19.62\,(5)$ & $16.66\,(9)$ & $14.38\,(11)$ & $\underline{\mathbf{35.73\,(1)}}$ & $18.23\,(7)$ & $26.53\,(2)$ & $14.32\,(12)$ & $20.34\,(4)$ & $15.94\,(10)$ & $21.56\,(3)$ & $16.72\,(8)$ \\
    PenDigits\_0 & $75.40\,(6)$ & $95.44\,(2)$ & $80.26\,(4)$ & $36.94\,(8)$ & $20.37\,(10)$ & $29.09\,(9)$ & $76.85\,(5)$ & $20.24\,(11)$ & $63.82\,(7)$ & $11.50\,(12)$ & $90.25\,(3)$ & $\underline{\mathbf{97.40\,(1)}}$ \\
    PenDigits\_1 & $20.35\,(10)$ & $\underline{\mathbf{68.78\,(1)}}$ & $37.04\,(6)$ & $40.80\,(4)$ & $12.70\,(12)$ & $23.59\,(9)$ & $38.02\,(5)$ & $24.75\,(8)$ & $33.68\,(7)$ & $15.62\,(11)$ & $52.31\,(3)$ & $60.15\,(2)$ \\
    PenDigits\_2 & $15.65\,(5)$ & $\underline{\mathbf{40.56\,(1)}}$ & $12.08\,(9)$ & $11.20\,(10)$ & $13.72\,(7)$ & $9.57\,(11)$ & $23.58\,(4)$ & $13.28\,(8)$ & $15.37\,(6)$ & $7.15\,(12)$ & $35.84\,(3)$ & $39.83\,(2)$ \\
    PenDigits\_3 & $8.52\,(9)$ & $53.83\,(2)$ & $5.85\,(12)$ & $7.34\,(10)$ & $19.75\,(6)$ & $13.14\,(8)$ & $\underline{\mathbf{66.17\,(1)}}$ & $15.15\,(7)$ & $7.23\,(11)$ & $23.70\,(5)$ & $51.00\,(3)$ & $49.30\,(4)$ \\
    PenDigits\_4 & $21.15\,(8)$ & $79.66\,(3)$ & $12.57\,(11)$ & $10.88\,(12)$ & $25.09\,(6)$ & $22.85\,(7)$ & $\underline{\mathbf{89.47\,(1)}}$ & $45.64\,(5)$ & $12.93\,(10)$ & $18.35\,(9)$ & $77.26\,(4)$ & $80.85\,(2)$ \\
    PenDigits\_5 & $39.80\,(6)$ & $\underline{\mathbf{78.32\,(1)}}$ & $39.00\,(7)$ & $48.60\,(5)$ & $27.11\,(10)$ & $11.27\,(12)$ & $55.91\,(4)$ & $27.62\,(9)$ & $29.37\,(8)$ & $16.00\,(11)$ & $60.43\,(3)$ & $72.65\,(2)$ \\
    PenDigits\_6 & $39.03\,(6)$ & $\underline{\mathbf{92.89\,(1)}}$ & $28.13\,(8)$ & $24.82\,(10)$ & $35.49\,(7)$ & $13.70\,(11)$ & $80.41\,(2)$ & $48.79\,(5)$ & $25.71\,(9)$ & $9.66\,(12)$ & $75.46\,(4)$ & $78.09\,(3)$ \\
    PenDigits\_7 & $15.45\,(7)$ & $\underline{\mathbf{89.18\,(1)}}$ & $10.35\,(10)$ & $10.94\,(8)$ & $26.27\,(5)$ & $10.35\,(10)$ & $61.48\,(3)$ & $17.52\,(6)$ & $10.05\,(11)$ & $9.07\,(12)$ & $57.60\,(4)$ & $63.50\,(2)$ \\
    PenDigits\_8 & $65.74\,(5)$ & $\underline{\mathbf{96.98\,(1)}}$ & $56.51\,(6)$ & $39.52\,(8)$ & $19.61\,(11)$ & $20.22\,(10)$ & $85.70\,(3)$ & $16.58\,(12)$ & $35.67\,(9)$ & $55.21\,(7)$ & $86.18\,(2)$ & $83.77\,(4)$ \\
    PenDigits\_9 & $21.40\,(10)$ & $\underline{\mathbf{88.53\,(1)}}$ & $35.09\,(7)$ & $44.43\,(6)$ & $30.78\,(8)$ & $15.26\,(11)$ & $73.26\,(4)$ & $28.53\,(9)$ & $11.00\,(12)$ & $52.77\,(5)$ & $81.63\,(2)$ & $74.44\,(3)$ \\
  \bottomrule
  \end{tabular}
}
}
\end{table*}

\section{Statistical Assessment of the Results}
\label{app:statistical}

We apply the Friedman test over the results reported in Tables~\ref{tab:datasets_roc_rev0}, \ref{tab:datasets_pr_rev0}, \ref{tab:datasets_roc_rev1}, and \ref{tab:datasets_pr_rev1} to test the null hypothesis that the performance of all the methods is equal. As we can reject this hypothesis (p-value smaller than 0.05), we proceed with all the pairwise comparisons using a non-parametric approach (Wilcoxon signed-rank test) and we correct for multiple testing using Shaffer's correction.

\begin{figure*}[t]
    \centering
    
    \begin{subfigure}[t]{0.48\textwidth}
        \centering
        \includegraphics[width=\linewidth]{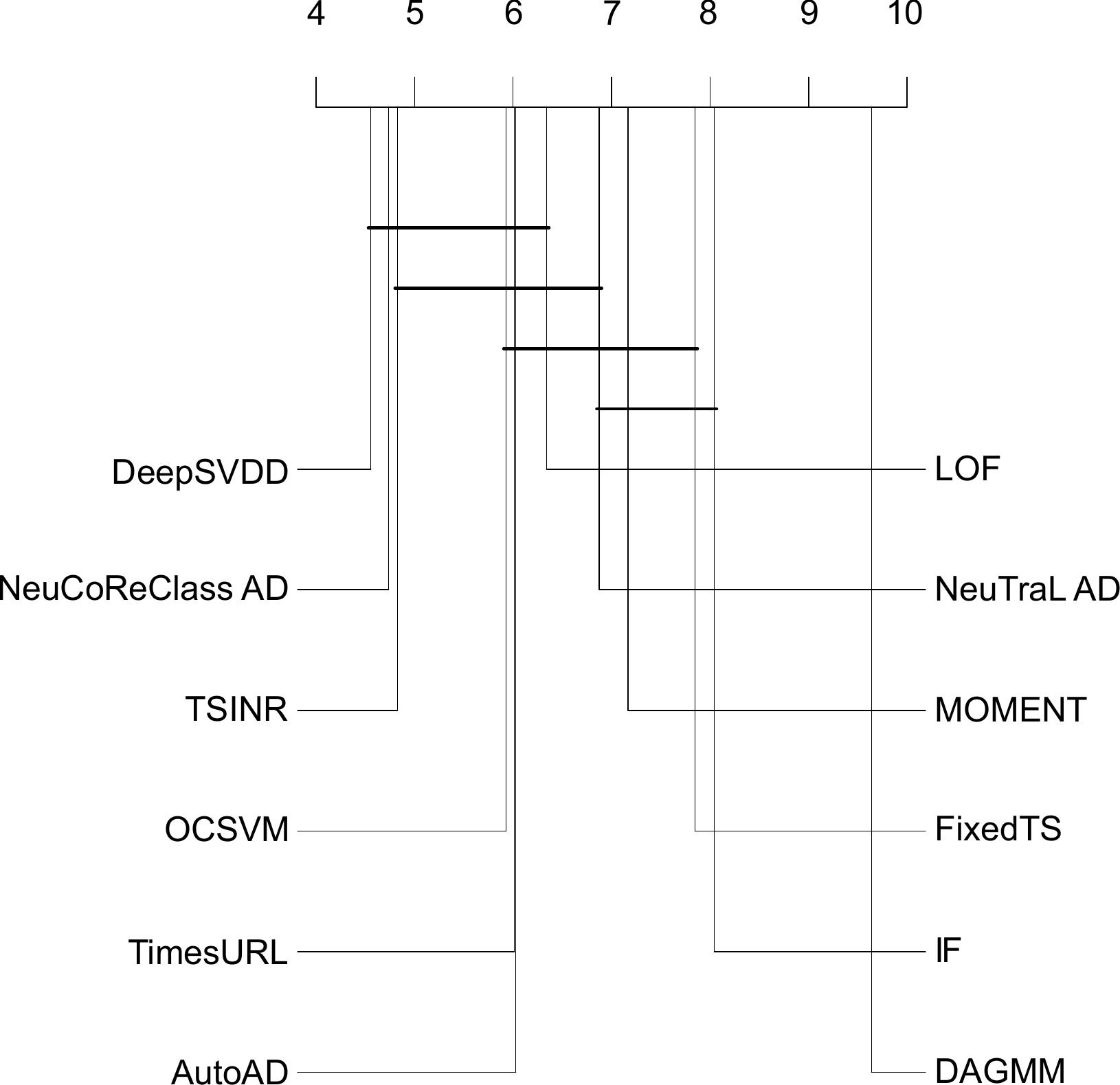}
        \caption{AUROC, \textit{one-vs-rest} setting.}
    \end{subfigure}
    \hfill
    \begin{subfigure}[t]{0.47\textwidth}
        \centering
        \includegraphics[width=\linewidth]{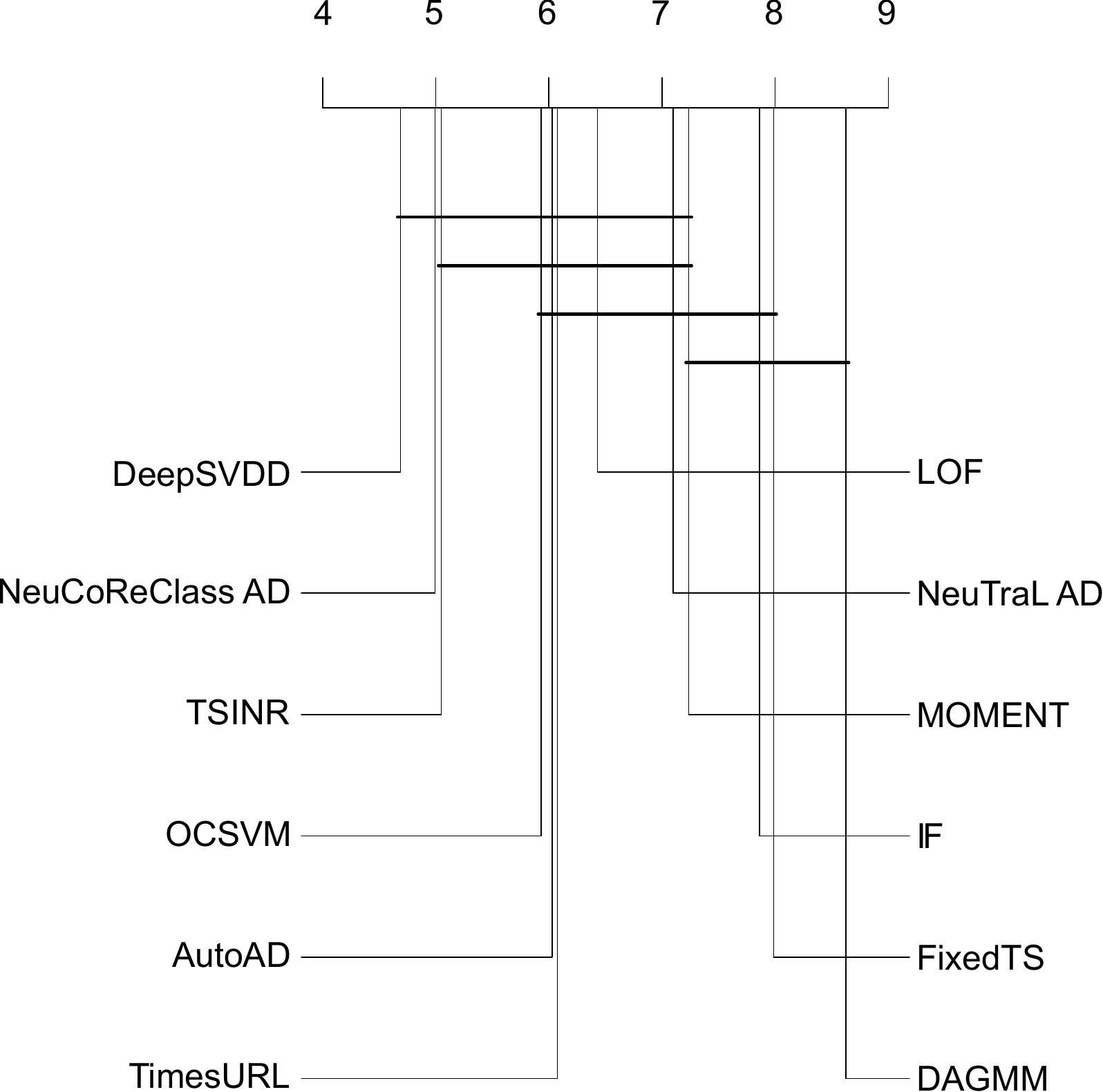}
        \caption{AUPR, \textit{one-vs-rest} setting.}
    \end{subfigure}

    \vspace{0.5em}

    \begin{subfigure}[t]{0.48\textwidth}
        \centering
        \includegraphics[width=\linewidth]{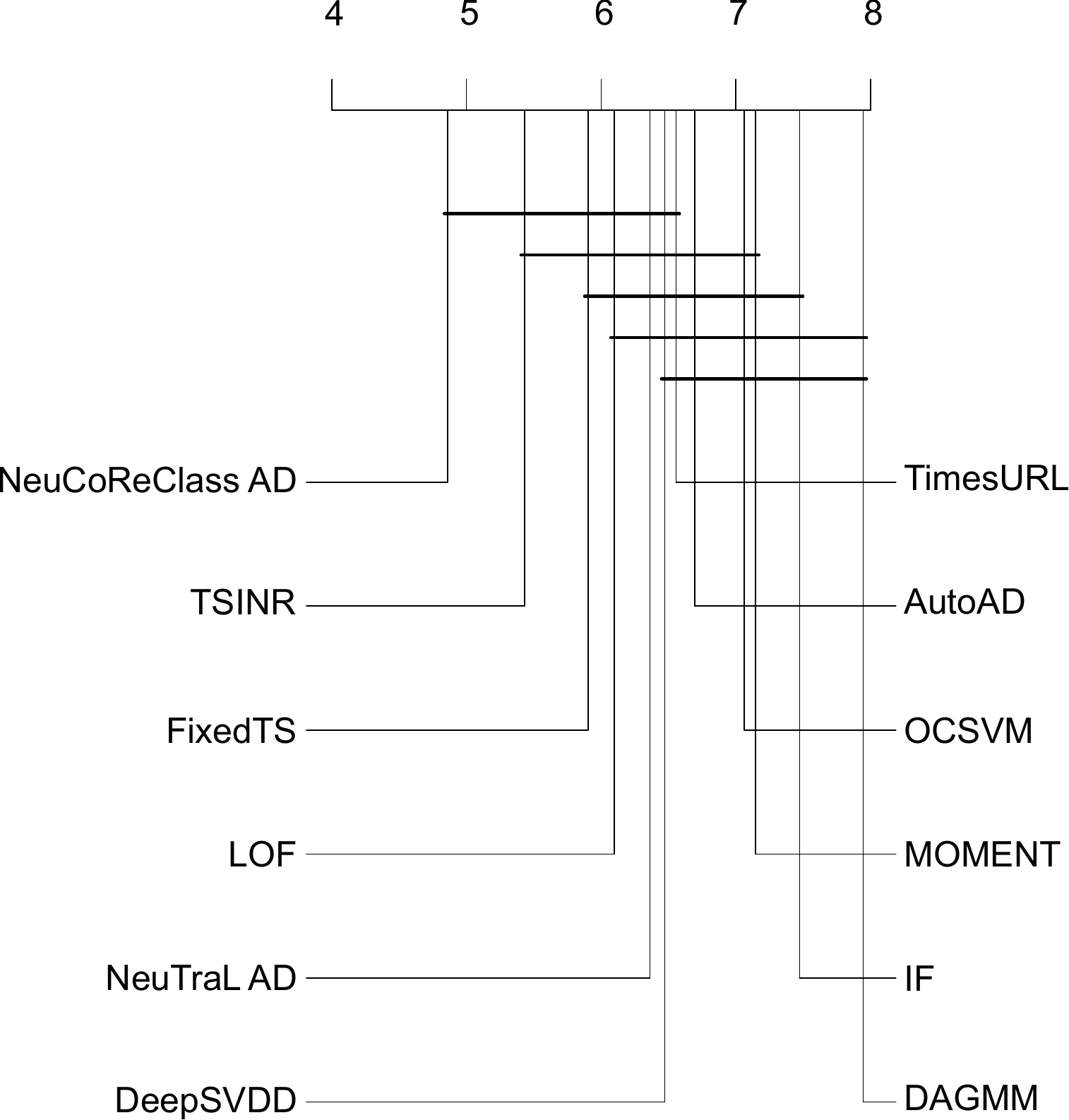}
        \caption{AUROC, \textit{(N-1)-vs-rest} setting.}
    \end{subfigure}
    \hfill
    \begin{subfigure}[t]{0.48\textwidth}
        \centering
        \includegraphics[width=\linewidth]{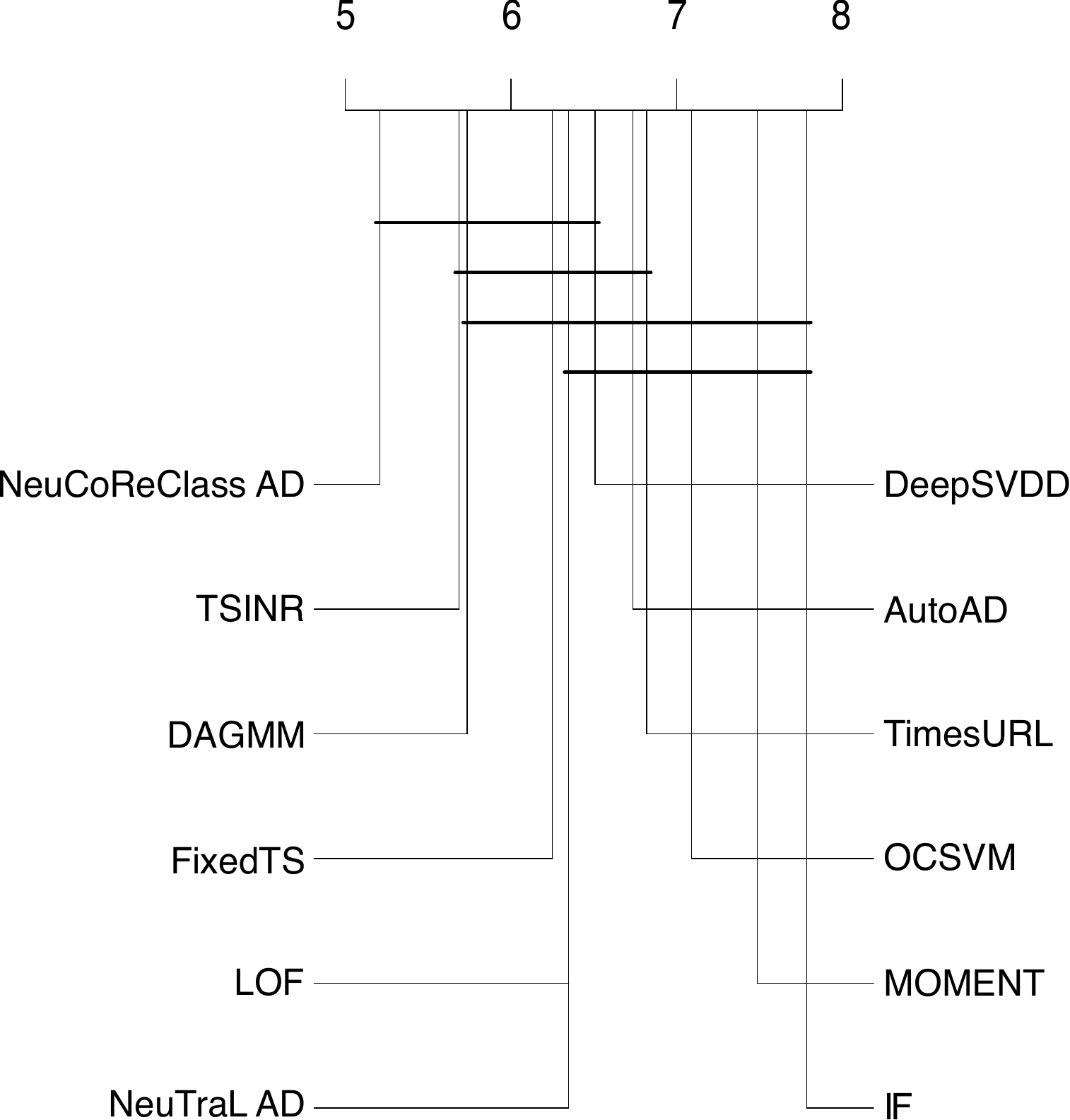}
        \caption{AUPR, \textit{(N-1)-vs-rest} setting.}
    \end{subfigure}

    \caption{
    Statistical comparison of methods using the Friedman test followed by pairwise Wilcoxon signed-rank tests with Shaffer's correction for multiple comparisons. 
    }
    
    \label{fig:statistical_tests}
\end{figure*}

Figure~\ref{fig:statistical_tests} shows the resulting ranking diagrams. These diagrams present the average rank of each method across all problems for both the \textit{one-vs-rest} and \textit{(N-1)-vs-rest} settings, and for the AUROC and AUPR metrics, where lower values indicate better performance. Groups of methods connected by horizontal lines are not significantly different at $\alpha = 0.05$.

The results show that no statistically significant differences are observed among the top-performing methods, nor among the worst-performing ones, across all settings and metrics. However, significant differences do emerge when comparing top-performing methods against the worst-performing ones, indicating that the test is able to discriminate between clearly different levels of performance.

The large number of methods considered in the experimental study has a significant impact on the power of the statistical analysis. The number of pairwise comparisons grows quadratically with the number of methods and, as a result, after correcting for multiple testing, many comparisons fail to reach statistical significance. Therefore, although there are clear differences among top- and worst-performing methods, we do not have sufficient statistical evidence to claim significant differences across all methods. Obtaining such evidence would require including substantially more datasets, which is computationally infeasible.

Despite this, the ranking diagrams are consistent with the conclusions drawn from the performance tables. In particular, NeuCoReClass AD consistently ranks among the top-performing methods. In the \textit{one-vs-rest} setting, its performance remains close to that of DeepSVDD, with only slight differences in both metrics. In contrast, in the \textit{(N-1)-vs-rest} setting, NeuCoReClass AD exhibits a clearer advantage over the second-best method for both metrics.

\section{Model Complexity and Computational Cost}\label{app:cost}

We analyze the computational cost and model complexity of the evaluated methods from both the empirical runtime and the hardware-agnostic perspectives. Table~\ref{tab:time} reports the runtime during training and inference, including total training time, total inference time, and average inference time per sample. Experiments with shallow methods were conducted on a 96-core AMD EPYC 955P processor at 2.6\,GHz with 256\,GB of memory and AVX support. Experiments with the remaining methods were performed on a machine equipped with an NVIDIA RTX A5000 GPU with 24\,GB of memory. The reported times are obtained by averaging over random seeds and aggregating across all problems. 

As expected, classical methods such as IF, LOF, and OCSVM exhibit the lowest computational cost, while deep learning-based approaches introduce a substantial increase in runtime. Recent self-supervised methods, particularly those based on neural transformation learning and large-scale architectures, show significantly higher training costs. Finally, transformer-based methods such as TSINR and MOMENT, as well as TimesURL, present substantially increased computational demands, with MOMENT being the most expensive method in both training and inference, reflecting the cost of large-scale pre-trained architectures.

Table~\ref{tab:complexity} complements this analysis by reporting the number of parameters and forward-pass FLOPs for training and inference, providing a hardware-agnostic estimate of computational cost for the non-shallow baselines. The reported number of parameters corresponds to the average across all datasets, as it may vary depending on the dimensionality of the input time series and certain architecture-dependent components. Similarly, FLOPs depend on the dimensionality of the input data; therefore, they are computed for each dataset and subsequently averaged across all datasets.

FLOPs are computed using the \texttt{fvcore} library~\citep{fvcore}. It is important to note that this estimation may not fully capture the cost of certain operations. In particular, for methods such as TimesURL, which rely on frequency-domain transformations (e.g., Fourier transforms), some computationally intensive operations are not fully accounted for by the profiler. This results in an apparent discrepancy where TimesURL exhibits lower FLOPs than NeuCoReClass AD, while requiring substantially higher training time in practice. However, the FLOPs estimates are generally consistent with the observed runtime across most methods.

Overall, NeuCoReClass AD introduces an additional computational overhead compared to simpler baselines due to its multi-task learning framework. Nevertheless, its training time remains significantly lower than that of computationally demanding methods such as TSINR, TimesURL, and MOMENT, while achieving slightly better performance in the \textit{one-vs-rest} setting and substantially outperforming them in the \textit{(N-1)-vs-rest} setting. This indicates that NeuCoReClass AD achieves a favorable trade-off between computational cost and representational capacity.

\begin{table*}[ht]
\centering
\caption{Empirical runtime of the evaluated methods during training and inference. We report total training time, total inference time, and average inference time per sample.}
\label{tab:time}
{
\begin{tabular}{lccc}
\toprule
Method & Train Time (s) & Inference Time (s) & Inference Time (ms/sample) \\
\midrule
IF & 5.30e+01 & {1.13e+00} & {0.01} \\
{LOF} & {2.80e+02} & {1.51e+01} & {0.18} \\
{OCSVM} & {1.04e+02} & {1.27e+02} & {1.04} \\
{AutoAD} & {1.92e+04} & {1.07e+03} & {4.54} \\
{DAGMM} & {2.20e+04} & {2.40e+03} & {9.23} \\
{DeepSVDD} & {3.03e+04} & {1.10e+03} & {4.65} \\
{FixedTS} & {1.73e+04} & {6.79e+02} & {3.09} \\
{NeuTraL AD} & {5.40e+04} & {3.78e+03} & {16.94} \\
{TimesURL} & {8.56e+05} & {1.24e+03} & {5.11} \\
{MOMENT} & {1.96e+06} & {7.23e+03} & {99.05} \\
{TSINR} & {4.47e+05} &  {3.65e+03} &  {16.36} \\
{NeuCoReClass AD} & {1.33e+05} & {3.82e+03} & {17.15} \\
\bottomrule
\end{tabular}
}

\end{table*}

\arrayrulecolor{black}

\begin{table*}[ht]
\centering
\caption{Model complexity and computational cost of the evaluated methods, measured in number of parameters and forward-pass FLOPs for training and inference.}
\label{tab:complexity}

{
\begin{tabular}{lccc}
\toprule
{Method} & {Params} & {FLOPs (Train)} & {FLOPs (Inference)} \\
\midrule
{AutoAD} & {1.58e+06} & {2.06e+10} & {8.32e+08} \\
{DAGMM} & {4.97e+05} & {6.44e+09} & {2.60e+08} \\
{DeepSVDD} & {1.58e+06} & {8.36e+09} & {8.32e+08} \\
{FixedTS} & {6.45e+05} & {1.00e+11} & {3.37e+08} \\
{NeuTraL AD} & {1.33e+07} & {1.61e+11} & {6.01e+09} \\
{TimesURL} & {1.14e+06} & {1.00e+11} & {3.37e+08} \\
{MOMENT} & {3.41e+08} & {1.79e+13} & {7.60e+11} \\
{TSINR} & {9.02e+07} & {8.40e+11} & {3.52e+10} \\
{NeuCoReClass AD} & {1.42e+07} & {3.08e+11} & {1.20e+10} \\
\bottomrule
\end{tabular}
}

\end{table*}

\arrayrulecolor{black}

\section{Ablation Studies}\label{app:ablation}

To assess the contribution of each self-supervised objective in NeuCoReClass AD, we conduct an ablation study by comparing the full model against six reduced variants. These variants selectively consider one or two of the three training objectives: self-supervised reconstruction, transformation classification, and contrastive learning. The evaluated combinations include: (1) Classification only, (2) Contrastive only, (3) Reconstruction only, (4) Contrastive + Classification, (5) Reconstruction + Classification, and (6) Reconstruction + Contrastive. The results are reported in Table~\ref{tab:ablation_r0} and Table~\ref{tab:ablation_r1}, which correspond to the two evaluation settings used in our experiments: \textit{one-vs-rest} and \textit{(N-1)-vs-rest}, respectively.

Each table shows the average AUROC and AUPR scores across all problems for each ablation setting, together with the performance of the full NeuCoReClass AD model. We also include the relative performance gap (in parentheses) between each variant and the full model.

\begin{table*}[ht]
\centering
\caption{Ablation results under the \textit{one-vs-rest} setting.}
\label{tab:ablation_r0}
\begin{tabular}{lcc}
\toprule
Task Combination & AUROC (\%) & AUPR (\%) \\
\midrule
Classification & 61.19 (-12.03) & 77.12 (-4.30) \\
Contrastive & 67.71 (-5.51) & 79.23 (-2.19) \\
Reconstruction & 72.82 (-0.40) & 81.33 (-0.09) \\
Contrastive + Classification & 65.38 (-7.84) & 78.49 (-2.93) \\
Reconstruction + Classification & 73.18 (-0.04) &\textbf{ 81.50 (+0.08)} \\
Reconstruction + Contrastive & 73.06 (-0.16) & 81.38 (-0.04) \\
\midrule 
NeuCoReClass AD & \textbf{73.22} & 81.42 \\
\bottomrule
\end{tabular}
\end{table*}

\begin{table*}[ht]
\centering
\caption{Ablation results under the \textit{(N-1)-vs-rest} setting.}
\label{tab:ablation_r1}
\begin{tabular}{lcc}
\toprule
Task Combination & AUROC (\%) & AUPR (\%) \\
\midrule
Classification & 54.68 (-11.95) & 34.22 (-12.55) \\
Contrastive & 56.74 (-9.89) & 34.85 (-11.92) \\
Reconstruction & 66.76 (+0.13) & 46.70 (-0.07) \\
Contrastive + Classification & 55.32 (-11.31) & 34.17 (-12.60) \\
Reconstruction + Classification & \textbf{66.85 (+0.22)} & \textbf{47.10 (+0.33)} \\
Reconstruction + Contrastive & 66.68 (+0.05) & 46.67 (-0.10) \\
\midrule
NeuCoReClass AD & 66.63 & 46.77 \\
\bottomrule
\end{tabular}
\end{table*}

Interestingly, we observe that the configuration combining only reconstruction and classification achieves slightly better scores than the full model. This can be attributed to the fact that both contrastive and classification losses encourage similar latent space structure—fulfilling the disruption and diversity properties described in Section~\ref{sec:revisiting}. However, despite this marginal improvement, we argue that the contrastive loss plays a key role in enabling semantic characterization of the anomalies.

\begin{figure*}[ht]
    \centering
    \begin{subfigure}{0.48\textwidth}
        \centering
        \includegraphics[width=\linewidth]{Figures/epilepsy0_originalseparability.pdf}
        \caption{Dimensionality reduction over the original samples in the evaluation set.}
        \label{fig:ablationoriginal_space}
    \end{subfigure}
    \hfill
    \begin{subfigure}{0.48\textwidth}
        \centering
        \includegraphics[width=\linewidth]{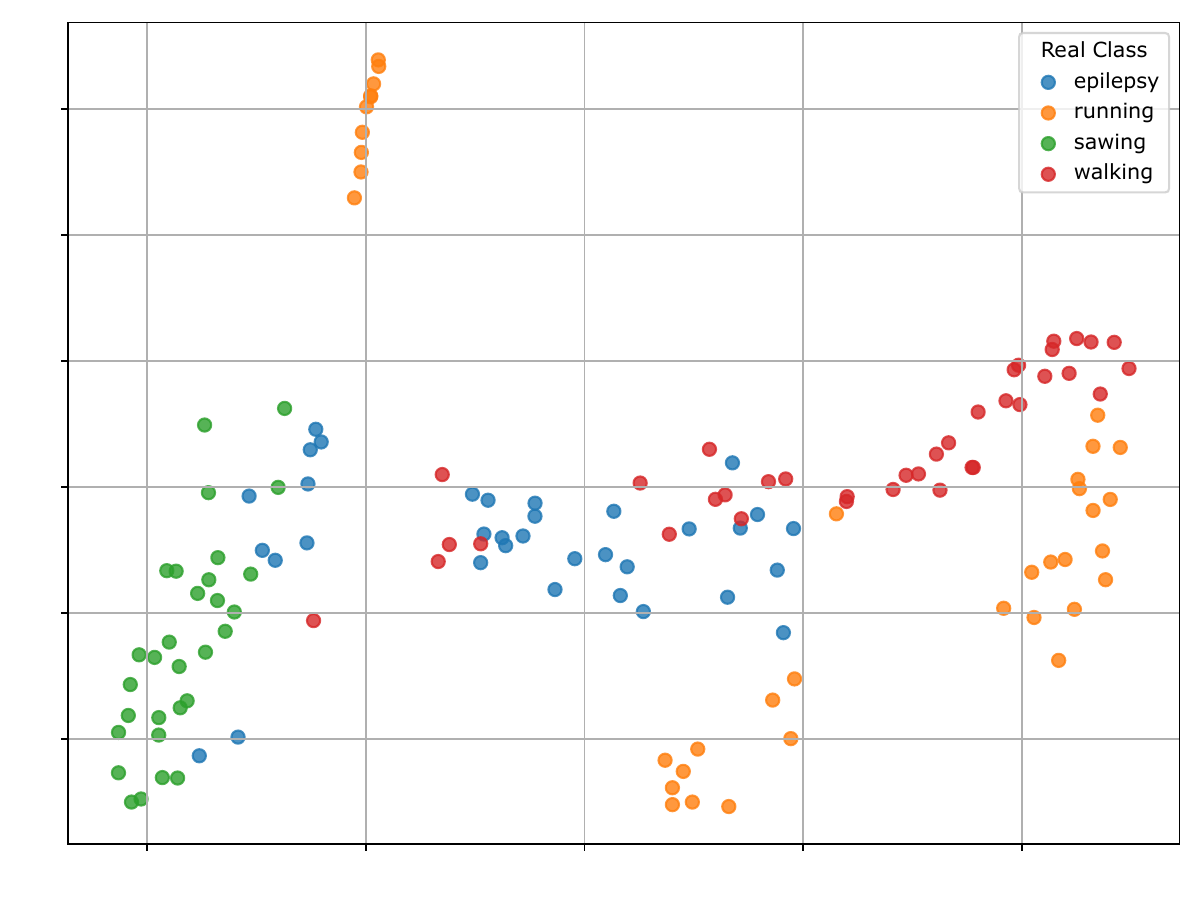}
        \caption{Dimensionality reduction over the transformation-wise anomaly contributions of the ablated model.}
        \label{fig:ablationscore}
    \end{subfigure}
    
    \caption{Class separability in the \textit{Epilepsy} dataset across different feature spaces for the ablated model. Each point corresponds to a sample from the evaluation set, and each color indicates one of the ground-truth classes.}
    \label{fig:ablationvsoriginal}
\end{figure*}

To support this claim, we repeat the characterization experiment introduced in Section~\ref{sec:characterization} using the ablated model that combines only the reconstruction and classification objectives. We conduct this analysis using the \textit{Epilepsy} dataset with the class \textit{sawing} defined as normal and the rest as anomalies. We compare two feature spaces: (i) the original input space, and (ii) the transformation-contribution space, where each sample is represented by the anomaly scores assigned to each transformation. We use t-SNE to project both spaces into two dimensions for visualization, as shown in Figure~\ref{fig:ablationvsoriginal}.

In the original experiment conducted considering the three learning objectives of NeuCoReClass AD, the projection based on the original features reveals no apparent class structure, while the projection based on transformation contributions uncovers clear and well-separated clusters that correspond to distinct anomaly types. In contrast, when repeating this experiment with the ablated model that excludes the contrastive loss, such structure is noT clearly observed. This indicates that the contrastive loss is essential for enriching the characterization of different anomaly types in an unsupervised way.

\section{Sensitivity to Hyperparameters}\label{app:sensitivity}

Fixing hyperparameter values in unsupervised scenarios can be challenging, as ground truth annotations are lacking for validation \citep{fan2020hyperparameter}. Therefore, we fix the key hyperparameters of NeuCoReClass AD to popular values from the literature—the number of neural transformations \(K\) and the temperature \(\tau\) in the contrastive loss—and evaluate its robustness with respect to them. This analysis shows that our method maintains stable performance across different values, thereby demonstrating its robustness and its practical applicability.

\subsection{Sensitivity to Number of Transformations}

\begin{figure*}[h!]
  \centering

  \begin{subfigure}[b]{0.45\textwidth}
    \centering
    \includegraphics[width=\textwidth]{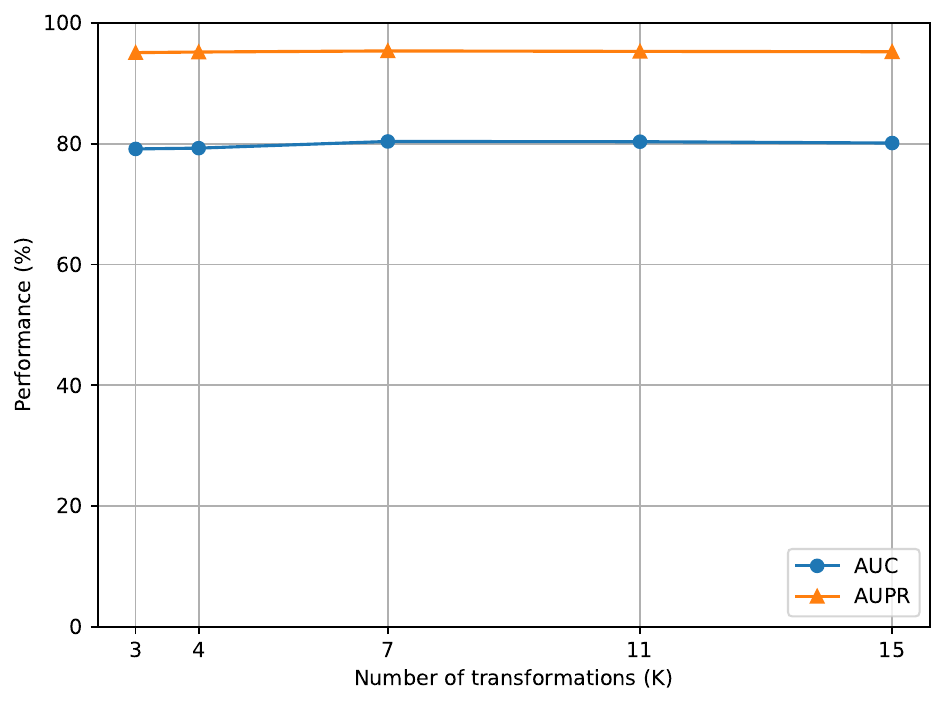}
    \caption{\textit{one-vs-rest} setting.}
    \label{fig:transforms_electric_1vr}
  \end{subfigure}
  \hfill
  \begin{subfigure}[b]{0.45\textwidth}
    \centering
    \includegraphics[width=\textwidth]{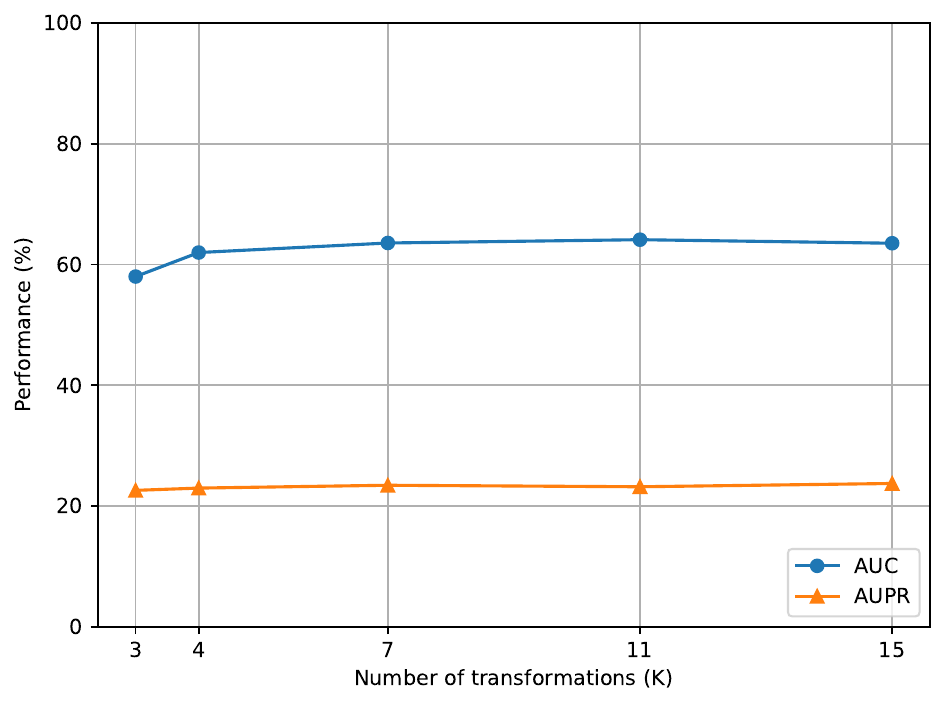}
    \caption{\textit{\((N-1)\)-vs-rest} setting.}
    \label{fig:transforms_electric_n1vr}
  \end{subfigure}

  \caption{Sensitivity of NeuCoReClass AD to the number of transformations \(K\) on the \textit{Electric Devices} dataset.}
  \label{fig:transforms_electric}
\end{figure*}

\begin{figure*}[h!]
  \centering

  \begin{subfigure}[b]{0.45\textwidth}
    \centering
    \includegraphics[width=\textwidth]{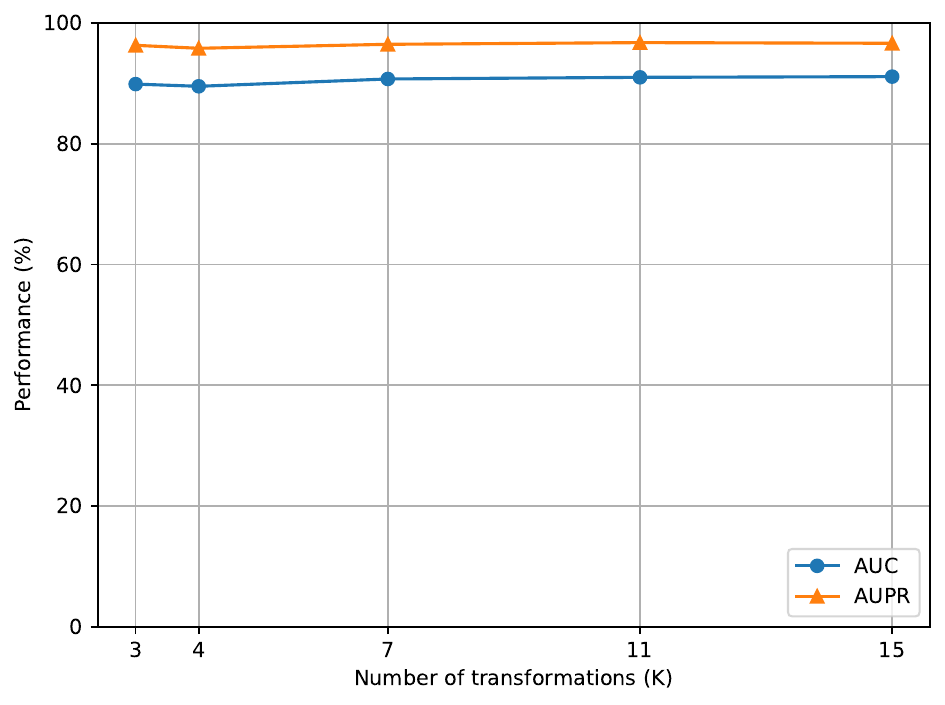}
    \caption{\textit{one-vs-rest} setting.}
    \label{fig:transforms_epilepsy_1vr}
  \end{subfigure}
  \hfill
  \begin{subfigure}[b]{0.45\textwidth}
    \centering
    \includegraphics[width=\textwidth]{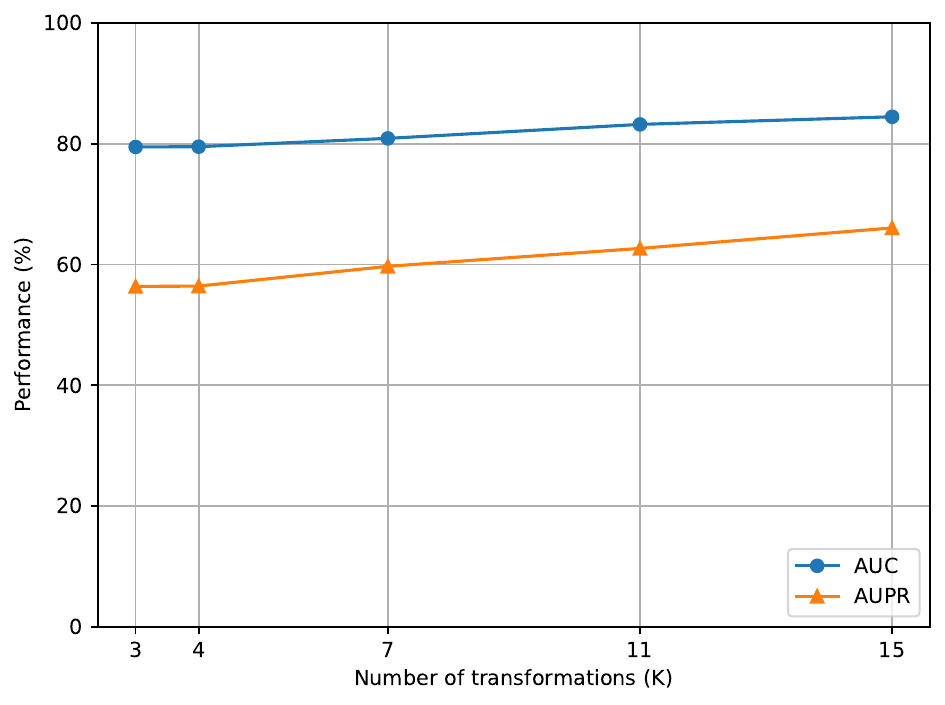}
    \caption{\textit{\((N-1)\)-vs-rest} setting.}
    \label{fig:transforms_epilepsy_n1vr}
  \end{subfigure}

  \caption{Sensitivity of NeuCoReClass AD to the number of transformations \(K\) on the \textit{Epilepsy} dataset.}
  \label{fig:transforms_epilepsy}
\end{figure*}

We vary the number of learnable transformations \(K\) with \(K \in \{3, 4, 7, 11, 15\}\) on the \textit{Electric Devices} (univariate) and \textit{Epilepsy} (multivariate) datasets under both the \textit{one-vs-rest} and the \((N-1)\)-vs-rest settings. For each value of $K$, we train NeuCoReClass AD with five random seeds for each problem derived from each dataset and report the mean AUROC and AUPR in percentage terms. Figures \ref{fig:transforms_electric} and \ref{fig:transforms_epilepsy} show that both metrics improve slightly as \(K\) increases, yet remain largely stable, indicating robustness to the exact choice of transformations. This trend holds across both the one-vs-rest and \((N-1)\)-vs-rest settings.

\subsection{Sensitivity to Temperature}

\begin{figure*}[h!]
  \centering

  \begin{subfigure}[b]{0.45\textwidth}
    \centering
    \includegraphics[width=\textwidth]{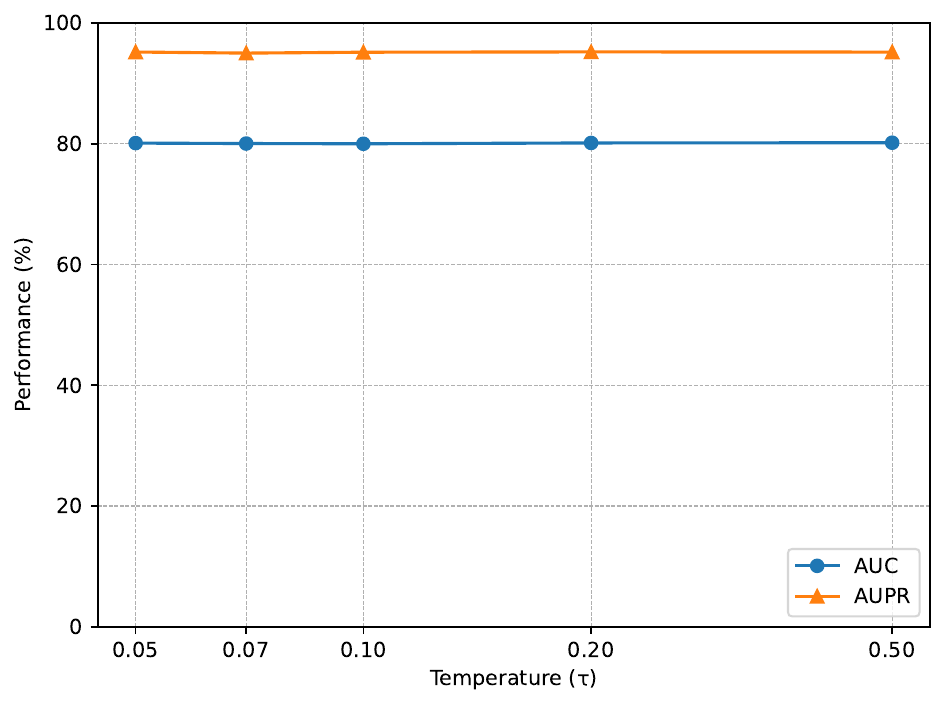}
    \caption{\textit{one-vs-rest} setting.}
    \label{fig:sensitivity_k_1vr}
  \end{subfigure}
  \hfill
  \begin{subfigure}[b]{0.45\textwidth}
    \centering
    \includegraphics[width=\textwidth]{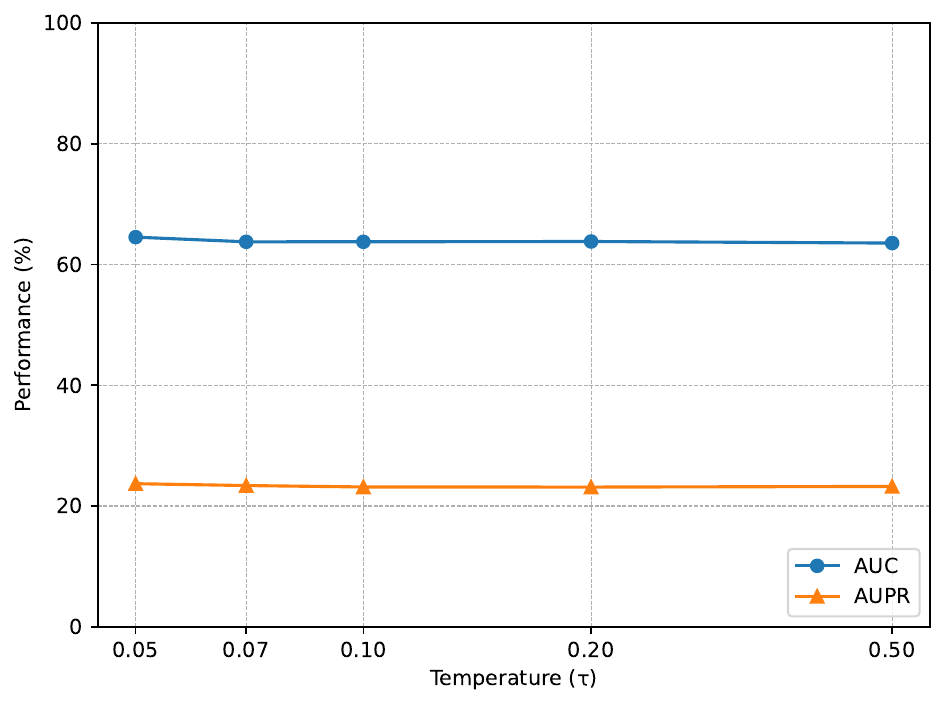}
    \caption{\textit{\((N-1)\)-vs-rest} setting.}
    \label{fig:sensitivity_k_n1vr}
  \end{subfigure}

  \caption{Sensitivity of NeuCoReClass AD to the temperature parameter \(\tau\) on the \textit{Electric Devices} dataset.}
  \label{fig:temps_electric}
\end{figure*}

\begin{figure*}[h!]
  \centering

  \begin{subfigure}[b]{0.45\textwidth}
    \centering
    \includegraphics[width=\textwidth]{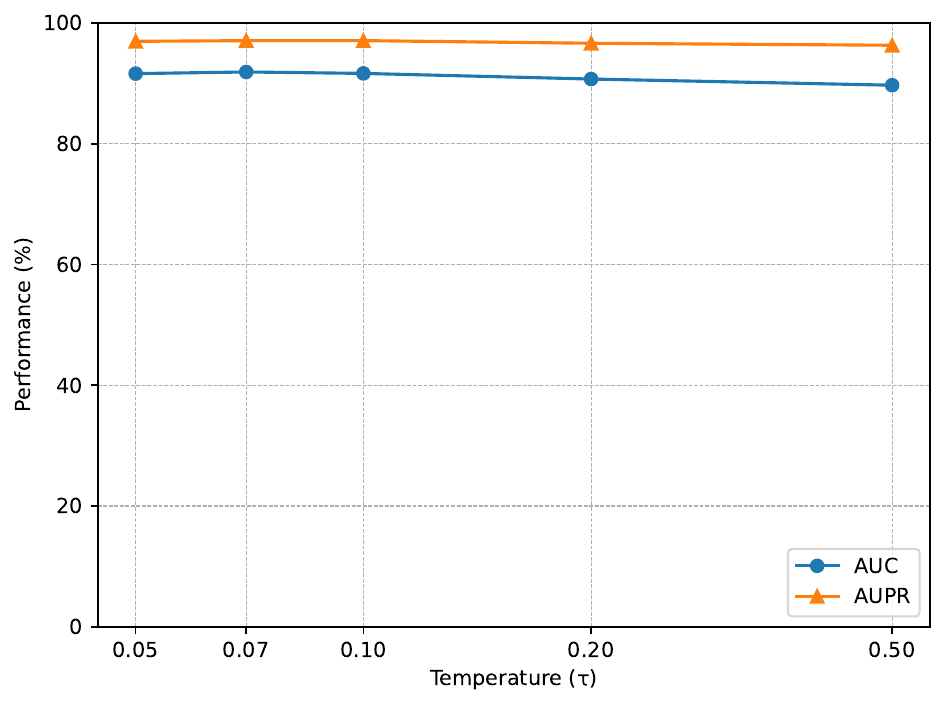}
    \caption{\textit{one-vs-rest} setting.}
    \label{fig:sensitivity_k_1vr}
  \end{subfigure}
  \hfill
  \begin{subfigure}[b]{0.45\textwidth}
    \centering
    \includegraphics[width=\textwidth]{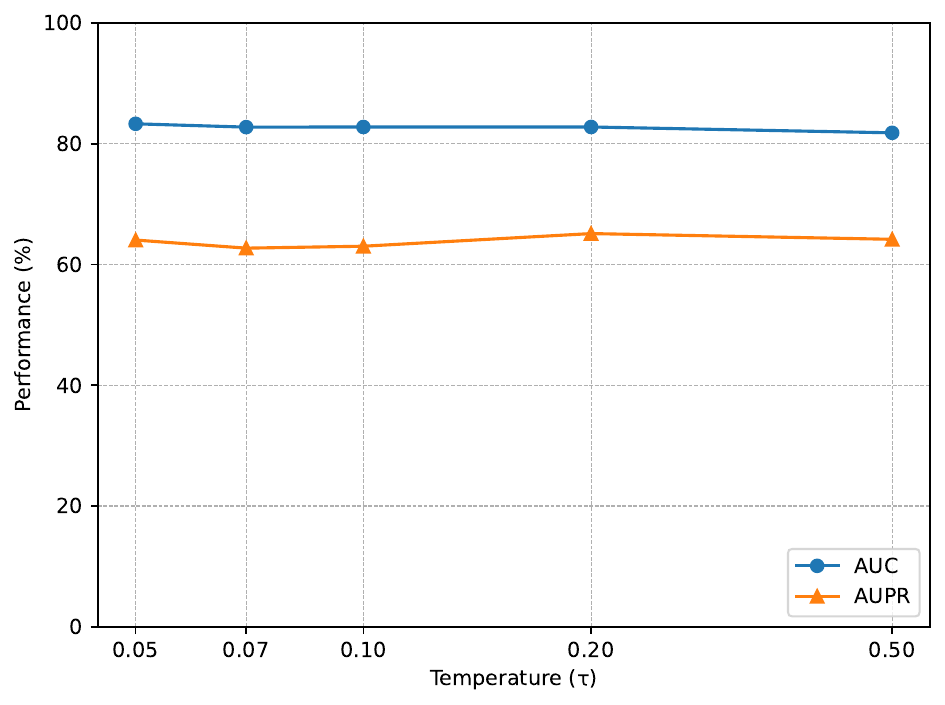}
    \caption{\textit{\((N-1)\)-vs-rest} setting.}
    \label{fig:sensitivity_k_n1vr}
  \end{subfigure}
    \centering
  \caption{Sensitivity of NeuCoReClass AD to the temperature parameter \(\tau\) on the \textit{Epilepsy} dataset.}
  \label{fig:temps_epilepsy}
\end{figure*}

To verify robustness with respect to the contrastive loss temperature \(\tau\), we try different popular values for $\tau$ in the literature, \(\tau\in\{0.05,0.07,0.1,0.2, 0.5\}\), under both evaluation settings on the Electric Devices (univariate) and Epilepsy (multivariate) datasets. For each \(\tau\), NeuCoReClass AD is trained with five seeds in each anomaly detection problem derived from the dataset, and we record mean AUROC and AUPR. Figures \ref{fig:temps_electric} and \ref{fig:temps_epilepsy} plot both metrics versus \(\tau\) (log scale), showing that the method performs consistently for all the values considered.


\end{document}